\documentclass{article}

\usepackage{arxiv}

\usepackage{natbib}

\usepackage{amsmath,amsfonts,bm}

\def\eqref#1{equation~\ref{#1}}

\def\1{\bm{1}}

\def\rvu{{\mathbf{i}}}

\def\rvu{{\mathbf{u}}}
\def\rvv{{\mathbf{v}}}

\def\rvx{{\mathbf{x}}}
\def\rvy{{\mathbf{y}}}

\def\vn{{\bm{n}}}

\def\vx{{\bm{x}}}

\def\vz{{\bm{z}}}

\DeclareMathAlphabet{\mathsfit}{\encodingdefault}{\sfdefault}{m}{sl}
\SetMathAlphabet{\mathsfit}{bold}{\encodingdefault}{\sfdefault}{bx}{n}

\newcommand{\E}{\mathbb{E}}

\usepackage{hyperref}
\usepackage{url}

\usepackage[utf8]{inputenc} 
\usepackage[T1]{fontenc}    
\usepackage{hyperref}       
\usepackage{url}            
\usepackage{booktabs}       
\usepackage{amsfonts}       
\usepackage{nicefrac}       
\usepackage{microtype}      
\usepackage{xcolor}         
\usepackage{amsmath}
\usepackage{amssymb}
\usepackage{amsthm}

\usepackage{listings}

\usepackage{microtype}
\usepackage{graphicx}
\usepackage{subcaption}

\usepackage{booktabs} 
\usepackage{xcolor}

\usepackage{algorithmic}
\usepackage{multirow}
\usepackage[ruled, vlined, linesnumbered]{algorithm2e}

\usepackage{wrapfig}
\usepackage{booktabs}

\usepackage{xcolor}
\usepackage{listings}
\usepackage{inconsolata} 

\definecolor{codebg}{RGB}{248,248,248}
\definecolor{codeframe}{RGB}{220,220,220}
\definecolor{codeblue}{RGB}{33,102,172}
\definecolor{codegreen}{RGB}{0,120,90}
\definecolor{codered}{RGB}{180,60,60}
\definecolor{codegray}{RGB}{120,120,120}

\lstdefinestyle{pyflax}{
  language=Python,
  basicstyle=\ttfamily\fontsize{5}{6}\selectfont,
  numberstyle=\ttfamily\fontsize{3.8}{4.5}\selectfont\color{codegray},
  numbers=left,
  stepnumber=1,
  numbersep=6pt,
  showstringspaces=false,
  tabsize=2,
  breaklines=true,
  breakatwhitespace=true,
  columns=fullflexible,
  keepspaces=true,
  frame=single,
  rulecolor=\color{codeframe},
  backgroundcolor=\color{codebg},
  xleftmargin=1.5em,
  framexleftmargin=1.2em,
  framexrightmargin=0.4em,
  framextopmargin=0.4em,
  framexbottommargin=0.4em,
  captionpos=b,
  keywordstyle=\color{codeblue}\bfseries,
  commentstyle=\color{codegreen}\itshape,
  stringstyle=\color{codered},
}

\title{Dynamic Neural Optimal Transport for Subset Alignment\thanks{Research was sponsored by the Department of the Navy, Office of Naval Research under ONR award numbers N00014-21-1-2300 and N00014-24-1-2259.  Any opinions, findings, and conclusions or recommendations expressed in
 this material are those of the author(s) and do not necessarily reflect the views of the
 Office of Naval Research.} }
\date{September 28, 2026}

\usepackage{authblk}

\author[1]{%
{Bilal Riaz\thanks{\texttt{bilalria@udel.edu}}}%
}
\author[1,2]{%
{Austin J.~Brockmeier\thanks{\texttt{ajbrock@udel.edu}}}%
}
\affil[1]{Department of Electrical and Computer Engineering, University of Delaware, Newark, DE, 19711}
\affil[2]{Department of Computer and Information Sciences, University of Delaware, Newark, DE, 19711}

\renewcommand{\shorttitle}{Dynamic Neural OT for Subset Alignment}

\begin{document}

\maketitle

\begin{abstract}
We study optimal transport between distributions when only a subset of the target population is relevant to the source. The goal is to map the source distribution to a reweighted target distribution whose density ratio with respect to the full target lies in $[0, c]$. This constraint enables the transported distribution to select an appropriate target subpopulation while limiting how strongly any region of the target can be emphasized. We propose a dynamic neural optimal transport framework for this setting using two neural networks. The first parameterizes a time-dependent potential whose spatial gradient defines the velocity field and whose terminal value enforces the density-ratio constraint. The second parameterizes an interpolation between the source distribution and the learned terminal distribution. Unlike prior neural dynamic optimal transport methods, which typically assume a fixed terminal distribution, our formulation learns the terminal distribution within the support of the target. We demonstrate the method on positive-unlabeled learning, unpaired domain translation, and trajectory inference, where meaningful alignments exist between source and target distributions, but direct distribution matching is inappropriate because the target contains additional or irrelevant components.
\end{abstract}

\section{Introduction}
Distribution alignment is a problem of immense importance across machine learning, statistics, and optimization. Wasserstein distance and \textit{f}-divergences are two widely used families of measures of disparity between distributions. Given source and target distributions $\mu$ and $\nu$ defined on subsets of a convex set $\Omega\subseteq \mathbb{R}^d$, specifically on $\operatorname{supp}(\mu)$ and $\operatorname{supp}(\nu)$, respectively, the objective of optimal transport is to obtain a mapping
$T:\operatorname{supp}(\mu)\to\operatorname{supp}(\nu)$
from the source distribution to the target distribution that minimizes the expected transport cost, thereby characterizing the Wasserstein distance. This can be relaxed by mapping the source to a distribution near the target, using an \textit{f}-divergence to quantify the disparity between the output of the map and the target distribution. In particular, we consider a dynamic deterministic formulation of $T$ that uses a neural network to parameterize the velocity field. We also use a particular choice of $f$-divergence that constrains the mapped density to lie within a subset of $\operatorname{supp}(\nu)$ while satisfying a prescribed density-ratio bound. While previous work has studied relaxed neural optimal transport with static transport maps~\citep{gazdieva2023extremal, choi2023generative, choi2024analyzing} and unbalanced dynamic optimal transport with mass growth~\citep{neklyudov2024wasserstein}, there is no existing neural dynamic optimal transport (OT) formulation that learns a map from the source distribution to a subset of the target distribution.

Specifically, our formulation of support subset alignment relies on a relaxation of the Benamou-Brenier problem, formulated in terms of a probability path $\rho_t$ and velocity field $\rvv_t$~\citep{benamou2000computational,santambrogio2015optimal,figalli2021invitation}. For probability measures $\mu,\nu\in\mathcal{P}(\Omega)$, the semi-relaxed dynamic OT problem we study is
\begin{equation} 
\label{benamou_us}
\begin{aligned}
    \inf_{\rho_t,\rvv_t}\!
    &\int_0^1
    {\mathbb{E}}_{\rvx\sim\rho_t}\!\left[
    \frac{1}{2}\|\rvv_t(\rvx)\|^2
    \right]dt,\quad
    \mathrm{s.t.} \quad
    \partial_t\rho_t(\rvx)    \!+\!\operatorname{div}\!\left(\rho_t(\rvx)\rvv_t(\rvx)\right)\!=\!0,\,
    \rho_0\!=\!\mu,\, \rho_{1}\!\leq\! c\nu,
\end{aligned}
\end{equation}
where the initial marginal constraint $\rho_0=\mu$ is preserved, whereas the terminal marginal $\rho_1$ is allowed to vary subject to the density-ratio constraint $\rho_1/\nu\in[0,c]$. This allows some regions of $\operatorname{supp}(\nu)$ to be up-weighted while others are down-weighted or removed. The relaxed terminal constraint is equivalent to a partial optimal transport relaxation of static optimal transport using $f$-divergences, as introduced by \citet{sejourne2023unbalanced}. The case $c=1$ corresponds to prescribing the endpoint marginal as $\rho_1=\nu$. Under suitable regularity, weak solutions of the continuity equation induce deterministic flows via mass conservation \citep[p.~15]{villani2009optimal}, while locally Lipschitz flows induce unique probability paths \citep[Theorem~5.34]{villani2021topics}.

In practice, when only samples are available, we use neural networks to parameterize both the velocity field $\rvv_t$ and the probability path $\rho_t$. Dynamic neural OT is closely connected to flow matching, diffusion-type models, trajectory inference, and generative modeling \citep{lipmanflow,liu2023flow,albergo2023building,tong2020trajectorynet,neklyudov2023action,neklyudov2024wasserstein}. 

To position our formulation, we introduce necessary notation and discuss unbalanced optimal transport (UOT) and the dynamic version. Let $\mathcal{P}(\mathcal{X})$ and $\mathcal{M}_{+}(\mathcal{X})$ denote probability
and finite nonnegative measures on $\mathcal{X}\subseteq\Omega$. For a map $S$,
$S_{\#}\mu$ is the pushforward; for
$\pi\in\mathcal{M}_{+}(\mathcal{X}\times\mathcal{Y})$,
$(\mathrm{pr}_{\mathcal{X}})_{\#}\pi$ and
$(\mathrm{pr}_{\mathcal{Y}})_{\#}\pi$ are its marginals. We write
$\tilde{\nu}\ll\nu$ if $\nu(A)=0\Rightarrow\tilde{\nu}(A)=0$. When absolutely
continuous, we use the same symbol for a measure and its density; thus, if
$d\tilde{\nu}=\tilde{\nu}(\rvx)d\rvx$ and $d\nu=\nu(\rvx)d\rvx$, then
$d\tilde{\nu}/d\nu=\tilde{\nu}/\nu$ on $\{\nu>0\}$. We use $\rho_t$, $\rvv_t$,
and $g_t$ for time-varying measures, velocity fields, and growth fields,
respectively, and $\mathbb{E}$ for expectations under probability measures.

\textbf{Unbalanced optimal transport} (UOT) generalizes balanced OT~\citep{santambrogio2015optimal} by allowing transport between finite nonnegative measures $\mu \in \mathcal{M}_{+}(\mathcal{X})$ and $\nu \in \mathcal{M}_{+}(\mathcal{Y})$ for which $\mu(\mathcal{X})$ and $\nu(\mathcal{Y})$ need not be equal. Let
$f_1,f_2:\mathbb{R}_{+}\to\mathbb{R}\cup\{+\infty\}$ be proper, convex, lower semicontinuous functions satisfying $f_1(1)=f_2(1)=0$. Given $\pi\in\mathcal{M}_{+}(\mathcal{X}\times\mathcal{Y})$, define
$\tilde{\mu}=(\mathrm{pr}_{\mathcal{X}})_{\#}\pi$ and
$\tilde{\nu}=(\mathrm{pr}_{\mathcal{Y}})_{\#}\pi$. The UOT problem is then stated as~\citep{learn_wasserstein_15,sejourne2023unbalanced}
\begin{equation}
\label{static_uot}
\begin{aligned}
    \mathcal{W}_{\mathrm{UOT}}^{\{f_1,f_2\}}(\mu,\nu)
    =
    \inf_{\pi\in\mathcal{M}_{+}(\mathcal{X}\times\mathcal{Y})}
    \int_{\mathcal{X}\times\mathcal{Y}}
    \mathtt{c}(\rvx,\rvy)\,d\pi(\rvx,\rvy)
    +
    \gamma_1\mathcal{D}_{f_1}(\tilde{\mu}\Vert\mu)
    +
    \gamma_2\mathcal{D}_{f_2}(\tilde{\nu}\Vert\nu),
\end{aligned}
\end{equation}
where $\gamma_1,\gamma_2>0$, and $\mathcal{D}_{f_1}$ and $\mathcal{D}_{f_2}$ denote the $f$-divergences corresponding to $f_1$ and $f_2$, respectively. Even when $\mu$ and $\nu$ are probability measures, the optimal solution of \eqref{static_uot} is generally not a coupling of $\mu$ and $\nu$. Thus, UOT relaxes the strict marginal equality constraints of OT by replacing them with $f$-divergence penalties, making it applicable to general finite nonnegative measures. UOT has also been used to estimate transport maps $T$ using neural network parameterizations~\citep{yang2018scalable, choi2023generative, choi2024analyzing, gazdieva2023extremal, gazdieva2024light}.

\textbf{Dynamic UOT} handles general positive measures by explicitly modeling mass creation and destruction through a scalar growth field $g_t$ \citep{chizat2018unbalanced,chizat2018interpolating}, and has been applied to problems in generative modeling and trajectory inference \citep{neklyudov2023action, neklyudov2024wasserstein,neklyudov24a}. Dynamic UOT between $\mu, \nu \in \mathcal{M}_{+}$ is 
\begin{small}
\begin{equation}
\label{dynamic_uot}
\begin{aligned}
    \inf_{\rho_t,\rvv_t, g_t} \! 
    & \! 
    \int_0^1 \! \! \! 
    \int_{\Omega} \! \!
    \left[ \!
    \frac{\|\rvv_t(\rvx)\|^2 \!+\! g_t^2(\rvx)}{2}
    \! \right] \! \!
    d\rho_{t}(\rvx)dt, 
    \,\textrm{s.t.} \, 
    \frac{\partial}{\partial t}\rho_t(\rvx)
    \! + \! 
    \operatorname{div} \! 
    \left( 
    \rho_t(\rvx)\rvv_t(\rvx)
    \right)
    \! \! = \! \! 
    g_{t}(\rvx) \rho_{t}(\rvx), \
    \rho_0\!\!=\!\!\mu, \
    \rho_1\!\!=\!\!\nu. \! \!
\end{aligned}
\end{equation} 
\end{small}
Thus, UOT penalizes deviations from prescribed marginals, whereas dynamic UOT modifies the continuity equation through a mass growth term. When $\mu$ and $\nu$ are probability measures and mass creation/destruction is not needed, one may set $g_t=0$, recovering balanced dynamic OT.

\textbf{Positioning and contributions.}
Unlike UOT, which allows positive measures and relaxes marginal constraints, and dynamic UOT, which models mass creation/destruction but preserves marginals, our dynamic subset alignment preserves the source mass and transports it to a terminal distribution supported on a subset of the target support and within a density-ratio bound of the target. This is achieved by replacing the terminal marginal equality constraint with an $f$-divergence-based penalty, which recovers balanced OT as a special case. We derive a dual dynamic formulation whose potential identifies the selected target subset and whose gradient defines the transport velocity. We use 
neural network-based parameterizations of the velocity field and path distribution to achieve the alignment and subset selection. Furthermore, we demonstrate the framework on unpaired domain translation, PU-learning, and trajectory inference problems.

\section{Related Work}
Our work on dynamic subset alignment is most directly related to Lagrangian neural optimal transport~\citep{pooladian2024neural}, action matching~\citep{neklyudov2023action}, and Wasserstein Lagrangian flows~\citep{neklyudov24a}. The neural optimal transport framework with Lagrangian costs~\citep{pooladian2024neural} studies optimal transport with general Lagrangian potentials in Euclidean space. Wasserstein Lagrangian flows~\citep{neklyudov24a} are primarily developed for applications in cellular trajectory inference and quantum many-body problems and extend to more general settings such as the Wasserstein-Fisher-Rao metric~\citep{chizat2018interpolating, chizat2018unbalanced, sejourne2023unbalanced}, with the ability to model mass growth/destruction and different types of dynamics.

These approaches, as well as standard flow-based models, are developed for settings in which the endpoint marginals are prescribed, which prevents their direct application to the subset-alignment problem. A more extensive review of recent work on dynamic neural optimal transport is included in Appendix~\ref{app:dynamic-survey}.  

In contrast, our proposed dynamic support subset-selection framework preserves one marginal while dynamically transporting its mass to a subset of the support of the other marginal while minimizing the transport cost. Therefore, our approach can be viewed as a novel extension of prior work~\citep{neklyudov24a}. Although we focus on the $\ell_2^2$ cost, our method is compatible with other Lagrangian costs~\citep{pooladian2024neural}, which could be useful for incorporating side information, as in semi-supervised domain adaptation.

Additionally, we review advances in static neural optimal transport in Appendix~\ref{app:static-survey}. In Appendix~\ref{static-derive}, we also discuss the static formulation of subset alignment. The static formulation of our framework can be related to incomplete optimal transport (IOT)~\citep{gazdieva2023extremal} and unbalanced optimal transport mapping (UOTM)~\citep{choi2023generative}. IOT~\citep{gazdieva2023extremal} can be interpreted as a proximal form of static subset alignment. UOTM can also be reduced to static subset alignment between two probability measures by choosing, in \eqref{static_uot}, $f_1$ as the $f$-divergence corresponding to the indicator function
$\imath_{\{1\}}(x)=
    \begin{cases}
    0, & x=1,\\
    +\infty, & x\neq 1,
    \end{cases}$
and $f_2=\imath_{[0,c]}$.

\section{Methodology}
\label{Dynamic subset-selection}
In our semi-relaxed dynamic OT formulation~\eqref{benamou_us}, we can replace the inequality $\rho_{1}\leq c\nu$ constraint by adding the equivalent $f$-divergence penalty term $\mathcal{D}_{{\imath_{[0,c]}}}$ to the objective, where $\mathcal{D}_{\imath_{[0,c]}}$ is a special case of the range divergence $\mathcal{D}_{\imath_{[a,b]}}$ corresponding to the convex indicator function $\imath_{[a,b]}$, which, along with its Legendre--Fenchel conjugate $\imath^{*}_{[a,b]}$, is given as 
\begin{equation}
    {\imath_{[a,b]}}(r) =  \begin{cases}
        0, & r \in [a, b] \\
        +\infty, & \text{o.w.}
    \end{cases}, \quad     \quad  \imath^{*}_{[a,b]}(t) = \sup_{u \in [a,b]} (u \cdot t) =\max (-at, bt).
\end{equation}
Setting $a=0$ and $b=c$, one obtains $\mathcal{D}_{\imath_{[0,c]}}(\rho_1\Vert \nu)$ with $\imath^{*}_{[0,c]}(u)=c\cdot \max(0, u)$. Since $\imath_{[0,c]}$ is convex and lower semicontinuous, $\imath_{[0,c]}=\imath_{[0,c]}^{**}$. The variational form of $f$-divergences \citep{nguyen2010estimating,nowozin2016f}, \citep[Theorem 7.26]{Polyanskiy_Wu_2024} $\mathcal{D}_{f}(\tilde{\nu}\Vert\nu)
 = \sup_{\eta:\Omega\to\mathbb{R}}
\int_{\Omega}\eta(\rvx)d\tilde{\nu}(\rvx)-
\int_{\Omega}f^{*}(\eta(\rvx))d\nu(\rvx)$,
yields
\begin{equation} \label{eq:f_0c}
    \mathcal{D}_{\imath_{[0,c]}}(\rho_{1}\Vert\nu) =\! \sup_{\eta:\Omega \rightarrow \mathbb{R}} \underset{\rvx\sim \rho_{1}}{\mathbb{E}}\left[\eta(\rvx)\right] -\!\!  \underset{\rvx\sim\nu}{\mathbb{E}}\left[c \cdot \max(0, \eta(\rvx))\right].
\end{equation}
To solve the relaxed dynamic optimal transport problem in \eqref{benamou_us}, we replace the terminal inequality constraint with the additive penalty
$\mathcal{D}_{\imath_{[0,c]}}(\rho_1\Vert\nu)$ in the objective. We introduce the Lagrange multiplier $\varphi_t$ for the continuity equation and the Lagrange multiplier $\psi_0$ for the initial marginal constraint $\rho_0=\mu$. After further simplification (see Appendix~\ref{Derivation of Dynamic Neural Subset Selection} for details), one obtains the conditions.
\begin{align}
\rvv_t(\rvx) = \nabla\varphi_t(\rvx), \
\psi_0(\rvx) = \varphi_0(\rvx), \
\eta(\rvx) = -\varphi_1(\rvx).
\end{align}
These conditions can then be used to eliminate $\rvv_t(\rvx)$, $\psi_0(\rvx)$, and $\eta(\rvx)$, yielding the optimization problem
$\sup_{\rho_t}\inf_{\varphi_t}\mathcal{L}(\rho_t,\varphi_t)$ with the Lagrangian
\begin{small}
\begin{equation}
\label{opt_lag_final}
\mathcal{L}(\rho_t, \varphi_t) \!= \!
\overbrace{
\underset{\rvx \sim {\mu}}{\mathbb{E}}\!\big[ \varphi_0(\rvx) \big]
+\underset{\rvx \sim {\nu}}{\mathbb{E}}\!\big[c\cdot \max(0,-\varphi_{1}(\rvx))\big]
}^{\text{boundary part}} 
+ \overbrace{\int_{0}^{1} \underset{\rvx_t \sim \rho_{t}}{\mathbb{E}}
\left[\frac{\partial}{\partial t} \varphi_{t}(\rvx_t)
+ \frac{\Vert \nabla \varphi_{t}(\rvx_t)\Vert^{2}}{2} \right] dt}^{\text{intermediate part}} .
\end{equation}
\end{small}
\begin{figure*}[t]
    \centering
    \includegraphics[width=1.0\linewidth]{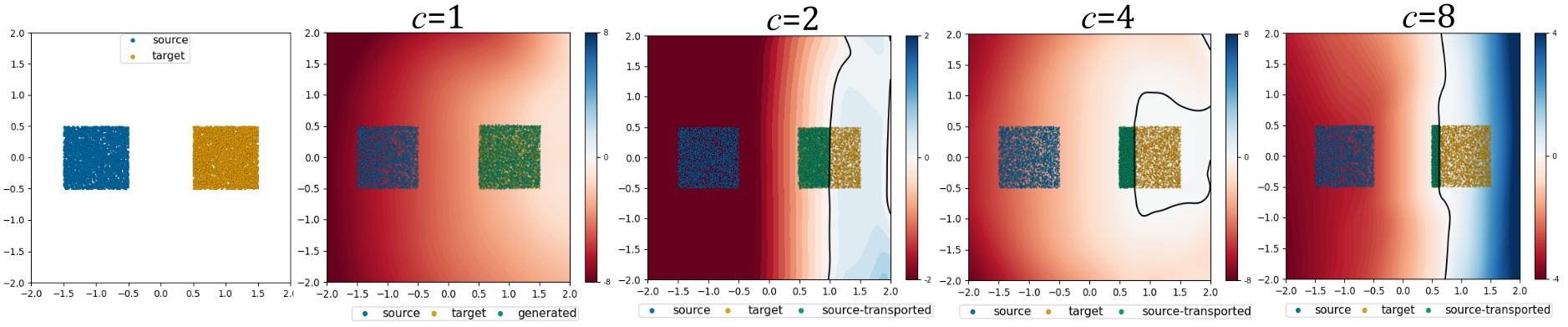}
    \caption{Subset alignment between two uniform distributions in $\mathbb{R}^{2}$ at $t=1$, obtained using Algorithm~\ref{neural-dynamic_0} with $c\in\{1,2,4,8\}$. The black contour line indicates the zero-level set of $\varphi_{1}(\rvx)$.}
    \label{squares-final-all}
\end{figure*}
To approximate the solution, we directly parameterize the dual potential $\varphi_t$ with a neural network and parameterize the probability path $\rho_t$ using a neural network that acts on samples from the terminal marginals $\mu$ and $\nu$. A similar parameterization is used in the computational framework for learning Wasserstein--Lagrangian flows (WLF)~\citep{neklyudov24a}. In contrast, flow-based models typically do not learn $\rho_t$ directly, but instead use analytically defined interpolants between two fixed marginals during training~\citep{lipmanflow, liu2023flow, albergo2023building}. Our problem differs from both flow matching and WLF because we aim to map the source distribution $\mu$ to an unknown distribution $\rho_1^*$ supported on a subset of $\operatorname{supp}(\nu)$. To this end, for $t\in[0,1]$, we construct an interpolant $\rvx_t\sim\rho_t$ from independently sampled points $\rvx_0\sim\mu$ and $\rvx_1\sim\nu$ as
\begin{equation}
\label{eq:dynamic-interp}
\rvx_t = (1-t)\rvx_0 + t\left(\rvx_1 + \mathit{Q}_t(\rvx_0, \rvx_1) \right),    
\end{equation}
where $\mathit{Q}_t$ is a time-dependent neural network. At $t=1$, we have
$\rvx_1+\mathit{Q}_1(\rvx_0,\rvx_1)\sim \rho_1$. Since in general, $\rho_1\neq\nu$, the network must learn a new terminal marginal. Let $\Phi_t:\Omega\to\Omega$ denote the flow induced by the velocity field $\rvv_t$, so that $(\Phi_t)_{\#}\mu=\rho_t$ and $
 \Phi_t(\rvx_0) =
\rvx_0+\int_0^t \rvv_\tau(\rvx_\tau)\,d\tau,
\qquad
\rvx_0\in\operatorname{supp}(\mu)$. The optimal correction $\mathit{Q}_t^\star$ corresponding to this flow is then given by $\mathit{Q}_1^\star(\rvx_0,\rvx_1)=
\Phi_1(\rvx_0)-\rvx_1$ and $\mathit{Q}_t^\star(\rvx_0,\rvx_1)=
\frac{1}{t}\Big(\Phi_t(\rvx_0)-(1-t)\rvx_0\Big)-\rvx_1 $. By the standard neural-network universal approximation theorem~\citep{hornik1991}, bounds on the $\epsilon$-approximation error $\|\mathit{Q}_t-\mathit{Q}_t^\star\|$ translate into bounds on $\|\Phi_t-\Phi_t^\star\|$. Consequently, for any admissible velocity field, one obtains
$
W_2\!\left((\Phi_t)_{\#}\mu,\rho_t\right)\leq C\epsilon
$
for some constant $C>0$.

Range-based penalty~\ref{eq:f_0c} also allows us to distinguish between data points sampled from the selected and rejected subsets of the target support $\operatorname{supp}(\nu)$. Let $\Phi_t^{\star}$ denote the optimal flow operator corresponding to the optimal dual potential $\varphi_t^{\star}$. Then points sampled from the selected subset $\operatorname{supp}((\Phi_1^{\star})_\# \mu)$ and from the rejected subset $\operatorname{supp}(\nu) \setminus \operatorname{supp}((\Phi_1^{\star})_\# \mu)$ can be separated by the following test
\begin{equation}
    \begin{aligned}
       \varphi_{1}^{\star}(\rvx) \leq 0, & \quad \text{almost surely on } \operatorname{supp}((\Phi_1^{\star})_\# \mu), \\
       \varphi_{1}^{\star}(\rvx) > 0, & \quad \text{almost surely on } \operatorname{supp}(\nu) \setminus \operatorname{supp}((\Phi_1^{\star})_\# \mu).
    \end{aligned} \label{sp_d_0}
\end{equation}
Test~\ref{sp_d_0} follows directly from the optimality conditions for the range penalty and dual $\varphi_t$ (see Appendix~\ref{appendix: Thresholding for PU{-}Learning and Rejection Sampling}) and is used in our PU-learning and melanoma experiments.

\begin{algorithm}[ht]
{\footnotesize
\caption{\texttt{Dynamic-Neural-SS}: Learning Algorithm for Dynamic Subset Selection}
\label{neural-dynamic_0}
\begin{algorithmic}[1]

\REQUIRE Samples from distributions $\mu$ and $\nu$, time-dependent neural network
$\varphi_t(\cdot;\theta_{\varphi})$, interpolant network
$\mathit{Q}_t(\cdot,\cdot;\theta_\rho)$, batch size $N$, number of updates
$n_\varphi$ and $n_\rho$, gradient-based optimizers $\mathrm{optim}_\varphi$
and $\mathrm{optim}_\rho$.

\ENSURE Sample-based neural estimate for $\varphi_t(\cdot;\theta_{\varphi})$.

\FOR{learning iteration $k = 0,1,\ldots$}
  \FOR{$u = 1,\ldots,n_\varphi$}

    \STATE Sample $\{\rvx_0^i\}_{i=1}^N \sim \mu$, $\{\rvx_1^i\}_{i=1}^N \sim \nu$, and $\{t^i\}_{i=1}^N \sim \mathrm{Uniform}([0,1])$.

    \STATE $\displaystyle
    \{{\rvx}_t^i \leftarrow (1-t^i)\rvx_0^i +
    t^i({\rvx}_1^i + Q_{t^i}(\rvx_0^i,\rvx_1^i;\theta_\rho))\}_{i=1}^{N}$.

    \STATE $\displaystyle
    \mathrm{grad}_{\theta_{\varphi}} \leftarrow  
    \nabla_{\theta_{\varphi}}\frac{1}{N}\sum_{i=1}^{N}\bigg[
    \frac{\partial}{\partial t}\varphi_{t^i}({\rvx}_t^i;\theta_{\varphi})
    +\frac{\|\nabla\varphi_{t^i}({\rvx}_t^i;\theta_{\varphi})\|^2}{2}
    +\varphi_0(\rvx_0^i;\theta_{\varphi})
    +c\max\!\Bigl(0,-\varphi_1(\rvx_1^i;\theta_{\varphi})\Bigr)
    \bigg]$.

    \STATE Using $\mathrm{optim}_\varphi$ and $\mathrm{grad}_{\theta_{\varphi}}$ to update $\theta_{\varphi}$ via gradient descent.
  \ENDFOR

  \FOR{$u = 1,\ldots,n_\rho$}

    \STATE Sample $\{\rvx_0^i\}_{i=1}^N \sim \mu$, $\{\rvx_1^i\}_{i=1}^N \sim \nu$, and $\{t^i\}_{i=1}^N \sim \mathrm{Uniform}([0,1])$.

    \STATE $\displaystyle
    \{{\rvx}_t^i \leftarrow (1-t^i)\rvx_0^i +
    t^i({\rvx}_1^i + Q_{t^i}(\rvx_0^i,\rvx_1^i;\theta_\rho))\}_{i=1}^{N}$.

    \STATE $\displaystyle
    \mathrm{grad}_{\theta_{\rho}} \leftarrow
    \nabla_{\theta_{\rho}}\frac{1}{N}\sum_{i=1}^{N}\bigg[
    \frac{\partial}{\partial t}\varphi_{t^i}({\rvx}_t^i(\theta_{\rho});\theta_{\varphi})
    +\frac{\|\nabla\varphi_{t^i}({\rvx}_t^i(\theta_{\rho});\theta_{\varphi})\|^2}{2}
    \bigg]$.

    \STATE Update $\theta_{\rho}$ using $\mathrm{optim}_\rho$ and $\mathrm{grad}_{\theta_{\rho}}$ via gradient ascent.
  \ENDFOR
\ENDFOR

\end{algorithmic}
}
\end{algorithm}

\section{Experiments and Results}~\label{Experiments and Results}
We present experimental results for subset alignment on an easily interpretable image-to-image task, where the MNIST dataset~\citep{MNIST} is the source and the EMNIST dataset~\citep{cohen2017emnist} is the target. In this setting, digit images form a subset of the characters in EMNIST.
We also applied dynamic subset alignment to the PU-Learning task.
We then apply our proposed approach to domain translation on the FFHQ dataset~\citep{karras2019style} in the 512-dimensional latent space of ALAE~\citep{pidhorskyi2020adversarial}. We use Algorithm~\ref{neural-dynamic_0} to solve all the subset alignment problems discussed above. We also extended dynamic subset alignment to a multi-marginal setting and applied it melanoma dataset from ~\citet{bunne2023learning}.

\subsection{Toy Examples}
\begin{figure*}[t]
    \centering
    \vspace{-4pt}
    \includegraphics[
        width=0.8\textwidth]{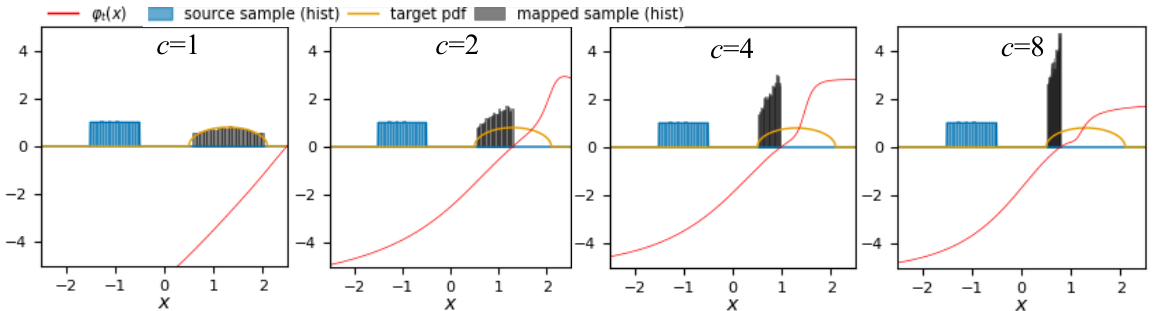}
    \caption{Source, target, and transported source histograms with sampling densities for 1D subset alignment between a uniform distribution and a semicircular density.  $\varphi_1$ is shown in red.}
    \label{uniform2semicircle-final-all}
\end{figure*}

First, we apply subset alignment to 1D and 2D point distributions. For the 1D toy example, we align a uniform distribution with a 1D semicircular distribution. Results for $c\in\{1,2,4,8\}$ are shown in Figure~\ref{uniform2semicircle-final-all}. As the parameter $c$ increases, the source distribution $\mu$ is transported to a subset of the support of the target distribution~$\nu$, for which the potential satisfies $\varphi_{1} \le 0$. The transport between two 2D uniform distributions for $c\in\{1,2,4,8\}$ is shown in Figure~\ref{squares-final-all}, where one can observe that $\varphi_1 (\vx)$ is non-positive in selected subsets. Additional toy examples are provided in Appendix~\ref{app:add-toy}, including subset alignment between two 1D uniform distributions and between two bimodal 2D distributions.

\subsection{MNIST to EMNIST Domain Translation} \label{mnist_to_emnist}
\vspace{-4pt}
The MNIST dataset contains 60,000 images of digits from 0 to 9 in the training set and 10,000 images in the test set. The MNIST dataset is roughly balanced, in the sense that the proportions of the classes are approximately the same. The EMNIST ByClass (EMNIST) dataset contains English letters and digits. EMNIST consists of 62 imbalanced classes, of which 10 classes correspond to digits between 0--9, while the remaining 52 classes correspond to uppercase and lowercase letters of the English alphabet. Roughly, 16\% of EMNIST consists of digits, while the remaining 84\% consists of letters.

Since our goal is to transfer MNIST images to EMNIST images in such a way that MNIST digits are mapped to EMNIST digits while ignoring letters, we trained a neural network classifier to distinguish between digits and letters and used it to evaluate the learned mapping; see Appendix~\ref{app:classifier} for implementation details. After training the classifier, we used both static and dynamic subset-selection approaches for domain translation between MNIST and EMNIST. Implementation details of the underlying models and their training are provided in Appendix~\ref{app:emnist}.
\begin{table}[htb]
\vspace{-2pt}
\small
\caption{Classification accuracies of translated images for MNIST$\rightarrow$EMNIST.}
\label{tab:emnist_results}
\centering
\begin{tabular}{lcccc}
\hline
\textbf{Method} & \textbf{$c=1$} & \textbf{$c=2$} & \textbf{$c=4$} & \textbf{$c=8$} \\
\hline
static~\citep{gazdieva2023extremal} & 46.93 & 75.33 & 82.44 & 87.32 \\
\hline
dynamic (one step) & 64.85 & 75.16 & 93.57 & 95.84 \\
dynamic (ODE) & 58.80 & 70.47 & 92.68 & 95.00 \\
\hline
\end{tabular}
\end{table}
We evaluated classification accuracy on translations of the entire MNIST dataset; the corresponding confusion matrices are provided in Appendix~\ref{appendix-mnist-emnist}. Accuracies of the translated outputs are in Table~\ref{tab:emnist_results}.

\begin{figure}[htb]
    \centering
    \begin{subfigure}[t]{0.30\textwidth}
        \centering
        \includegraphics[width=\linewidth]{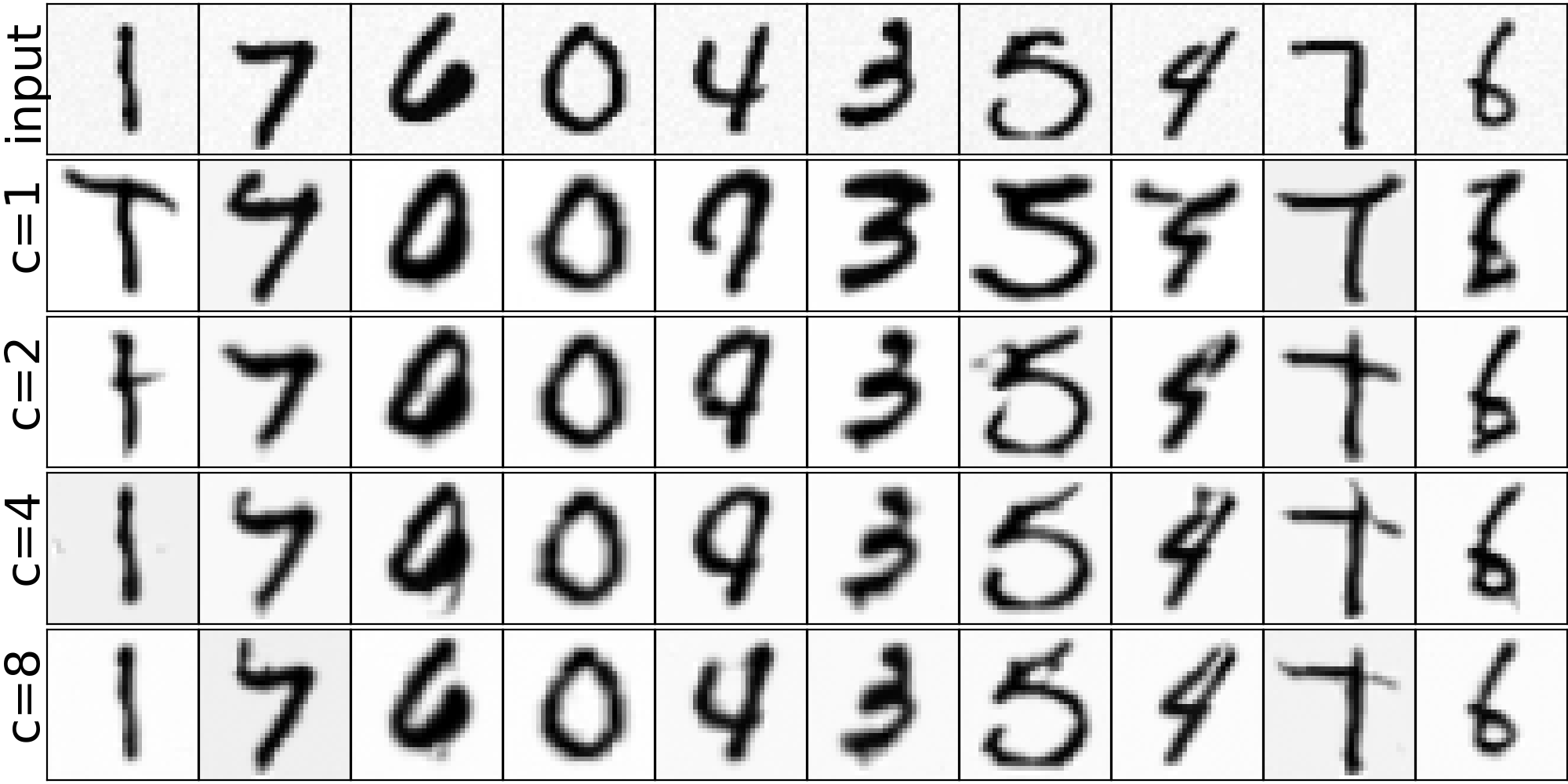}
        \caption{static \citep{gazdieva2023extremal}}
        \label{fig:emnist-static}
    \end{subfigure}\hfill
    \begin{subfigure}[t]{0.30\textwidth}
        \centering
        \includegraphics[width=\linewidth]{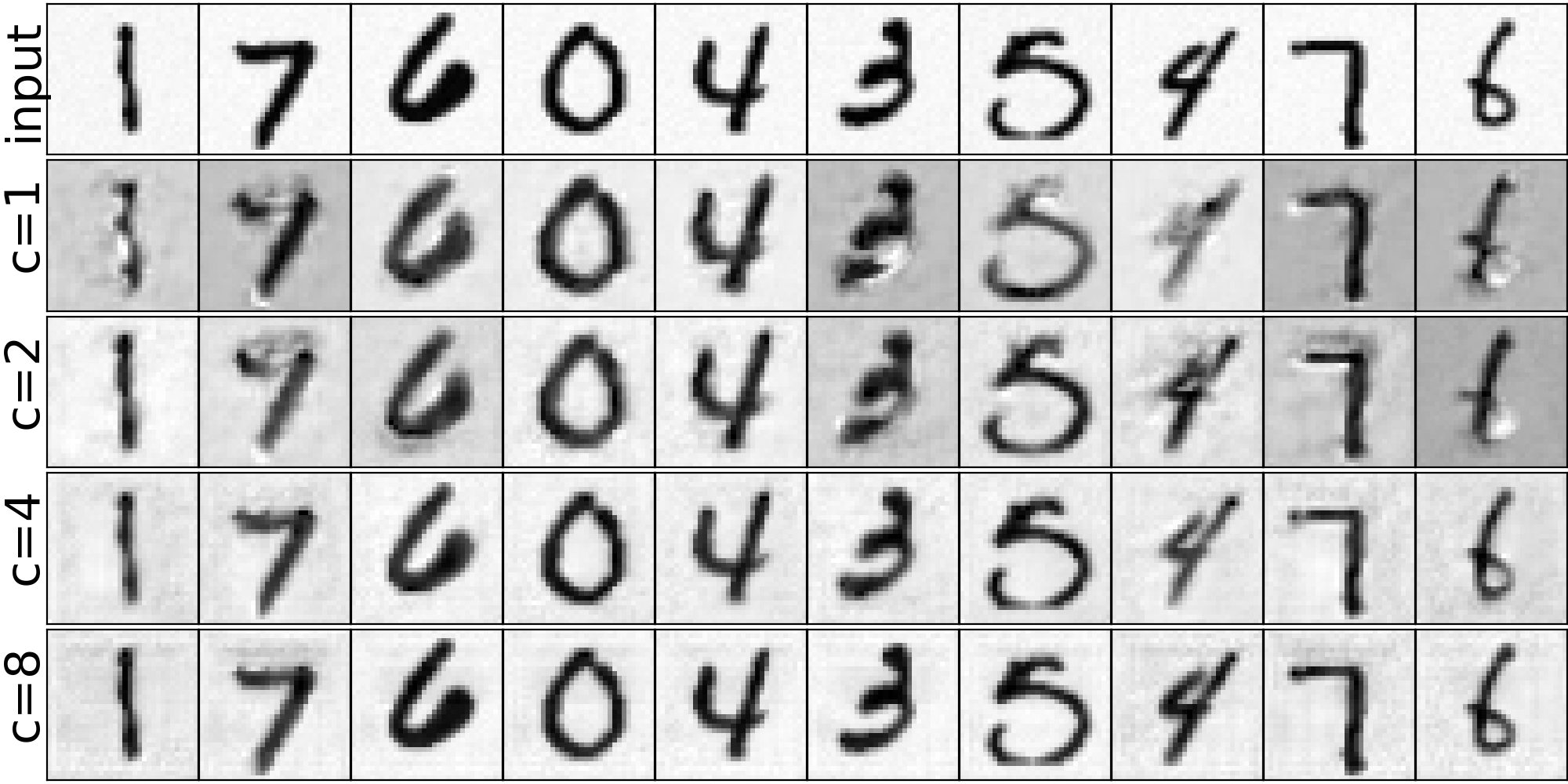}
        \caption{dynamic (one step)}
        \label{fig:emnist-dyn-one}
    \end{subfigure}\hfill
    \begin{subfigure}[t]{0.30\textwidth}
        \centering
        \includegraphics[width=\linewidth]{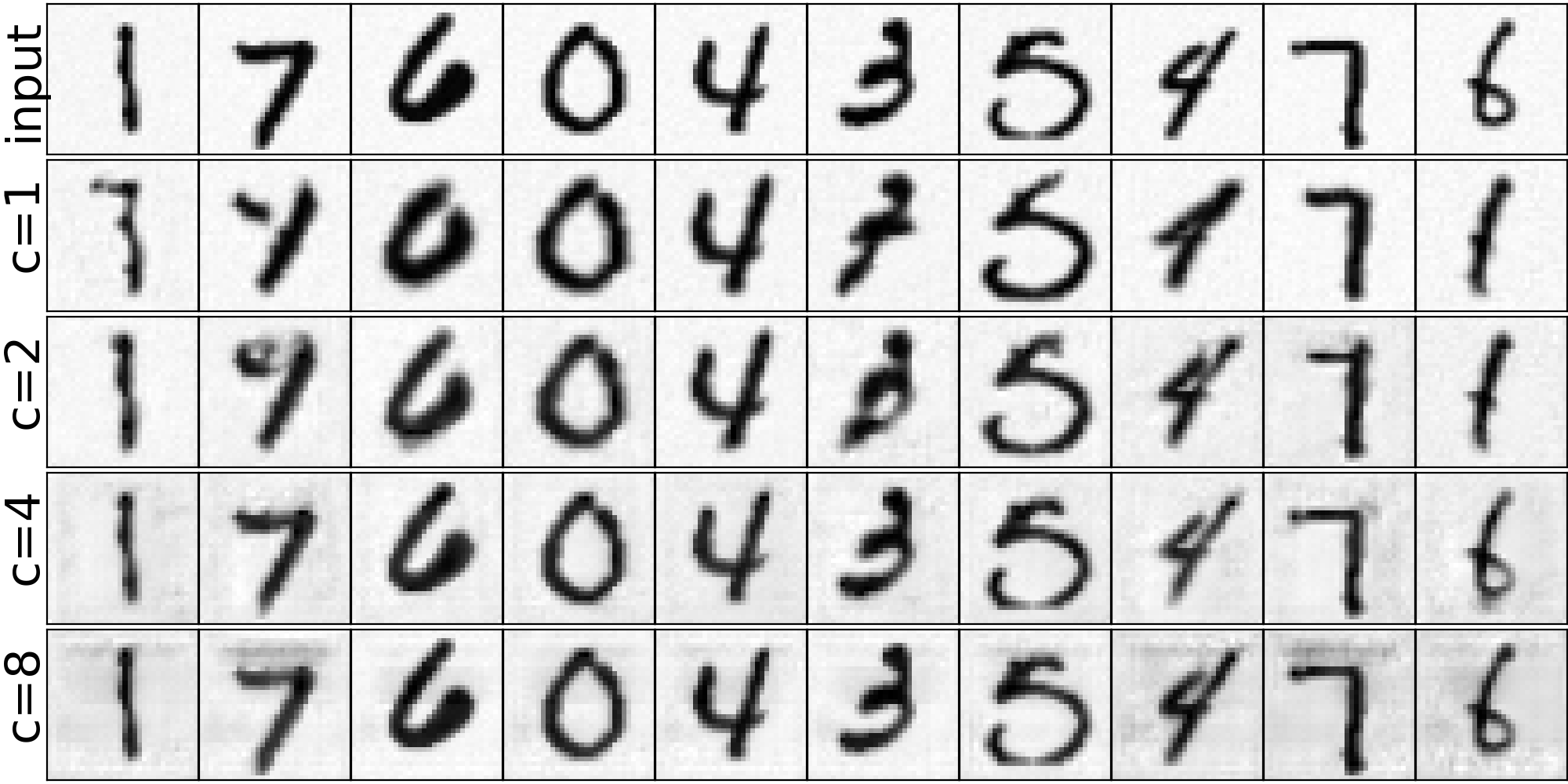}
        \caption{dynamic (ODE integration)}
        \label{fig:emnist-dyn-ode}
    \end{subfigure}
    \caption{Image translation outputs for MNIST $\rightarrow$ EMNIST.}
    \label{emnist-translation}
\end{figure}

In our experiments, we trained and evaluated both static and dynamic models using the corresponding subset-selection frameworks for $c\in \{1, 2, 4, 8 \}$. Similar to flow-based generative models, dynamic subset alignment requires numerical integration; in our experiments, we use ODE integration with an Euler-type numerical integrator with 100 integration steps. However, one-step integration can also be used~\citep{liu2023flow, liu2024instaflow}. Figure~\ref{emnist-translation} shows that, perceptually, one-step integration performs worse than both static and ODE-based generation, but classification accuracies are better. 

The classifier-based evaluation should be interpreted together with the qualitative results. As discussed above, the classifier is trained on the target data, where the digit classes are assigned separate labels $(0${-}$9)$, while all alphabet classes are grouped into a single additional label. Therefore, this evaluation primarily measures whether translated MNIST samples are classified as digit-like EMNIST samples rather than alphabet-like samples. For small $c$, one-step translation can achieve higher accuracy because it only slightly perturbs the source image and often stays within the same classifier decision region, even if the resulting image is visually less aligned with EMNIST. In contrast, the ODE solution follows the learned transport more faithfully, which can cause more label changes for small $c$ and lower classifier accuracy. As $c$ increases, the relevant target subset becomes easier to identify, and the gap between one-step and ODE results becomes small both quantitatively and qualitatively.

\subsection{Positive-Unlabeled Learning}
\label{Positive-Unlabeled Learning}
Positive-unlabeled (PU) learning is a binary classification problem in which only a subset of positive samples is labeled. This labeled subset is then used to train a model that distinguishes positive samples from negative samples within an unlabeled dataset containing both positive and negative examples~\citep{bekker2020learning, kato2018learning, chapel2020partial, riaz2023partial}. Since the sign of the optimal potential function $\varphi_t$ in our framework differs between the selected and unselected subsets, it can be used to distinguish positive samples from unlabeled samples; see Appendix~\ref{appendix: Thresholding for PU{-}Learning and Rejection Sampling}.

We applied both static and dynamic optimal transport methods to PU learning on the 20 UCI datasets~\citep{UCI-datasets}, following the setting of~\citet{teisseyre2025learning}. We used the same 75/25 train-test split for each dataset and the sampled-completely-at-random (SCAR) mechanism to select and label positive points. Dynamic subset alignment employs Algorithm~\ref{neural-dynamic_0}, while static subset alignment employs Algorithm~\ref{neural-static}, described in Appendix~\ref{Static support subset alignment}.

\begin{table}
\setlength{\tabcolsep}{1.3pt}
\renewcommand{\arraystretch}{0.75}
\captionof{table}{Comparison of average balanced accuracies of 20 models trained using dynamic subset alignment methods with PUSB and NTC-MI reported by \citet{teisseyre2025learning}, and our formulation of incomplete extremal optimal transport from \citet{gazdieva2023extremal} (labeled static). The best methods are colored red and second best blue.}
\label{PU-comparison}
\centering 
{\scriptsize 
\begin{tabular}{l|l|l|l|l|c|c}
\multicolumn{1}{c|}{\multirow{2}{*}{\textbf{Dataset}}} &
\multicolumn{1}{c|}{\multirow{2}{*}{\textbf{PUSB}}} &
\multicolumn{1}{c|}{\multirow{2}{*}{\textbf{NTC-MI}}} &
\multicolumn{2}{c|}{\textbf{static}} &
\multicolumn{2}{c}{\textbf{dynamic}} \\
\multicolumn{1}{c|}{} &
\multicolumn{1}{c|}{} &
\multicolumn{1}{c|}{} &
\multicolumn{1}{c|}{sorted} & \multicolumn{1}{c|}{sign} &
\multicolumn{1}{c|}{sorted} & \multicolumn{1}{c}{sign} \\
\hline
Abalone    & 0.544 ± 0.060 & \textcolor{blue}{0.575 ± 0.025} & 0.561 ± 0.029 & 0.503 ± 0.008 & \textcolor{red}{0.578 ± 0.025} & 0.540 ± 0.054 \\
Banknote   & 0.829 ± 0.050 & 0.922 ± 0.019 & 0.883 ± 0.037 & 0.882 ± 0.039 & \textcolor{red}{0.952 ± 0.022} & \textcolor{blue}{0.939 ± 0.018} \\
Breast-w   & 0.766 ± 0.145 & 0.870 ± 0.028 & 0.930 ± 0.028 & \textcolor{red}{0.941 ± 0.027} & 0.927 ± 0.024 & \textcolor{blue}{0.931 ± 0.026} \\
Diabetes   & 0.546 ± 0.042 & \textcolor{red}{0.700 ± 0.039} & \textcolor{blue}{0.635 ± 0.044} & 0.629 ± 0.044 & 0.629 ± 0.043 & 0.630 ± 0.046 \\
Haberman   & 0.513 ± 0.023 & 0.532 ± 0.066 & 0.539 ± 0.066 & \textcolor{blue}{0.540 ± 0.067} & \textcolor{red}{0.544 ± 0.074} & 0.540 ± 0.075 \\
Heart      & 0.527 ± 0.033 & \textcolor{red}{0.757 ± 0.053} & 0.637 ± 0.093 & 0.623 ± 0.089 & \textcolor{blue}{0.656 ± 0.080} & 0.653 ± 0.094 \\
Ionosphere & 0.440 ± 0.085 & 0.755 ± 0.059 & 0.773 ± 0.091 & 0.762 ± 0.088 & \textcolor{red}{0.827 ± 0.066} & \textcolor{blue}{0.807 ± 0.070} \\
Isolet     & 0.793 ± 0.072 & 0.725 ± 0.006 & \textcolor{blue}{0.881 ± 0.028} & \textcolor{red}{0.923 ± 0.030} & 0.731 ± 0.140 & 0.707 ± 0.182 \\
Jm1        & \textcolor{red}{0.628 ± 0.016} & \textcolor{red}{0.628 ± 0.013} & 0.576 ± 0.015 & 0.575 ± 0.010 & \textcolor{blue}{0.593 ± 0.023} & 0.558 ± 0.029 \\
Kc1        & 0.645 ± 0.075 & \textcolor{red}{0.679 ± 0.030} & 0.604 ± 0.036 & 0.607 ± 0.035 & \textcolor{blue}{0.649 ± 0.030} & 0.573 ± 0.032 \\
Madelon    & 0.496 ± 0.030 & 0.519 ± 0.028 & \textcolor{blue}{0.533 ± 0.025} & 0.523 ± 0.015 & \textcolor{red}{0.535 ± 0.018} & 0.517 ± 0.019 \\
Musk       & 0.712 ± 0.036 & 0.767 ± 0.012 & 0.841 ± 0.018 & \textcolor{blue}{0.847 ± 0.018} & \textcolor{red}{0.857 ± 0.021} & \textcolor{blue}{0.847 ± 0.046} \\
Segment    & 0.848 ± 0.074 & 0.803 ± 0.014 & 0.898 ± 0.038 & \textcolor{blue}{0.927 ± 0.031} & 0.911 ± 0.038 & \textcolor{red}{0.936 ± 0.024} \\
Semeion    & 0.569 ± 0.055 & 0.755 ± 0.022 & \textcolor{blue}{0.824 ± 0.044} & \textcolor{red}{0.850 ± 0.067} & 0.603 ± 0.096 & 0.544 ± 0.067 \\
Sonar      & 0.497 ± 0.041 & 0.573 ± 0.057 & 0.561 ± 0.091 & 0.524 ± 0.074 & \textcolor{red}{0.602 ± 0.089} & \textcolor{blue}{0.590 ± 0.101} \\
Spambase   & \textcolor{blue}{0.821 ± 0.031} & \textcolor{red}{0.887 ± 0.014} & 0.786 ± 0.011 & 0.775 ± 0.010 & 0.799 ± 0.017 & 0.776 ± 0.018 \\
Vehicle    & 0.549 ± 0.067 & 0.804 ± 0.042 & \textcolor{blue}{0.806 ± 0.037} & \textcolor{red}{0.823 ± 0.032} & 0.622 ± 0.090 & 0.556 ± 0.077 \\
Waveform   & \textcolor{red}{0.860 ± 0.012} & \textcolor{blue}{0.829 ± 0.012} & 0.795 ± 0.015 & 0.743 ± 0.013 & 0.772 ± 0.0129 & 0.729 ± 0.016 \\
Wdbc       & 0.798 ± 0.155 & 0.801 ± 0.043 & 0.861 ± 0.068 & 0.845 ± 0.063 & \textcolor{red}{0.884 ± 0.050} & \textcolor{blue}{0.872 ± 0.039} \\
Yeast      & 0.517 ± 0.051 & \textcolor{red}{0.657 ± 0.024} & \textcolor{blue}{0.630 ± 0.040} & 0.612 ± 0.049 & 0.627 ± 0.029 & 0.533 ± 0.046 \\\hline
\end{tabular}%
}
\end{table}

Neural networks consisting of five MLP layers with appropriate input and output dimensions and Swish activation functions are used to train both the static and dynamic models. The learning rates are held constant during training for both models. Architecture and parameter details for each model are provided in Appendix~\ref{app:pu-learning-models}. We trained 20 different models for each dataset using different train-test splits. In total, we trained 400 models for static subset alignment and 400 models for dynamic subset alignment. We adopted alternative sign-based and value-based label-assignment strategies for the unlabeled dataset. The performance of our approaches, measured in terms of balanced accuracy, along with the top-performing baselines PUSB~\citep{kato2018learning} and NTC-MI~\citep{teisseyre2025learning}, is reported in Table~\ref{PU-comparison}. In all PU learning experiments, $c$ is fixed as $c=\frac{1}{\pi}$, where $\pi$ denotes the known prior proportion of positively labeled points in the dataset. Details are discussed in Appendix~\ref{app:pu-learning-models}.

\textcolor{black}{Since OT methods rely on the assumption that data points in a dataset can be geometrically related through a metric, we used t-SNE~\citep{vandermaaten08a} embeddings with perplexity $30$ for all datasets in Table~\ref{PU-comparison} to evaluate the spatial separability of the positive and negative classes. We observed that our methods perform better or comparably to competing methods when the positive class $(1)$ and negative class $(0)$ have spatially separable t-SNE embeddings as shown by the t-SNE embeddings are provided in Appendix~\ref{app:tsne-uci}.}  
\subsection{FFHQ Image Translation}
We also apply our proposed approaches to the unpaired image translation problem. We followed the experimental setup of~\cite{gazdieva2024light}, where the FFHQ dataset, embedded in the latent space of the Adversarial Latent Autoencoder (ALAE)~\citep{pidhorskyi2020adversarial}, is divided either by gender (man or woman) or by age, as two orthogonal labels. \begin{wraptable}[6]{r}{3.3cm}    
    \caption{Division of FFHQ train images.}
    \label{FFHQ_data}
    \setlength{\tabcolsep}{3pt}
    \renewcommand{\arraystretch}{0.9}
    \begin{tabular}{c|c|c}
    \textbf{Class} & Man & Woman \\
    \hline
    Young & 15K  & 23K \\
    \hline
    Old   & 7K   & 3.5K
    \end{tabular}
\end{wraptable}
Table~\ref{FFHQ_data}, adapted from~\cite{gazdieva2024light}, shows the number of images for each class; images with ages less than 16 are ignored, ages between 16 and 43 are labeled young, and the remainder are labeled old. Given these classes, the task is to learn to map a source distribution to a target distribution. There are four cases: young to old, old to young, man to woman, and woman to man. To evaluate the translation process, two classifiers pretrained in the ALAE latent space are used; one classifier is trained to classify young versus old, and the other to distinguish man versus woman. The target accuracy quantifies the proportion of translated images that lie within the target-class boundary. The source accuracy quantifies whether the translated images retain the orthogonal label. For example, with young$\rightarrow$old, the source accuracy is whether the ``aged'' image of a young source image retains the same gender.

Implementation details for both static and dynamic subset selection to the FFHQ dataset are given in Appendix~\ref{app:ffhq}. Between young$\rightarrow$old, old$\rightarrow$young, man$\rightarrow$woman and woman$\rightarrow$man, it was observed that larger values of $c$ tend to preserve the source accuracy but often have lower target accuracy. This can be related to the fact that for larger values of $c$, it takes more training steps to achieve the optimal subset selection.
\begin{figure}[htb]
    \centering
    \includegraphics[width=0.7\linewidth]{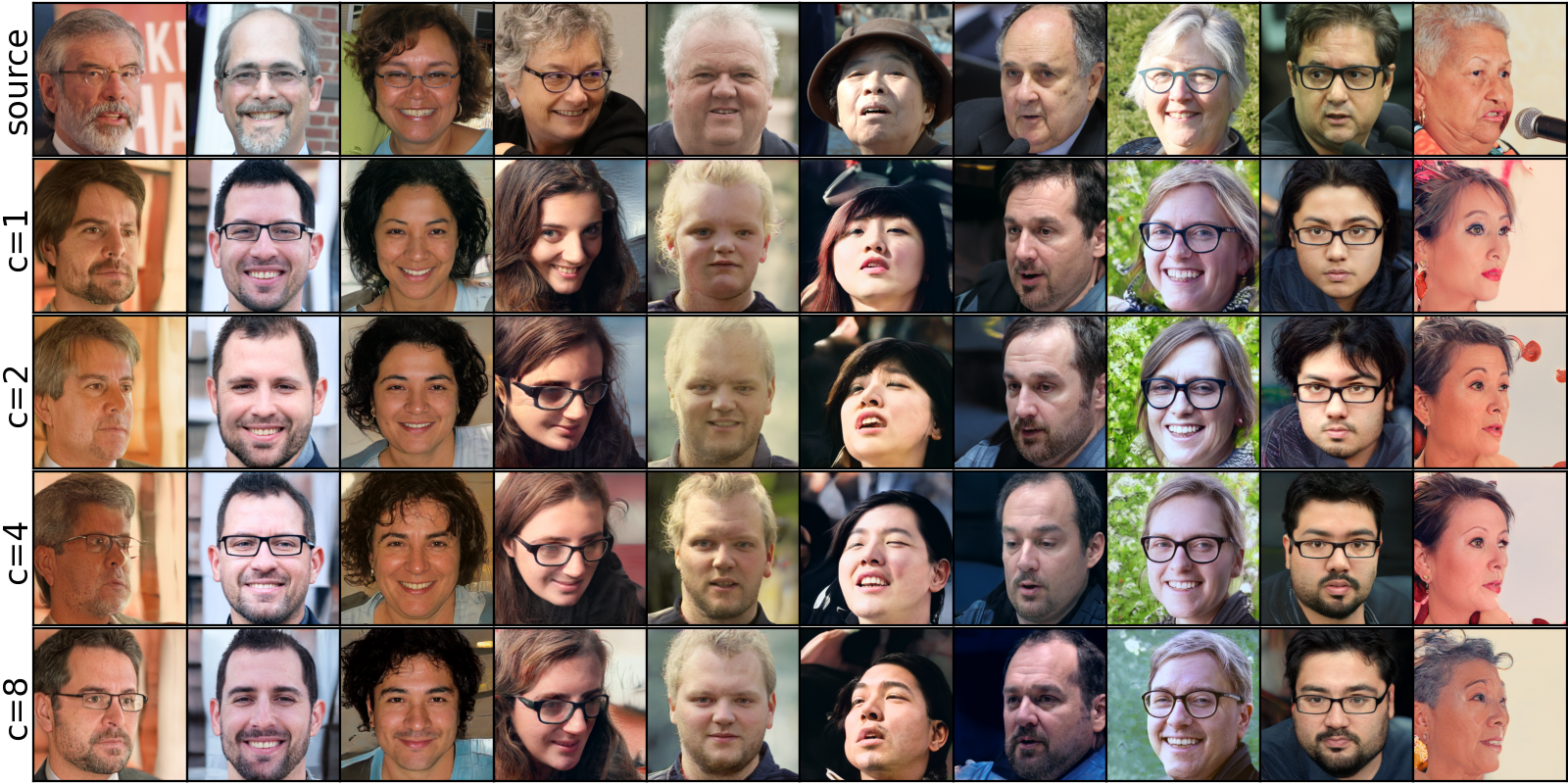}
    \caption{FFHQ old$\rightarrow$young  using dynamic subset alignment with Euler integration (100 steps).}
    \label{old-to-young}
\end{figure}
Example translated images for old$\rightarrow$young translation with dynamic subset alignment are shown in Figure~\ref{old-to-young}, with other cases provided in Appendix~\ref{app:ffhq_more}.

\begin{table*}[h]
\centering
\scriptsize 
\setlength{\tabcolsep}{0.9pt} 
\renewcommand{\arraystretch}{1.0} 
\caption{Target and source accuracy (\%) for different domain translations on the FFHQ dataset. Dynamic subset selection is evaluated using Euler integration with 100 steps.} 
\begin{tabular}{ll | cc|cc|cc|cc|c|c|c|c|c|c}
& \textbf{}               
& \multicolumn{2}{c|}{\textbf{c=1}} 
& \multicolumn{2}{c|}{\textbf{c=2}} 
& \multicolumn{2}{c|}{\textbf{c=4}} 
& \multicolumn{2}{c|}{\textbf{c=8}} 
& \multicolumn{1}{c|}{\textbf{UOT}} 
& \textbf{UOT} 
& \textbf{U}  
& \textbf{UOTM}
& \textbf{UOTM}
& \textbf{UOTM}
\\
Task  & Accuracy
& static & dynamic                  
& static & dynamic                  
& static & dynamic                  
& static & dynamic                  
& \textbf{-SD}                                    
& \textbf{-GAN}                
& \textbf{-LOT}                
& \textbf{KL-KL}
& \textbf{$\chi^2$-KL}
& \textbf{lin.--$\chi^2$}
\\ 
\hline

Young$\rightarrow$Old 
& Target 
& 84.09 & 83.45 
& 70.47 & 79.23 
& 47.63 & 74.51 
& 27.07 & 73.93 
& 87.33 & 84.25 & 81.78 
& 71.06 & 72.68 & 80.78
\\
& Class 
& 70.43 & 71.55 
& 91.41 & 84.31 
& 95.47 & 90.03 
& 96.06 & 90.30 
& 45.71 & 73.85 & 84.49 
& 90.17 & 88.68 & 78.77
\\ 
\addlinespace

Old$\rightarrow$Young
& Target 
& 96.06 & 93.36 
& 92.55 & 86.65 
& 85.04 & 85.03 
& 72.93 & 83.33 
& 97.39 & 95.88 & 87.79 
& 88.67 & 89.70 & 94.19
\\
& Class 
& 71.77 & 71.92 
& 89.46 & 87.69 
& 94.43 & 89.84 
& 95.45 & 90.88 
& 49.30 & 74.74 & 89.48 
& 89.64 & 88.19 & 80.68
\\ 
\addlinespace

Man$\rightarrow$Woman
& Target 
& 96.11 & 94.53 
& 92.18 & 91.74 
& 85.66 & 90.96 
& 77.55 & 90.29 
& 98.16 & 97.38 & 90.23 
& 90.63 & 91.19 & 94.51
\\
& Class 
& 83.68 & 83.64 
& 87.45 & 87.43 
& 90.05 & 88.52 
& 91.27 & 89.11 
& 75.50 & 84.04 & 90.30 
& 89.52 & 88.83 & 85.31
\\ 
\addlinespace

Woman$\rightarrow$Man 
& Target 
& 93.34 & 92.26 
& 87.32 & 88.09 
& 78.32 & 84.28 
& 64.86 & 81.89 
& 94.96 & 92.91 & 88.59 
& 83.85 & 84.10 & 91.41
\\
& Class 
& 82.39 & 82.68 
& 92.33 & 91.51 
& 93.89 & 92.59 
& 94.35 & 93.22 
& 72.03 & 84.56 & 89.66 
& 91.79 & 90.85 & 87.17\\
\hline 
\end{tabular}
\label{ffhq_class_table}
\end{table*}
\begin{figure}[h!]
    \centering
    \begin{subfigure}[t]{0.2475\textwidth}
        \centering
        \includegraphics[width=\linewidth,clip=true,trim={0 0 0 1.125cm}]{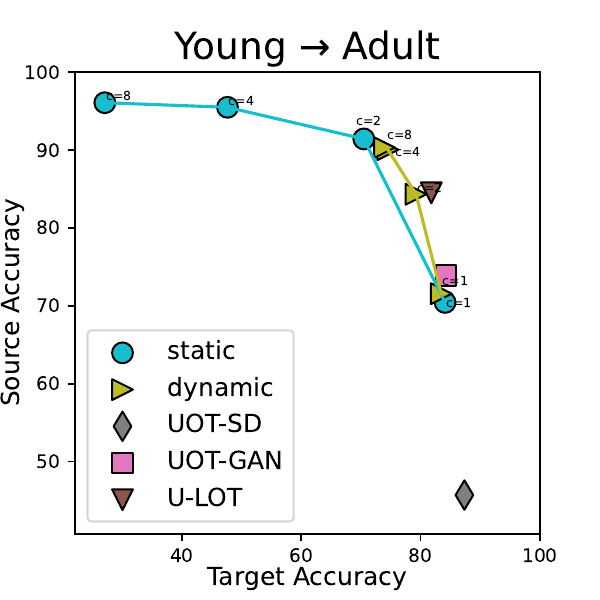}
        \caption{Young$\rightarrow$Old}
        \label{fig:pareto-ya}
    \end{subfigure}\hspace{0mm}%
    \begin{subfigure}[t]{0.2475\textwidth}
        \centering
        \includegraphics[width=\linewidth,clip=true,trim={0 0 0 1.125cm}]{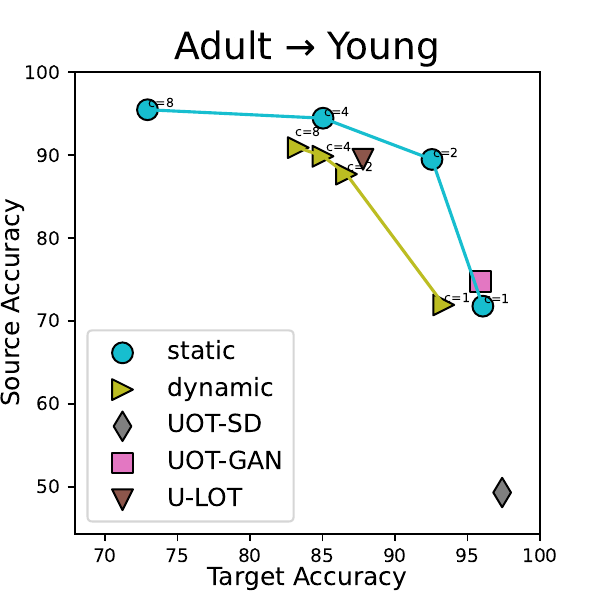}
        \caption{Old$\rightarrow$Young}
        \label{fig:pareto-ay}
    \end{subfigure}\hspace{0mm}%
    \begin{subfigure}[t]{0.2475\textwidth}
        \centering
        \includegraphics[width=\linewidth,clip=true,trim={0 0 0 1.125cm}]{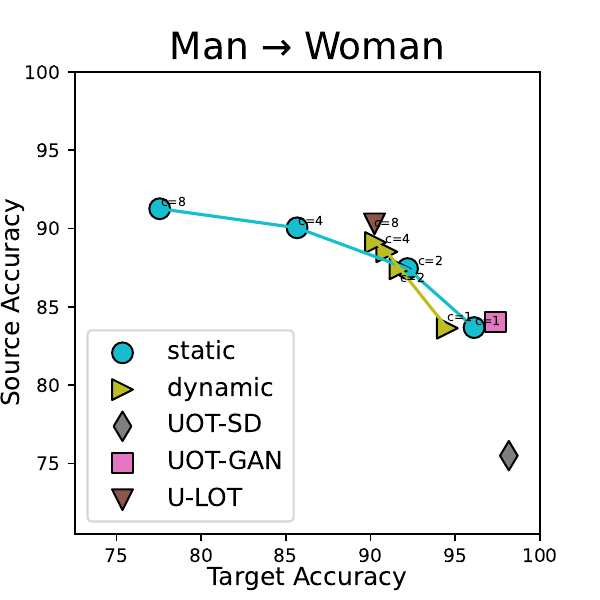}
        \caption{Man$\rightarrow$Woman}
        \label{fig:pareto-mw}
    \end{subfigure}\hspace{0mm}%
    \begin{subfigure}[t]{0.2475\textwidth}
        \centering
        \includegraphics[width=\linewidth,clip=true,trim={0 0 0 1.125cm}]{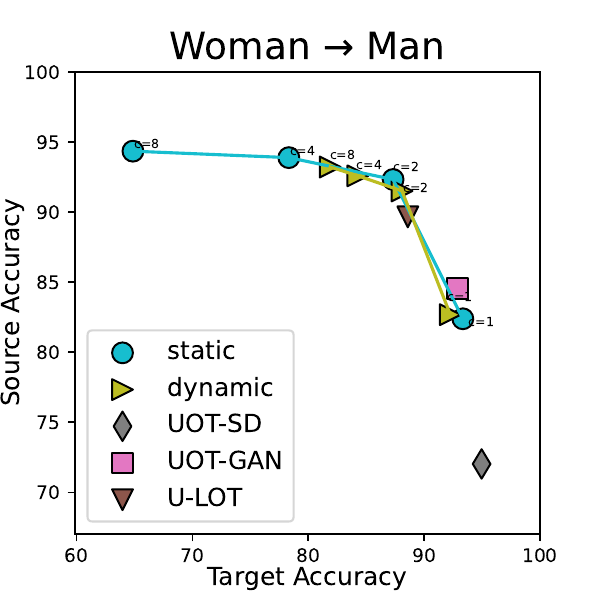}
        \caption{Woman$\rightarrow$Man}
        \label{fig:pareto-wm}
    \end{subfigure} 
    \caption{Accuracy curves for $c\in\{1,2,4,8\}$, in comparison to results from LOT~ \citep{gazdieva2024light}, UOT-GAN~\citep{yang2018scalable}, UOT-SD and UOTM~\citep{choi2023generative, choi2024analyzing}.}
    \label{mnist-emnist-results-comparisons}
\end{figure}
We compare with LOT~\cite{gazdieva2024light}, UOT-GAN~\cite{yang2018scalable}, UOTM, and UOT-SD~\cite{choi2023generative, choi2024analyzing}. Results in  Table~\ref{ffhq_class_table} and Figure~\ref{mnist-emnist-results-comparisons}, using accuracy values for UOT-SD, UOT-GAN, and U-LOT reported by~\citet{gazdieva2024light}, whereas the implementation for UOTM is discussed in Appendix~\ref{Static support subset alignment}. Methods with better target accuracy are worse for source/class accuracy.

\subsection{Melanoma Dataset}
The subset-alignment framework is primarily developed for aligning a source distribution $\mu$ to an unknown distribution $\tilde{\nu}$ whose support is contained within the support of a target distribution $\nu$. Following~\citet{neklyudov24a} and~\citet{pariset2023unbalanced}, we adapt our dynamic subset-alignment method to the melanoma treatment data of~\citet{bunne2023learning}, using the preprocessed 50-dimensional representation. Unlike our other experiments, the melanoma setting involves three observed marginals at $t\in\{8\text{hr},24\text{hr},48\text{hr}\}$ and we must match the intermediate marginal exactly. We therefore order the populations in reverse chronological order, $\mu_0$ (48h), $\mu_1$ (24h), and $\mu_2$ (8h), and train a multi-marginal interpolant that preserves $\mu_0$ and $\mu_1$ while leaving the terminal marginal (here, the 8h population) unconstrained so that sub-setting acts on $\mu_2$. After training, we select cells $\rvx\sim\mu_2$ with $\varphi_{t_2}(\rvx)\le 0$ and transport the selected subset back to $t_0$ using the learned velocity field $\nabla\varphi_t$.  (We use normalized times $t_0=0$, $t_1=0.6$, $t_2=1$. )  For evaluation, we integrate the ODE with explicit Euler (step size $10^{-3}$) and report the best checkpoint based on validation MMD. The training procedure, neural parameterizations and experimental parameters for the Melanoma experiment are given in the Appendix~\ref{app: Melanoma Experiment}.  As shown in Table~\ref{tab:melanoma_results}, subset alignment improves over previous neural Schr\"odinger-bridge baselines~\citep{chen2022likelihood,pariset2023unbalanced}, while remaining competitive with WLF~\citep{neklyudov24a}.

\begin{table}[h]
\caption{Results for the Melanoma dataset~\citep[Table~1]{pariset2023unbalanced} with three marginals and train/test splits. MMD and $W_2$ are measured between $\nu_{t=48}$ and $\rho_{t=48}$ obtained by running dynamics from $\mu_{t=8}$. For subset alignment, dynamics are run from the selected subset of $\mu_{t=8}$ (cells with $\varphi_{t=8}\leq 0$; number of selected test cells in parentheses), and results are reported at the iteration with the lowest MMD on a validation set; note that the validation cells are a subset of the training cells.}
\label{tab:melanoma_results}
\setlength{\tabcolsep}{4pt}
{\scriptsize
\begin{tabular}{lcc @{\hspace{2em}} lcc}
\toprule
\textbf{Model} & \textbf{MMD} & \textbf{$W_2$} & \textbf{Model} & \textbf{MMD} & \textbf{$W_2$} \\
\midrule
SB-FBSDE \citep{chen2022likelihood} & $1.86\mathrm{e}{-2}$ & $6.23$ & Subset Alignment ($c=1$, 476/490) & $5.21\mathrm{e}{-3}$ & $4.83$ \\
uDSB (no growth) \citep{pariset2023unbalanced} & $1.86\mathrm{e}{-2}$ & $6.27$ & Subset Alignment ($c=2.0$, 247/490) & $6.10\mathrm{e}{-3}$ & $\mathbf{4.68}$ \\
uDSB (w/growth) \citep{pariset2023unbalanced} & $1.75\mathrm{e}{-2}$ & $6.11$ & Subset Alignment ($c=4.0$, 148/490) & $9.24\mathrm{e}{-3}$ & $5.40$ \\
WLF-OT no growth \citep{neklyudov24a} & $\mathbf{5.04\mathrm{e}{-3}}$ & $5.20$ & Subset Alignment ($c=8.0$, 103/490) & $1.48\mathrm{e}{-2}$ & $6.60$ \\
WLF-UOT $\lambda=0.1$ \citep{neklyudov24a} & $9.16\mathrm{e}{-3}$ & $5.01$ & & & \\
\bottomrule
\vspace{-10pt}
\end{tabular}
}\end{table}

\subsection{Celeb-A to Anime Domain Translation}
\begin{wrapfigure}[11]{r}{0.34\linewidth}
\includegraphics[width=0.33\textwidth]{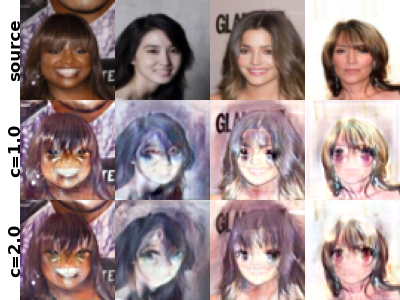}
    \caption{CelebA$\rightarrow$Anime.}
    \label{fig:celeba2anime}
\end{wrapfigure}

For domain translation between the CelebA~\citep{liu2015faceattributes} and Anime~\citep{chao2019-online} datasets, we use highly trimmed variants of the DDPM++ architecture~\citep{song2021scorebased} to parameterize both $Q_t$ and $\varphi_t$ so that the models fit within GPU memory. Because our subset-alignment framework maps the source distribution to a \emph{subset} of the target distribution, FID is not an appropriate evaluation metric for this setting. We therefore rely on qualitative visual inspection. In our experiments, dynamic subset alignment successfully translates the source distribution toward the target domain; however, larger values of $c$ generally require longer training to achieve comparable visual quality and stability. A sample translation is shown in Figure~\ref{fig:celeba2anime}.

\section{Limitations}
Practically, one point of concern is that the sample estimators of Wasserstein distances suffer from the curse of dimensionality~\citep{Weed2019, fournier2015rate}. This issue can be alleviated to some extent by employing entropic regularization~\citep{genevay2019sample, feydy2019interpolating}, which, in the dynamic setting, is intimately connected to Schr\"odinger bridges~\citep{peluchetti2023diffusion, bortoli2024schrodinger}.

Recently, entropically regularized UOT has been studied for modeling birth or death processes in population dynamics~\citep{pariset2023unbalanced, neklyudov2023action}. Our approach can also be applied to model death processes in settings where there is a canonical relationship between temporally ordered populations. In such cases, we treat $\mu$ as the final population of survivors and $\nu$ as the initial population. The learned mapping then relates the survivors to a subset of the initial population. Moreover, the sign of the potential can be used to distinguish members of the initial population that are likely to survive from those that are not. We note that, in this and other applications, the choice of $c$ is critical. While $c$ has a clear interpretation as a constraint on the density ratio, automatic selection of $c$ based on the resulting transport cost may be possible, as previously studied for partial optimal transport in the discrete setting~\citep{phatak2022computing}. One consolidated approach would be to sample $c$ from a prescribed range and use multi-task learning to optimize networks across varying values of $c$. From an implementation perspective, this can be achieved by using scalar embeddings of $c$, analogous to the time-variable embeddings used in dynamic networks.

 \begin{wraptable}{r}{0.50\textwidth}
\centering
\caption{Average evaluation time per sample for FFHQ domain translation tasks from Table~4. Evaluation is performed with batch size 128.}
\label{tab:ffhq_eval_time}
\scriptsize
\setlength{\tabcolsep}{2pt}
\renewcommand{\arraystretch}{0.9}
\begin{tabular}{lcc}
\hline
\textbf{Metric} & \textbf{One-Step} & \textbf{ODE} \\
\hline
Time/sample (s) & $1.44{\times}10^{-4}$ & $2.84{\times}10^{-3}$ \\
\hline
\end{tabular}
\end{wraptable}

Another key limitation of dynamic models, including frameworks similar to ours, is that their iterative generation process requires more computational resources and longer inference time~\citep{liu2023flow}. For example, Table~\ref{tab:ffhq_eval_time} reports the average per-sample generation time for FFHQ image translation. These results show that, in this setting, one-step generation is at least an order of magnitude faster than ODE integration. Considerable recent effort has been devoted to improving the generation efficiency of dynamic models~\citep{liu2024instaflow, geng2026mean}.

We note that although this work focuses on relaxing the target distribution, the range-divergence framework could also be used to relax the source distribution. A fully relaxed version, in which both source and target marginals are relaxed, may be applicable to other classes of problems.
\section{Conclusion}
Our neural optimal transport approach with subset selection achieves subset alignment by translating between two distributions while reweighting and selecting a subset of the target support. Our results on image-translation tasks across two datasets and 20 tabular PU-learning tasks show that a meaningful subset can be learned simultaneously with a Monge map. Unlike previous work, our dynamic formulation allows the terminal distribution to differ from the original target marginal, thereby inducing flows toward the nearest target subset while maintaining the prescribed density-ratio constraint.

\subsection*{AI use statement}
We used generative AI tools primarily for copy-editing (improving grammar, clarity, and flow). We did not use generative AI for code development, except in the Melanoma dataset section, where an AI tool helped identify a bug. All AI-assisted outputs were reviewed and verified by the authors. We take full responsibility for the final content of this work, including all text, claims, and artifacts.

\subsection*{Reproducibility statement}
To facilitate full reproducibility, we provide detailed descriptions of all experiments in the appendix. The main body summarizes the experimental setup and procedures for each experiment, while the appendix includes the complete specifications needed to replicate our results.


\appendix
\section{Preliminaries and Problem Formulation}
\label{prelim}
Gaspard Monge proposed the original idea of optimal transport as mathematical model for the problem of minimum-cost transportation of dirt from a source location to a destination~\citep{monge1781memoire}.  In more modern parlance, given probability measures, $\mu$ defined on compact set $\mathcal{X} \subseteq \mathbb{R}^{d}$, $\nu$ defined on compact set $\mathcal{Y} \subseteq \mathbb{R}^{d}$, and the bounded uniformly continuous cost $\mathtt{c}(\cdot, \cdot):\mathcal{X} \times \mathcal{Y} \rightarrow \mathbb{R}$,  the Monge formulation of optimal transport is stated as
\begin{equation} \label{monge0}
    \begin{aligned}
         \mathcal{D}_{\text{Monge}}(\mu, \nu) & = \underset{T \in \mathcal{J}\left(\mathcal{X}, \mathcal{Y}\right)}{\inf} \int_{\mathcal{X}} \mathtt{c}\left (\rvx, T(\rvx) \right) \mu(\rvx)  d\rvx 
        & \quad \text{s.t.} & \quad \ T_{\#}\mu = \nu
    \end{aligned}
\end{equation}
where the set $\mathcal{J}\left(\mathcal{X}, \mathcal{Y} \right)$ denotes the set of measurable maps between $\mathcal{X}$ and $\mathcal{Y}$. Monge formulation of the optimal transport problem requires that the transport map of $T$ to be a deterministic function. To satisfy the constraint in the Monge problem~\ref{monge0}, the transport map $T$ must cover $\nu$ up to some $\nu${-}null sets. Usually, the cost $\mathtt{c}$ is non-linearly dependent on the transportation map~$T$, making the problem~\ref{monge0} very cumbersome and very difficult to solve~\citep{santambrogio2015optimal, villani2009optimal}.
Kantorovich reformulated the Monge problem by relaxing the constraint that supports of $\mu$ and $\nu$ should be related to each other by a functional relation $T$ \citep{kantorovich1942translocation}. Instead, he allowed the marginals $\mu$ and $\nu$ to be related to each other by a joint measure. Kantorovich's reformulation of the problem is a linear program, and its solution exists for all convex lower-semi-continuous costs~\citep{santambrogio2015optimal, figalli2021invitation}. The Kantorovich problem is
\begin{equation} \label{Kantorovich0}
    \begin{aligned}
         \mathcal{W}(\mu, \nu) = \underset{\pi}{\inf} & \int_{\mathcal{X \times Y}} \mathtt{c}\left (\rvx, \rvy \right) \pi(\rvx, \rvy) d\rvx d\rvy &
        \text{s.t.} &  \int_{\mathcal{Y}} \pi(\rvx,\rvy) d\rvy = \mu(\rvx), \ \ \int_{\mathcal{X}} \pi(\rvx,\rvy) d\rvx = \nu(\rvy), 
    \end{aligned}
\end{equation}
where it can be observed that integrals $\int_{\mathcal{X}}$ and $\int_\mathcal{Y}$ marginalize with respect to spaces $\mathcal{Y}$ and $\mathcal{X}$ respectively. Therefore, $\pi$ must be the joint measure between $\mu$ and $\nu$ defined on the product space $\mathcal{X} \times \mathcal{Y}$, {i.e.}  $\pi \in \mathcal{P} \left(\mathcal{X} \times \mathcal{Y} \right)$. In other words, constraints in the Kantorovich problem ensure that every feasible solution $\pi$ must be a joint distribution of $\mu$ and~$\nu$. 
While in general, the Kantorovich problem is much easier to solve in comparison to the Monge problem, there are conditions of practical importance where one can employ the solutions of the Kantorovich problem to obtain the solution of the Monge problem. Those conditions are more clearly discussed in terms of the dual form of the  Kantorovich problem~\citep{santambrogio2015optimal} given as,
\begin{equation} \label{Kantorovich_dual0}
    \begin{aligned}
         \mathcal{W}(\mu, \nu) = \underset{f, g }{\sup} & \int_{\mathcal{X}}f(\rvx) \mu(\rvx) d\rvx+ \int_{\mathcal{Y}}g(\rvy) \mu(\rvy) d\rvy & 
        \text{s.t.} & \quad f(\rvx) + g(\rvy) \leq \mathtt{c}(\rvx, \rvy).
    \end{aligned}
\end{equation}
The functions $f(\rvx)$ and $g(\rvy)$ are called Kantorovich potentials. By defining $\mathtt{c}$-conjugate (also called $\mathtt{c}$-transform) of $f(\rvx)$, and $\bar{\mathtt{c}}$-conjugate (also called $\bar{\mathtt{c}}$-transform) of $g(\rvy)$ as 
\begin{align}
    f^{\mathtt{c}}(\rvy) &= \inf_{\rvx \in \mathcal{X}} \mathtt{c} (\rvx, \rvy) - f(\rvx), \\
    g^{\bar{\mathtt{c}}}(\rvx) &= \inf_{\rvx \in \mathcal{X}} \mathtt{c} (\rvx, \rvy) - g(\rvy),
\end{align}
Using ${\mathtt{c}}$ and $\bar{\mathtt{c}}$ conjugates, Kantorovich problem is expressed as
\begin{align}
\mathcal{W}(\mu, \nu) & = \underset{f}{\sup}\int_{\mathcal{X}}f(\rvx) \mu(\rvx) d\rvx + \int_{\mathcal{Y}}f^{\mathtt{c}}(\rvy) \nu(\rvy) d\rvy \\
\quad &= \underset{g}{\sup} \int_{\mathcal{X}}g^{\bar{\mathtt{c}}}(\rvx) \mu(\rvx) d\rvx + \int_{\mathcal{Y}}g(\rvy) \nu(\rvy) d\rvy
\end{align}
Under very general conditions, one can relate the cost $\mathtt{c}$ with the support of optimal coupling solution $\pi^{\star}$ and optimal Kantorovich potentials $f^{\star}(\rvx)$ and ${f^{\mathtt{c}}}^{\star}(\rvy)$~\cite[Theorem 1.37]{santambrogio2015optimal} by
\begin{equation}
\textrm{supp}(\pi^{\star}) \subset \{ (\rvx, \rvy) \in \mathcal{X} \times \mathcal{Y} : f^{\star}(x) + {f^{\star}}^{\mathtt{c}}(\rvy) = \mathtt{c} (\rvx, \rvy)\}
\end{equation}
In discrete domains, above result is equivalent to Karush-Kuhn-Tucker (KKT) conditions for optimality. One can further relate the optimal solutions of Monge and Kantorovich problems using a landmark result by~\cite{gangbo1996geometry}~\citep[Theorem 2.7.1]{figalli2021invitation}, which states that 
there exists an optimal Kantorovich coupling of the form $\pi^{\star} = (\textrm{Id} \times T^{\star} )_{\#}\mu$, where $T^{\star}$ is Monge map satisfying
\begin{equation}
    \nabla_{\rvx}\mathtt{c}(\rvx, T^{\star}(\rvx)) + \nabla f^{\star}(\rvx) = 0,
\end{equation}
if the following conditions are satisfied
\begin{itemize}
    \item $\mu$ is absolutely continuous,
    \item $\forall \ \rvy \in \mathcal{Y}$ map $ x \mapsto \mathtt{c}(\rvx, \rvy)$ is differentiable~$\forall \ \rvx \in \mathcal{X}$,
    \item  $\forall \ \rvx \in \mathcal{X}$ the gradient map $ y \mapsto \nabla_{\rvx} \mathtt{c}(\rvx, \rvy)$ is injective $\forall \ \rvy \in \mathcal{Y}$,
     \item  and, the gradient $ \nabla_{\rvx}\mathtt{c}(\rvx, \rvy) $ satisfies the local Lipschitz condition $\Vert \nabla_{\rvx} \mathtt{c}(\rvx, \rvy) \Vert \leq C_{r}$ for all $\rvx \in \mathcal{B}_{r}$ where $\mathcal{B}_{r}$ is a ball of radius $r$ around $\rvx$.
\end{itemize}
When the cost can be written as $\mathtt{c}(\rvx, \rvy) = h(\rvx-\rvy)$, where $h$ is strictly convex and translation-invariant function, one can further relate the Monge mapping with optimal dual potential by \citep[Theorem 1.17]{santambrogio2015optimal}
\begin{equation}
    T^{\star}(\rvx) = \rvx - \nabla h^{*} \circ \nabla f^{\star}(\rvx),
\end{equation}
where $h^{*}$ is the Legendre-Fenchel conjugate of $h$ given by $h^{*}(\rvy) = \underset{x\in \mathcal{X}}{\sup} \{ \langle \rvy, \rvx\rangle - h(\rvx) \}.$
In the result above, when $h(\rvx-\rvy)=\frac{1}{2}\Vert \rvx - \rvy \Vert_{2}^{2}$, one obtains the result of the celebrated Brenier theorem of optimal transport for squared-Euclidean costs with $T^{\star}(\rvx) = \nabla f^{\star}(\rvx)$, where $f^{\star}(\rvx)$ is convex~\citep{brenier1991polar}~\citep[Theorem 2.5.10]{figalli2021invitation}. The Brenier theorem on optimal transport differs from another important theorem on polar factorization~\citep{brenier1991polar} stating that under very general conditions a square-integrable vector field~$\rvv$ can be decomposed into the composition of the gradient of a unique convex function $\xi$ and a unique measure-preserving map~$\rvu$, \text{i.e.}~$\rvv(\rvx) = \nabla \xi \circ \rvu(\rvx)$.
Before the discussion on dynamic formulation of the problem, we would like to point out that much of the recent work on static neural optimal transport relies on the above results.

Benamou and Brenier formulated the Wasserstein distance with squared Euclidean cost as the expected kinetic energy minimization problem under the assumption that both the source~$\mu$ and target~$\nu$ distributions have finite second moments~\citep{benamou2000computational, santambrogio2015optimal, figalli2021invitation}. Assuming that supports of both source and target distributions lie in a convex set~$\Omega\ \subseteq \mathbb{R}^{d}$, whose normal at the boundary is given by~$\mathbf{n}:\partial\Omega \rightarrow \mathbb{R}^{d}$. For a bounded and smooth velocity{-}field $\rvv_{t}(\rvx):[0 ,1]\times \Omega \rightarrow \mathbb{R}^{d}$ such that $\langle \rvv_{t}(\rvx), \mathbf{n} \rangle|_{\partial\Omega} = 0$, the flow corresponding to the velocity{-}field $\rvv_{t}$ is given by
\begin{equation} \label{d-flow1}
    \begin{aligned}
        \frac{d}{d t} \Phi_{t}(\rvx_{t}) = \rvv_{t}(\Phi_{t}(\rvx_{t})), \ \, & \, \Phi_{0}(\rvx_{0}) = \rvx_{0}.
    \end{aligned}
\end{equation}
Considering that there also exists a probability path $\rho_{t}(\rvx):[0,1] \times \Omega \rightarrow \mathbb{R}_{+}$ corresponding to the flow $\Phi_{t}(\rvx)$ such that $\rho_{t}(\rvx) = {\Phi_{t}}_{\#} \rho_{0}(\rvx)$,  Benamou-Brenier formulation of optimal transport is 
\begin{equation}
    \begin{aligned} \label{benamou0_1}
    \underset{\rho_{t}, \rvv_{t}}{\inf} & \int_{0}^{1}\int_{\Omega} \frac{\Vert \rvv_{t}(\rvx) \Vert^{2}}{2} \,\rho_{t}(\rvx) d\rvx \, dt  \\
    \text{s.t.} & \quad  \frac{\partial}{\partial_{t}}\rho_{t}(\rvx) + \textrm{div}{\left(\rho_{t}(\rvx) \rvv_{t}(\rvx)\right)} = 0, 
    \quad \rho_0(\rvx) = \mu(\rvx),  \quad \rho_1(\rvx) = \nu(\rvx), \quad 
\end{aligned}
\end{equation}
where, $\textrm{div}(\cdot)$ denotes the divergence operator mapping scalar or vector fields to scalar for the field $z_{t}(\rvx) $ by~$\textrm{div}(z_{t}(\rvx)) = \sum_{i}\frac{\partial }{\partial x_{i} } z_{t}(\rvx)$. The optimal flow $\Phi_{t}^{\star}$ is related to Monge mapping $T^{\star}$ by displacement interpolation~\citep{mccann1997convexity},
\begin{equation}
    \Phi_{t}^{\star} = (1-t)\text{Id} + tT^{\star}.
    \label{McCan}
\end{equation}
It is important to mention that the Benamou{-}Brenier formulation can be extended to Wasserstein-$p$ distances, for~$p>1$,  under the assumption that both source and target distributions have finite $p$th moments~\citep[Chapters 5 \& 6]{santambrogio2015optimal}. 

\subsection{Static support subset alignment}
\label{Static support subset alignment}
Our formulation of static support subset alignment with optimal transport is derived from relaxed optimal transport problem where the constraint on the first marginal of the joint density $\pi$ is maintained, while the second marginal $\int_\mathcal{X}\pi(\rvx,\rvy)d\rvx= \tilde{\nu}(\rvy)$ is allowed to vary from within a range $[0,c]$ of the target density~$\nu$, such that $0\le \frac{\tilde{\nu}(\rvy)}{\nu(\rvy)}\le c$. The density $\tilde{\nu}$ can be interpreted as a reweighted target density~$\tilde{\nu}(\rvy)=\omega(\rvy)\nu(\rvy), \quad 0\le \omega(\rvy)\le c$, where portions of the support can be up-weighted while others are down-weighted or removed. The relaxed constraint is equivalent to a case of the partial optimal transport relaxations using~$f$-divergences introduced by~\citet{sejourne2023unbalanced},
\begin{equation}
\label{static0}
    \begin{aligned}
        \inf_{\pi} & \int_{\mathcal{X}\times \mathcal{Y}} \mathtt{c}(\rvx,\rvy) \pi(\rvx,\rvy) d\rvx d\rvy  + \mathcal{D}_{\imath_{[a,b]}}(\tilde{{\nu}} \Vert {\nu}) &
        \textrm{s.t.} & \int_{\mathcal{Y}} \pi(\rvx,\rvy) d\rvy = \mu(\rvx), 
    \end{aligned}
\end{equation}
where $\mathcal{D}_{\imath_{[a,b]}}$ is the range divergence with $\imath_{[a,b]}$ being the convex indicator function,
\begin{equation}
    \begin{aligned}
        {\imath_{[a,b]}}(r) =  \begin{cases}
        0, & r \in [a, b] \\
        +\infty, & \text{o.w.}
    \end{cases}, \quad &
    \imath^{*}_{[a,b]}(t) = \sup_{u \in [a,b]} (u \cdot t) =\max (-at, bt), 
    \end{aligned}
\end{equation}
where $\imath^{*}_{[a,b]}$ denotes its Legendre-Fenchel conjugate. Since the function $\imath_{[a,b]}$ is convex lower semi-continuous, therefore $\imath_{[a,b] }= \imath_{[a,b]}^{**}$, we can apply the variational form of the $f$-divergence~\citep{nguyen2010estimating}, \citep[Theorem 7.26]{ Polyanskiy_Wu_2024}, exploited by $f$-GAN \citep{nowozin2016f}, leading to a form requiring only expected values 
\begin{align}
    \mathcal{D}_{\imath}\big(\tilde{\nu} \Vert {\nu} \big) & =\int_{\mathcal{Y}}  \imath_{[a,b]} \left( \frac{{\tilde{\nu}}}{\nu} (\rvy) \right) \nu(\rvy) d\rvy = \sup_{\eta} \underbrace{\int_{\mathcal{Y}} \eta(\rvy) {\tilde{\nu}}(\rvy) d\rvy}_{\E_{\tilde{\rvy}\sim\tilde{\nu}}[\eta(\tilde{\rvy})]}  - \underbrace{\int_{\mathcal{Y}} \imath^{*}_{[a,b]}(\eta(\rvy) ){\nu}(\rvy) d\rvy}_{\E_{\rvy \sim \nu} [\imath^{*}_{[a,b]}(\eta(\rvy))]}. \label{phi_3}  
\end{align}
To match \ref{static0}, we focus on $a=0$ and $b=c\ge 1$, such that $\imath^{*}_{[0,c]}(t) = c\cdot \max(0, t)$ and for notational compactness denote $\eta_+(\rvy)=\max(0,\eta(\rvy))$. Introducing  $\psi$ as a measurable function to act as a Lagrange multiplier to enforce the constraint in ~\ref{static0}  and combining with~\ref{phi_3} yields the problem 
\begin{equation}
 \begin{aligned}
\label{static-semidual01}
\inf_\pi \sup_{\psi,\eta} \int ( \mathtt{c}(\rvx,\rvy) + \eta(\rvy) - \psi(\rvx)) \pi(\rvx,\rvy) d\rvx d\rvy  + \int \psi(\rvx)  \mu(\rvx) d\rvx 
    - c \int \eta_+(\rvy) \nu(\rvy) d\rvy.
\end{aligned}   
\end{equation}
As described in Appendix~\ref{static-derive}, since $\mathtt{c}$ is convex and lower semi-continuous, we interchange the $\inf_\pi$ and $\sup_\eta$ and   apply what is known as the $\mathtt{c}$-transform of $-\eta(\rvy)$~\citep{santambrogio2015optimal, villani2009optimal} to obtain the dual problem with measurable map $T:\mathcal{X}\rightarrow\mathcal{Y}$
\begin{equation}
    \underset{\eta}{\sup} \, \underset{T}{\inf} \underbrace{\int_{\mathcal{X}} \big( \mathtt{c}(\rvx,T(\rvx)) +  \eta(T(\rvx)) \big) \mu(\rvx) d\rvx}_{\E_{\rvx \sim \mu} [\mathtt{c}(\rvx,T(\rvx)) +  \eta(T(\rvx))]}  -  \underbrace{c\int_{\mathcal{Y}}  \eta_+(\rvy)  \nu(\rvy) d\rvy}_{\E_{\rvy \sim \nu} [c\cdot \max (0,\eta(\rvy))]} , \label{static-semidual-main}
\end{equation}
For computational implementation $T$ and $\eta$ are parameterized using neural networks with associated parameters $\theta_T$ and $\theta_\eta$ and expectations are estimated using samples from $\mu$ and $\nu$ as described in the Algorithm \ref{neural-static}.
\begin{algorithm}[ht]
\small
\caption{\texttt{Static-Neural-SS}: Learning Algorithm for Static Subset Alignment}
\label{neural-static}
\footnotesize
\linespread{0.50}\selectfont
\begin{algorithmic}[1]
\REQUIRE Source distribution $\mu$ and target distribution $\nu$, cost function $\mathtt{c}(\cdot,\cdot)$,
reweighting bound $c$, neural networks $T(\cdot;\theta_T)$ and $\eta(\cdot;\theta_\eta)$, batch size $N$,
number of updates $n_T$ and $n_\eta$, optimizers $\mathrm{optim}_T$ and $\mathrm{optim}_\eta$.
\ENSURE Trained neural estimator for transport map $T$.
\FOR{$\ell = 1,2,\dots,L$}
  \FOR{$k = 1,2,\dots,n_T$}
    \STATE Sample $\{\rvx_i\}_{i=1}^N \sim \mu$ and $\{\rvy_j\}_{j=1}^N \sim \nu$
    \STATE 
    $\displaystyle
        \begin{aligned}
        \mathrm{grad}_{\theta_T} &\leftarrow \nabla_{\theta_T}\frac{1}{N}\sum_{i=1}^N
        \Bigl[\mathtt{c}(\rvx_i, T(\rvx_i;\theta_T)) + \eta(T(\rvx_i;\theta_T);\theta_\eta)\Bigr]
        \end{aligned}
    $  
    \STATE Using $\mathrm{optim}_T$ update $\theta_T$ via gradient descent
  \ENDFOR
  \FOR{$k = 1,2,\dots,n_\eta$}
    \STATE Sample $\{\rvx_i\}_{i=1}^N \sim \mu$ and $\{\rvy_j\}_{j=1}^N \sim \nu$
    \STATE
        $\displaystyle
        \begin{aligned}
        \mathrm{grad}_{\theta_\eta}
        &\leftarrow\nabla_{\theta_\eta}\left[
        \frac{c}{N}\sum_{j=1}^N \max\!\bigl(0,\eta(\rvy_j;\theta_\eta)\bigr)
        -\frac{1}{N}\sum_{i=1}^N \eta\!\bigl(T(\rvx_i;\theta_T);\theta_\eta\bigr)
        \right]
        \end{aligned}$
\STATE Using $\mathrm{optim}_\eta$ update $\theta_\eta$ via gradient ascent 
  \ENDFOR
\ENDFOR
\end{algorithmic}
\end{algorithm}
\subsection{Derivation of Static Neural Subset Selection}
\label{static-derive}
We denote the problem expressed in \ref{static-semidual01} as $\inf_\pi \sup_{\psi} \sup_{\eta}\mathcal{L} \left( \pi, \psi, \eta \right)$, where the Lagrangian is 
\begin{small}
    \begin{equation}
\label{lagrange}
\begin{aligned}
    \mathcal{L} \left( \pi, \psi, \eta \right) 
     \!= \! \int_{\mathcal{X} \times \mathcal{Y}} \!\left( \mathtt{c}(\rvx,\rvy) \!+\!  \eta(\rvy) \!+\! \psi(\rvx) \! \right) \pi(\rvx,\rvy) d\rvx d\rvy  \!+ \! \int_{\mathcal{X}} \psi(\rvx) \, \mu(\rvx) d\rvx 
    \!-\! c \int_{\mathcal{Y}} \eta_+(\rvy) \nu(\rvy) d\rvy.
\end{aligned}  
\end{equation}
\end{small}
We proceed to interchange the $\sup$ and $\inf$,\footnote{This requires us to verify Slater's constraint qualifications, which are: (i) primal is convex wrt $\pi$ ,(which is obvious), (ii) dual is concave wrt $\eta$, which is also obvious (iii) relative interior for inequality constraints set is non-empty, which can be verified by looking at the fact that for any $\tilde{\nu}$ the distribution $\pi(\rvx, \rvy) = \mu(\rvx)\tilde{\nu}(\rvy)$ is feasible and one can see that  if one defines the feasible set of coupling $\Pi_{c}(\mu, \nu) = \{\pi \in \mathcal{P}(\mathcal{X} \times \mathcal{Y}) : \pi_{\mathcal{X}}=\mu,  \pi_{\mathcal{Y}}\leq c\nu \}$, then for $ \forall \, 1 \leq c_{0} \leq c_{1}, \Pi_{c_{0}}(\mu, \nu) \subseteq \Pi_{c_{1}}(\mu, \nu) $, which in other words mean that $\Pi_{c=1}(\mu, \nu)$ is a subset of feasible solutions for all values of $c>1$, therefore relative-interior in non-empty.} which is allowed due to the strong duality property associated with optimal transport when the cost $\mathtt{c}$ is convex and lower semi-continuous,
\begin{equation}
    \begin{aligned}
    \underset{\pi}{\inf} \sup_{\psi, \eta} \, \mathcal{L} \left( \pi, \psi, \eta \right) = \underset{\eta,\psi }{\sup} \ \underset{\pi}{\inf} \,  \mathcal{L} \left( \pi, \psi, \eta \right).
\end{aligned}
\end{equation}
optimizing with respect to $\pi$ for given $\eta$ and $\psi$,  the integrand in the first term of~\ref{lagrange} is unbounded from below at any point ${\mathtt{c}}(\rvx,\rvy) + \eta(\rvy) - \psi(\rvx)<0$.  Thus, $\eta$ and $\psi$ need to ensure  that  $\mathtt{c}(\rvx,\rvy) + \eta(\rvy) - \psi(\rvx) \geq 0$. This constraint requires that, for any  $\rvx \in \text{supp}(\mu) \subseteq \mathcal{X}$, $\psi(\rvx) =  \inf_{\rvy\in\mathcal{Y}} ({\mathtt{c}}(\rvx, \rvy) + \eta(\rvy) )$~\citep[Theorem 5.10 and Remark 5.13]{villani2009optimal} . This definition of $\psi$ corresponds to the $\mathtt{c}$-transform of $-\eta(\rvy)$ in the optimal transport literature~\citep{santambrogio2015optimal, villani2009optimal}. Then, the inner infimum with respect to $\pi$ is attained with zero value  if $\forall \, (\rvx,\rvy) \in (\mathcal{X} \times \mathcal{Y}) : \pi(\rvx,\rvy)>0 \implies \mathit{\mathtt{c}}(\rvx,\rvy) + \eta(\rvy) - \psi(\rvx) = 0$. 
Therefore, the dual problem becomes
\begin{align}
    \underset{\eta}{\sup} \,  &  \int_{\mathcal{X}} \left( \inf_{\rvy\in\mathcal{Y}} ({\mathtt{c}}(\rvx, \rvy) + \eta(\rvy) \right )  \mu(\rvx) d\rvx  - c \int_{\mathcal{Y}} \eta_+(\rvy) \, \nu(\rvy)  d\rvy. \label{static-semidual03}\\
    &= \underset{\eta}{\sup} \, \underset{T}{\inf} \int_{\mathcal{X}} \big( \mathit{\mathtt{c}}(\rvx,T(\rvx)) +  \eta(T(\rvx)) \big) \mu(\rvx) d\rvx - c \int_{\mathcal{Y}} \eta_+(\rvy) \, \nu(\rvy) d\rvy \label{static-semidual04}\\
    &=\underset{\eta}{\sup} \, \underset{T}{\inf} \underset{\rvx\sim {\mu}}{\mathbb{E}} \big[ \mathit{\mathtt{c}}(\rvx,T(\rvx)) +  \eta(T(\rvx)) \big] - c\underset{\rvy\sim {\nu}}{\mathbb{E}} [\eta_+(\rvy)]. \label{static-semidual05}\\
   &= \underset{T}{\inf}\, \underset{\eta}{\sup}  \underset{\rvx\sim {\mu}}{\mathbb{E}} \big[ \mathit{\mathtt{c}}(\rvx,T(\rvx))  \big]  + \underset{\rvx\sim {\mu}}{\mathbb{E}} \big[\eta(T(\rvx) \big]  - c\underset{\rvy\sim {\nu}}{\mathbb{E}} [\eta_+(\rvy)]. \label{static-semidual06} \\
    &= \underset{T}{\inf}   \underset{\rvx\sim {\mu}}{\mathbb{E}} \big[ \mathit{\mathtt{c}}(\rvx,T(\rvx))  \big]  + \mathcal{D}_{\imath_{[0,c]}} ( T_\sharp \mu \Vert \nu)=\underset{\substack{T\\ T_\# \mu \le c \nu}}{\inf}   \underset{\rvx\sim {\mu}}{\mathbb{E}} \big[ {\mathtt{c}}(\rvx,T(\rvx))  \big]. \label{static-semidual07}
\end{align}
Equation~\ref{static-semidual03} and~\eqref{static-semidual04} are equal due to a theorem by~\citet[Theorem 3A]{rockafellar1976integral};~\eqref{static-semidual04} and~\eqref{static-semidual05} are equivalent by definition;~\eqref{static-semidual05} and~\eqref{static-semidual06} are equivalent since the function is convex with respect to $T$ and concave with respect to $\eta$; and~\eqref{static-semidual06} and~\eqref{static-semidual07} are equivalent by the variational formula for the $f$-divergence. Thus, we obtain a relaxed Monge formulation. 
\subsection{Derivation of Dynamic Neural Subset Selection}
\label{Derivation of Dynamic Neural Subset Selection}
The modified Benamou-Brenier problem takes the following minimax form
\begin{equation}
    \begin{aligned} \label{benamou_n}
    \underset{\rho_{t}, \rvv_{t}}{\inf} & \int_{0}^{1}\!\!\! \underset{x \sim \rho_{t}}{\mathbb{E}} \left[ \frac{\Vert \rvv_{t}(\rvx) \Vert^{2}}{2}\right] dt  + \sup_{\eta:\Omega \rightarrow \mathbb{R}} \mathbb{E}_{\rvx\sim \rho_{1}}\left[\eta(\rvx)\right] -  \underset{\rvx\sim\nu}{\mathbb{E}}\left[\imath^{*}_{[0,c]}(\eta(\rvx))\right] \\
    \text{s.t.} & \quad \frac{\partial}{\partial_{t}}\rho_{t}(\rvx) + \text{div}{\left(\rho_{t}(\rvx) \rvv_{t}(\rvx)\right)} = 0, 
    \quad \rho_0(\rvx) = \mu(\rvx)
\end{aligned}
\end{equation}
Combining the objective and constraints in~\eqref{benamou_n} to obtain the Lagrangian
\begin{align}
    \mathcal{L}(\rvv_{t}, \rho_{t}, \psi_{0}, \varphi_{t}, \eta) = \overbrace{\int_{0}^{1}\!\!\!\int_{\Omega} \frac{\Vert \rvv_{t}(\rvx) \Vert^{2}}{2}  \rho_{t}(\rvx) d\rvx dt}^{ (\text{I})} \nonumber
    + \overbrace{\int_{\Omega}\psi_{0}(\rvx)\left(\rho_{0}(\rvx) - \mu(\rvx) \right) d\rvx}^{(\text{II})} \\  + 
    \overbrace{\int_{0}^{1}\!\!\!\int_{\Omega} \varphi_{t}(\rvx) \frac{\partial}{\partial_{t}}\rho_{t}(\rvx)d\rvx dt}^{(\text{III})} + \overbrace{\int_{0}^{1}\!\!\!\int_{\Omega} \varphi_{t}(\rvx)\textrm{div} \left(\rho_{t}(\rvx)\rvv_{t}(\rvx)\right)   d \rvx dt}^{(\text{IV})} \nonumber  \\   +  
    \overbrace{\sup_{\eta} \biggl( \int_{\Omega} \eta(\rvx) \rho_{1}(\rvx) d\rvx - c \int_{\Omega} \max(0,\eta(\rvx))  {\nu}(\rvx) d\rvx\biggr)}^{(\text{V})}.  \label{d-lagrangain0}
\end{align}
Since $\rho_t(\rvx)$ is supported on a bounded subset $\Omega \subset \mathbb{R}^d$, one can change the order of integration. Therefore, for term (III), by changing the order of integration and then computing integration by parts one obtains,
\begin{align}
    (\text{III})=\int_{\Omega}  \int_{0}^{1} \varphi_{t}(\rvx) \frac{\partial}{\partial_{t}} \rho_{t}(\rvx) d\rvx dt & = \int_{\Omega} \varphi_{1}(\rvx) \rho_{1}(\rvx) d\rvx- \int_{\Omega} \varphi_{0}(\rvx) \rho_{0}(\rvx)d\rvx \nonumber \\ 
    & \quad -\int_{0}^{1}\!\!\!\int_{\Omega} \frac{\partial}{\partial_{t}} \varphi_{t}(\rvx) \rho_{t}(\rvx) d\rvx dt. \label{lag_III}
\end{align}
To simplify $(\text{IV})$, we can use product rule of derivatives to write 
$$\varphi_{t}(\rvx) \text{div}{\left(  \rho_{t}(\rvx) \rvv_{t}(\rvx)\right)} = \textrm{div} \left( \varphi_{t}(\rvx) \rho_{t}(\rvx) \rvv_{t}(\rvx) \right) - \rho_{t}(\rvx) \langle \nabla \varphi_{t}(\rvx), \rvv_{t}(\rvx) \rangle. $$
Therefore by combining the above identity with Gauss's theorem, one obtains
\begin{equation}
\begin{aligned}
(\text{IV}) &=  \int_{0}^{1}\!\!\!\int_{\Omega} \textrm{div} \left( \varphi_{t}(\rvx) \rho_{t}(\rvx) \rvv_{t}(\rvx) \right) d\rvx dt - \int_{0}^{1}\!\!\!\int_{\Omega} \rho_{t}(\rvx) \langle \nabla \varphi_{t}(\rvx), \rvv_{t}(\rvx) \rangle d\rvx dt \\
&= \underbrace{\int_{0}^{1} \!\!\! \oint_{\partial\Omega} \varphi_{t}(\rvx_{t})  \rho_{t}(\rvx_{t}) \langle \rvv_{t}(\rvx), d\vn\rangle dt}_{= 0}  - \int_{0}^{1} \!\!\! \int_{\Omega} \rho_{t}(\rvx) \langle \nabla \varphi_{t}(\rvx), \rvv_{t}(\rvx) \rangle d\rvx dt. 
\end{aligned} \label{lag_IV_1}
\end{equation}
From the boundary condition on optimal transport (see the discussion above~\eqref{d-flow1}, and also~\citep[section 4.1]{figalli2021invitation}) , the first part of the right-hand side of~\eqref{lag_IV_1} is zero; therefore,
\begin{equation}
    (\text{IV}) = \int_{0}^{1} \!\!\!\int_{\Omega} \varphi_{t}(\rvx) \cdot \textrm{div}{\left(  \rho_{t}(\rvx) \rvv_{t}(\rvx)\right)} d\rvx dt = - \int_{0}^{1} \!\!\!\int_{\Omega}\langle \nabla\varphi_{t}(\rvx), \rvv_{t}(\rvx) \rangle \rho_{t}(\rvx) d\rvx dt. \label{lag_IV}
\end{equation}
To eliminate the primal variable $\rvv_{t}(\rvx)$, substitute~\eqref{lag_III} and~\eqref{lag_IV} into~\eqref{d-lagrangain0} and compute the variational derivative to obtain the stationary condition. For that one can write the terms of Lagrangian depending on $\rvv_t(\rvx)$ as
\begin{equation}
    \tilde{\mathcal{L}}(\rvv_{t}) = \int_{0}^{1}\!\!\!\int_{\Omega} \bigg(\frac{\Vert \rvv_{t}(\rvx)\Vert^{2}}{2} - \langle \nabla\varphi_{t}(\rvx), \rvv_{t}(\rvx) \rangle\bigg)\rho_{t}(\rvx) d \rvx dt.
\end{equation}
With the additive perturbation function $\bm{\tau}_t$ vanishing at $t=0$ and $t=1$ and a scalar $\varepsilon$, the Lagrangian  $\tilde{\mathcal{L}}(\rvv_{t} + \varepsilon\bm{\tau}_t)$ is
\begin{equation}
    \begin{aligned}
    \tilde{\mathcal{L}}(\rvv_{t} + \varepsilon\bm{\tau}_t) &= \int_{0}^{1}\!\!\!\int_{\Omega} \bigg(\frac{\Vert \rvv_{t}(\rvx) + \varepsilon\bm{\tau}_{t}(\rvx)\Vert^{2}}{2} - \big\langle \nabla\varphi_{t}(\rvx), \ \  \rvv_{t}(\rvx) + \varepsilon\bm{\tau}_t (\rvx) \big\rangle\bigg)\rho_{t}(\rvx) d\rvx dt \nonumber \\
    & = \int_{0}^{1}\!\!\!\int_{\Omega} \left( \frac{\Vert \rvv_{t}(\rvx)\Vert^{2}}{2} - \big\langle \nabla\varphi_{t}(\rvx), \rvv_{t}(\rvx) \big\rangle \right) \rho_{t}(\rvx) d\rvx dt \\
    &\quad + \int_{0}^{1}\!\!\!\int_{\Omega} \left( \varepsilon^{2} \frac{\Vert \bm{\tau}_{t}(\rvx)\Vert^2}{2} 
    + \varepsilon \big\langle \rvv_{t}(\rvx) - \nabla\varphi_{t}(\rvx) , \ \bm{\tau}_{t}(\rvx) \big\rangle \right) \rho_{t}(\rvx) d\rvx dt,
    \end{aligned}
\end{equation}
and variational derivative is 
\begin{equation}
    \left. \delta\tilde{\mathcal{L}}(\rvv_{t}(\rvx)) \right|_{\rvv_{t}} = \left. \frac{d}{d\varepsilon} \right|_{\varepsilon = 0} \tilde{\mathcal{L}}(\rvv_{t} + \varepsilon \bm{\tau}_{t}) = \int_{0}^{1}\!\!\!\int_{\Omega} \langle \rvv_{t}(\rvx) - \nabla\varphi_{t}(\rvx), \, \bm{\tau}_{t}(\rvx) \rangle \rho_{t}(\rvx) d\rvx d t. \label{v_stationarity}
\end{equation}
The stationarity condition requires $\delta_{\rvv_{t}}\tilde{\mathcal{L}}(\bm{\rvv}_{t}(\rvx)) = 0$. For arbitrary perturbation $\bm{\tau}_{t}(\rvx)$, the variational-derivative $\delta_{\rvv_{t}}\tilde{\mathcal{L}}(\bm{\rvv}_{t}(\rvx)) = 0 $ if and only if
\begin{equation}
\rvv_{t}(\rvx) = \nabla\varphi_{t}(\rvx).
\end{equation}
Therefore, one can write the Lagrangian as
\begin{equation}
  \begin{aligned}
    \mathcal{L}(\rho_{t}, \psi_{0}, \varphi_{t}, \eta) &= \int_{\Omega} \psi_{0}(\rvx) \cdot \left(\rho_{0}(\rvx) - \mu(\rvx)\right) d\rvx + \int_{\Omega} \varphi_{1}(\rvx) \rho_{1}(\rvx) d\rvx - \int_{\Omega} \varphi_{0}(\rvx) \rho_{0}(\rvx) d\rvx  \nonumber \\
    & \quad +   \int_{\Omega} \eta(\rvx) \rho_{1} (\rvx) d\rvx - c \cdot \int_{\Omega} \max(0,\eta(\rvx)) \nu(\rvx) d\rvx \nonumber \\
    & \quad - \int_{0}^{1} \!\!\!\int_{\Omega} \left( \frac{\partial}{\partial t} \varphi_{t}(\rvx) + \frac{\Vert \nabla \varphi_{t}(\rvx) \Vert^{2}}{2} \right) \rho_t(\rvx) d\rvx dt. \label{lag_final}
\end{aligned}  
\end{equation}
Similarly, by computing $\delta_{\psi_{0}} \mathcal{L}$, $\delta_{\rho_{1}} \mathcal{L}$, and using stationary conditions, one obtains the condition,
\begin{align}
\psi_{0}(\rvx) = \varphi_{0}(\rvx) \label{psi0}, \\
\eta(\rvx) = {-}\varphi_{1}(\rvx) \label{rho1}.
\end{align}
Therefore, Lagrangian is simplified to
\begin{equation}
\begin{aligned}
    \mathcal{L}(\rho_{t}, \varphi_{t}, \eta) &=  - \int_{\Omega} \varphi_{0}(\rvx) \mu (\rvx) d\rvx -c\int_{\Omega} \max(0,-\varphi_{1}(\rvx)) \nu(\rvx) d\rvx \\ 
    & \quad
    - \int_{0}^{1} \!\!\! \int_{\Omega} \left( \frac{\partial}{\partial t} \varphi_{t}(\rvx) + \frac{\Vert \nabla \varphi_{t}(\rvx) \Vert^{2}}{2} \right) \rho_{t}(\rvx) d\rvx dt. \label{lag_final_1}
\end{aligned}
\end{equation}
The simplified problem (\eqref{opt_lag_final} in the main body) is
\begin{small}
\begin{equation} \notag
\underset{\rho_{t}}{\sup} \, \underset{\varphi_{t}}{\inf} \underset{\rvx \sim {\mu}}{\mathbb{E}} \big[ \varphi_0(\rvx) \big] +c\cdot \underset{\rvx \sim {\nu}}{\mathbb{E}} \big[\max(0,-\varphi_{1}(\rvx))\big] + \int_{0}^{1} \underset{\rvx_t \sim \rho_{t}}{\mathbb{E}} \left[\frac{\partial}{\partial t} \varphi_{t}(\rvx_t) + \frac{\Vert \nabla \varphi_{t}(\rvx_t)\Vert^{2}}{2} \right]dt. 
\end{equation}
\end{small}

\section{Thresholding for PU{-}Learning and Rejection Sampling}
\label{appendix: Thresholding for PU{-}Learning and Rejection Sampling}
Our idea of rejection sampling and thresholding for PU{-}learning  relies on the fact that the dual form of range divergence is zero when the supremum in the dual is attained by the function $\eta^{\star}(\rvx)$ with $\tilde{\nu}(\rvx)=\rho_{1}^{\star}(\rvx)$ i.e.
\begin{equation}
    \underset{\rvx\sim \tilde{\nu}}{\mathbb{E}}[\eta^{\star}(\rvx)] -c \underset{\rvx \sim \nu}{\mathbb{E}}[\mathtt{ReLU}(\eta^{\star}(\rvx))]= 0 \label{rej1}
\end{equation}
By defining $\mathcal{A}\!\!=\!\mathrm{supp}(\nu)$ , $\mathcal{\tilde{A}}\!\!=\!\mathrm{supp} (\tilde{\nu})$, and $\mathcal{\bar{A}}\!=\! \mathcal{A}/\mathcal{\tilde{A}}$, one can also see that $\mathcal{\tilde{A}} \cap \mathcal{\bar{A}} = \emptyset $, therefore one can write \eqref{rej1} as
\begin{equation}
   \int_{\mathcal{\tilde{A}}} \eta^{\star}(\rvx) \tilde{\nu}(\rvx)d\rvx - c\int_{\mathcal{\tilde{A}}} \mathtt{ReLU}\left(\eta^{\star}(\rvx)\right) \nu(\rvx)d\rvx - c \int_{\mathcal{\bar{A}}} \mathtt{ReLU}(\eta^{\star}(\rvx)) \nu(\rvx)d\rvx = 0. \label{rej1p}
\end{equation}
One can further write 
\begin{equation}
    \eta^{\star}(\rvx) = \hspace{-1mm}\mathtt{ReLU}\left(\eta^{\star}(\rvx) \right) - \mathtt{ReLU}\left(-\eta^{\star}(\rvx) \right). \label{rej2}
\end{equation}
After substituting \eqref{rej2} into \eqref{rej1p} one obtains
\begin{equation}
    \overbrace{\int_{\mathcal{\tilde{A}}} \mathtt{ReLU}(\eta^{\star}(\rvx)) (\tilde{\nu}(\rvx) - c\nu(\rvx))d\rvx}^{\text{LHS}} = \overbrace{ \int_{\mathcal{\tilde{A}}} \mathtt{ReLU}(-\eta^{\star}(\rvx)) \tilde{\nu}(\rvx)d\rvx + c \int_{\mathcal{\bar{A}}} \mathtt{ReLU}(\eta^{\star}(\rvx)) \nu(\rvx)d\rvx}^{\text{RHS}} \label{rej3}
\end{equation}
The dual form  \eqref{rej1p} is optimal with zero duality gap, if the primal form satisfies $ \forall \rvx \in \mathcal{A}, \  \imath_{[0,c]} \big(\frac{{\tilde{\nu}}}{\nu} (\rvx)\big) = 0$, which can also be restricted to~$ \forall \ \rvx \in \mathcal{\tilde{A}} \  \imath_{[0,c]} \big(\frac{{\tilde{\nu}}}{\nu} (\rvx)\big) = 0$. This is equivalent to $\tilde{\nu}(\rvx) \leq c \nu(\rvx)  \ \textrm{almost-everywhere in} \ \mathcal{\tilde{A}}$. Therefore, one can say that $0 \geq \mathrm{LHS}$ and also
\begin{equation}
    0 \geq \overbrace{ \int_{\mathcal{\tilde{A}}} \mathtt{ReLU}(-\eta^{\star}(\rvx)) \nu(\rvx)d\rvx + c \int_{\mathcal{\bar{A}}} \mathtt{ReLU}(\eta^{\star}(\rvx)) \nu(\rvx)d\rvx}^{\text{RHS}} \label{rej4}
\end{equation}
We can now see that both integrands in~\eqref{rej3} are nonnegative and sum to a value less than or equal to zero, which is only possible if both are equal to zero. Therefore, one can write
\begin{equation}
    0 = \int_{\mathcal{\tilde{A}}} \mathtt{ReLU}(-\eta^{\star}(\rvx)) \nu(\rvx)d\rvx + c \int_{\mathcal{\bar{A}}} \mathtt{ReLU}(\eta^{\star}(\rvx)) \nu(\rvx)d\rvx \label{rej5}
\end{equation}
Further, two non-negative integrals are evaluated on two mutually exclusive sets; therefore, to have a sum equal to zero value, we can conclude that each integral is zero individually. Therefore, we can write
\begin{equation}
    0 = \int_{\mathcal{\tilde{A}}} \mathtt{ReLU}(-\eta^{\star}(\rvx)) \nu(\rvx)d\rvx = c \int_{\mathcal{\bar{A}}} \mathtt{ReLU}(\eta^{\star}(\rvx)) \nu(\rvx)d\rvx \label{rej6}
\end{equation}
Equation \ref{rej6} is therefore equivalent to the following element-wise test:
\begin{equation}
    \begin{aligned}
       \eta^{\star}(\rvx) \geq 0, \ \text{almost surely in} \ \mathcal{\tilde{A}} \\
       \eta^{\star}(\rvx) < 0, \ \text{almost surely in} \ \mathcal{\bar{A}} \\
    \end{aligned}
    \label{sp_s}
\end{equation}

Additionally, from the \eqref{rej4}, one can also conclude that 
\begin{equation}
    \overbrace{\int_{\mathcal{\tilde{A}}} \mathtt{ReLU}(\eta^{\star}(\rvx)) (\tilde{\nu}(\rvx) - c\nu(\rvx))d\rvx}^{\text{LHS}} = 0,
\end{equation}
which is a complementary slackness condition in the sense that $\tilde{\nu}(\rvx)<c\nu(\rvx) \implies \eta^{*}(\rvx)=0$ almost everywhere in $\mathcal{A}$. During the neural network training with finite data{-}points, the potential function $\eta$ is usually sub-optimal, and its sign cannot be relied on; therefore, instead of directly using the sign, one can sort values of potential at data points and select a predetermined proportion (prior) of data-points. Therefore, for training PU-learning models, we applied both sign-based and sorting-based filtration of data. From Figures~\ref{fig_toy-add1},~\ref{fig_toy-add2} and~\ref{fig_toy-add3}, one can observe that the optimal potential for static problem exactly follows \eqref{sp_s}, whereas in the dynamic case the sign of $\varphi_{1}$ is inverted,  which is due to the relation obtained in \eqref{rho1}, which ensures that for the at optimal $\varphi_{1}^{\star}$ following relation holds
\begin{equation}
    \begin{aligned}
       \varphi_{1}^{\star}(\rvx) \leq 0, \ \text{almost surely in} \ \mathcal{\tilde{A}} \\
       \quad \varphi_{1}^{\star}(\rvx) > 0, \ \text{almost surely in} \ \mathcal{\bar{A}}. \\
    \end{aligned}
    \label{sp_d}
\end{equation}

\section{Survey of Recent Work on Neural Optimal Transport}
In this section, we discuss the recent related work on computational optimal transport and its applications. More specifically, we consider the works that are related to neural estimation of optimal transport maps with occasional reference to theoretical developments.  

\subsection{Static Neural Optimal Transport}
\label{app:static-survey}
\citet{seguy2018large} employed stochastic gradient-based approaches in one of the earliest works to estimate the optimal transport map using neural networks. Notably, the work by \cite{seguy2018large} differed from early work by \citet{genevay16stoch} in the sense that the later work employed stochastic gradient-based methods to estimate the transport plan for large-scale data, whereas earlier work by \cite{genevay16stoch} only minimized the optimal transport loss using stochastic-gradient-based methods. This is also in contrast to the well-known Wasserstein-GAN \citep{arjovsky2017wasserstein, gulrajani2017improved} that employs the Kantorovich-Rubinstein duality to minimize the Wasserstein-1 loss function for generative modeling, where neural networks are employed as parameterizations for both dual-potential and data generator, but do not provide transport plans. Finally, the Sinkhorn-GAN employs an approximation of the discrete Wasserstein distance between latent representations of data and that of samples from non-informative prior \citep{genevay2018learning} for generative modeling. Now, we can see clear distinction between two different classes of approaches employing Wasserstein distances in generative modeling, the first class of works concerns with employing Wasserstein distance as a loss for generative modeling, without any explicit concern for obtaining the underlying transport plan across the distributions \citep{arjovsky2017wasserstein, gulrajani2017improved}. The second class seeks to learn a transport plan to realize the generative model. 

Efforts to learn Monge maps were motivated by a theorem by \cite{brenier1991polar}, which essentially states that, for continuous distributions with squared-Euclidean transportation cost, the optimal solution of the Monge problem is the gradient of a convex function \cite[Theorem 2.5.10]{figalli2021invitation}. Therefore, initially, gradients of input-convex neural networks (ICNN) \citep{amos2017input} were employed to estimate the transport plan for the Wasserstein-2 distance \citep{makkuva2020optimal, korotin2021wasserstein, korotin2021neural}. This approach has also been employed to supervised conditional neural Monge maps \citep{bunne2022supervised} and unbalanced optimal transport \citep{lubeck2022neural}. The study by \cite{amos2023meta} focuses on the development of an efficient neural optimal solution that could be implemented quickly in more practical scenarios. This approach to solve Wasserstein-2 distances employing convex potentials involves computationally challenging evaluation of the Fenchel conjugate of an ICNN-parameterized convex function. More recent work in this direction focuses on improved optimization strategies and better ICNN architectures to bypass problems related to Fenchel conjugate evaluations and ICNN training \citep{amosamortizing, vesseron2024on}.  Recent work also focuses on some batch-based methods, which have been devised to improve the regularity of learned neural Monge maps \citep{uscidda2023monge, eyring2024unbalancedness}.

Another recent direction of work is based on the idea that ICNNs can be overly restrictive; therefore, more general neural network architectures should be employed to directly parameterize the transport maps \citep{rout2022generative, korotin2023neural}.  The work by \citet{fan2022scalable, fan2023neural} focuses on employing neural networks to approximate the solution for Monge's transport problem and also draws inspiration from the recent developments in neural-network-based parametric realizations for approximating Kantorovich plans. Recently, neural optimal transport has also been extended to unbalanced transportation settings  \citep{yang2018scalable, choi2023generative}. Another work directly related to the static subset selection problem is the inexact optimal transport~\cite{gazdieva2023extremal}. 

Unless there is a corresponding Monge mapping \citep{choi2024analyzing, mokrov2024energyguided, geuter2025universal}, optimal transport requires a stochastic transport plan. A recent body of work \citep{korotin2023neural, korotin2023kernel, asadulaev2024neural} deals with learning transportation plans using a weaker formulation of optimal transport \citep{gozlan2017kantorovich, backhoff2019existence} along with noise outsourcing techniques, which have also been extended to more general costs. Apart from the applications in image translation \citep{korotin2023neural}, neural optimal transport has been applied for biomedical image registration \citep{kim2024otmorph} and to study single-cell perturbations \citep{bunne2023learning}. Neural optimal transport has also been employed for metric learning \citep{howard2024differentiable, scarvelis2023riemannian}.

\subsection{Dynamic Neural Optimal Transport}
\label{app:dynamic-survey}
The potential applications of dynamic optimal transport in the cellular trajectory inference \citep{tong2020trajectorynet} and its connections with flow-based models for generative modeling \citep{huang2021convex, huguet2022manifold} has been instrumental in the recent research developments in this direction. Jordan-Kinderlehrer-Otto flow (JKO) is a time discretization scheme to solve Wasserstein gradient flows for different energy functionals \citep{JKO1998,santambrogio2017euclidean}. Therefore, a lot of effort done in that regard is focused on neural network-parameterized schemes to solve the JKO-flow problem for both cellular trajectory inference and generative modeling \citep{ma2021learning, fan2022variational, lambert2022variational, bunne22a, chen2023_jko, choi2023generative, choiscalable, altekruger2023neural, mokrov2021large, alvarez-melis2022optimizing}. The JKO scheme has also been studied for the applications related to molecular discovery \citep{alvarez-melis2022optimizing}. A recent study deals with convergence properties of JKO-based generative models \citep{cheng2024convergence}.

Recent developments in flow-matching models based on flow matching \citep{lipmanflow, albergo2023building, liu2023flow} for generative modeling led to even more interest in the development of algorithms to solve dynamic optimal transportation problems. The action-matching-based framework led to the development of a more general framework to solve both trajectory inference and generative modeling problems \citep{neklyudov2023action} for the cases where one could also sample from the trajectory between two terminal marginals.  Rectified flow matching \citep{liu2023flow, liu2024instaflow} uses the neural optimal transport in an additional rectification step to improve the linearity of flows so that after training the model, images could be generated efficiently with only a single-step integration along straight-line paths. For generative modeling, in contrast to target-conditional flow matching \citep{lipmanflow}, where during training, flows are conditioned on target samples, discrete optimal transport-conditioned flow matching employs the mini-batch optimal transport to create the conditionals \citep{pooladian2023multisample, pooladian2024neural, tong24simulation}. Another recent work \citep{kornilov2024optimal} attempts to alleviate the error accumulation problems associated with mini-batch optimal transport by learning straight paths between source and target distributions in a single step. Flow-matching \citep{albergo2023building, albergo2023stochastic}, diffusion models \citep{dickstein15, song2020improved, song2021denoising}, and Schr\"odinger bridges \citep{wang21deep, liu2022deep, liu2024generalized, shi2023diffusion, gushchin2023building, gushchin2023entropic, somnath2023aligned} are deeply interconnected under the framework of generalized bridge matching \citep{tong24simulation, albergo2023stochastic, tong2024improving, shi2023diffusion}. Recently, there have also been attempts to understand diffusion models as approaches to minimize the dynamic Wasserstein distances \citep{kwonscore22, khrulkov2023understanding}. Recent works generalize flow matching from different perspectives, \citet{chen2024flow} generalize the flow matching to the flows on Riemannian manifolds, \citet{atanackovic2025meta} attempt to extend the flow models to return meaningful flows for the data beyond training distributions, and \citet{haviv2025wasserstein} generalize the flow matching to the cases where data can be treated as distributions of distributions.

Additionally, there has been a recent dynamic extension to the conditional neural optimal transport \citep{hosseini2023conditional, kerrigan2024dynamic}. There have also been efforts to study neural network-based scalable approaches to solve high-dimensional partial differential equations \citep{wan2023scalable}.

\section{Additional Toy Results}
\label{app:add-toy}
\begin{figure}
    \centering
    \begin{subfigure}[t]{0.50\textwidth}
        \centering
        \includegraphics[width=\linewidth,clip=true,trim={0 0 0 0}]{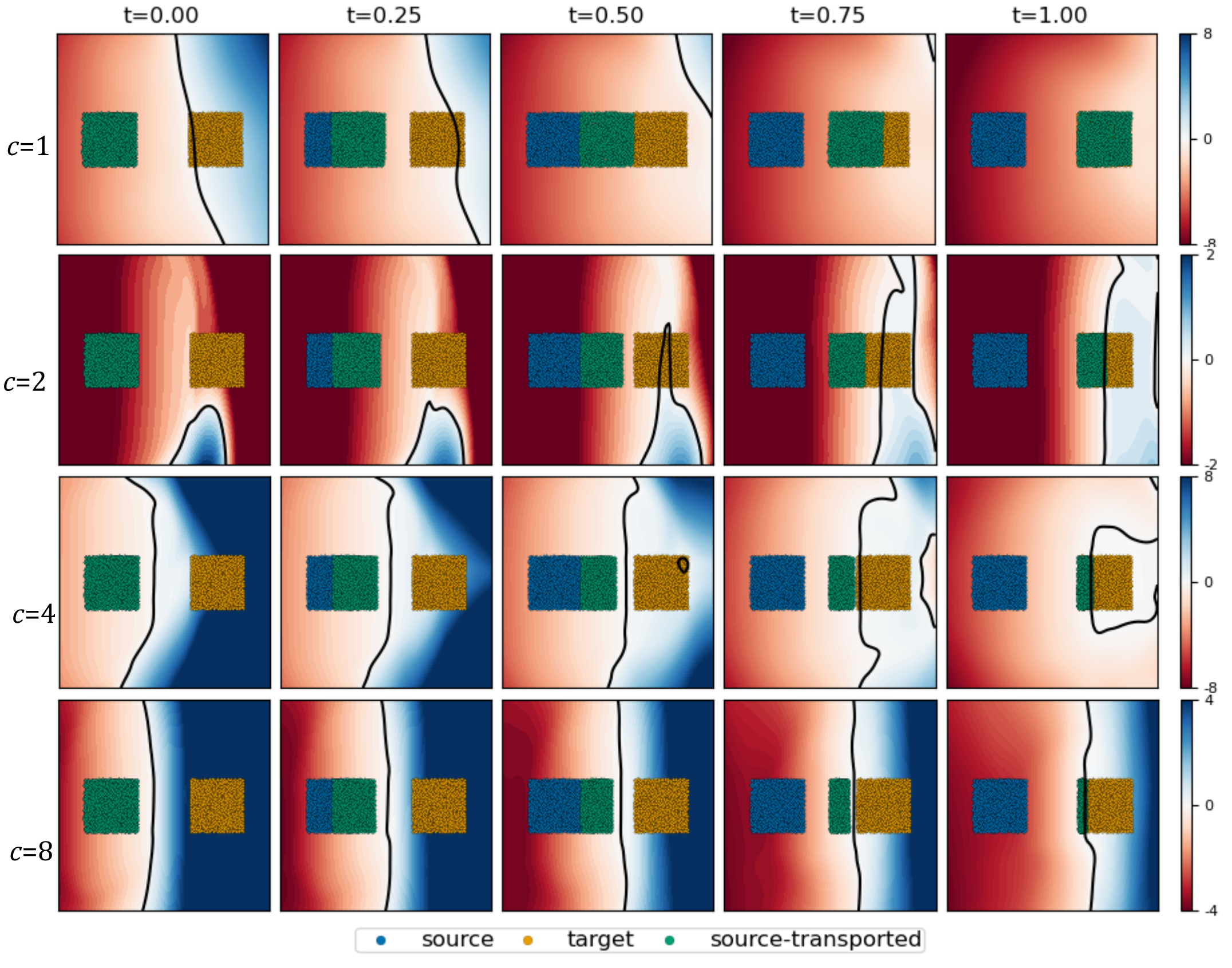}
        \caption{}
        \label{squares-evolution-all}
    \end{subfigure}\hspace{0mm}%
    \begin{subfigure}[t]{0.40\textwidth}
        \centering
        \includegraphics[width=\linewidth,clip=true,trim={0 0 0 0}]{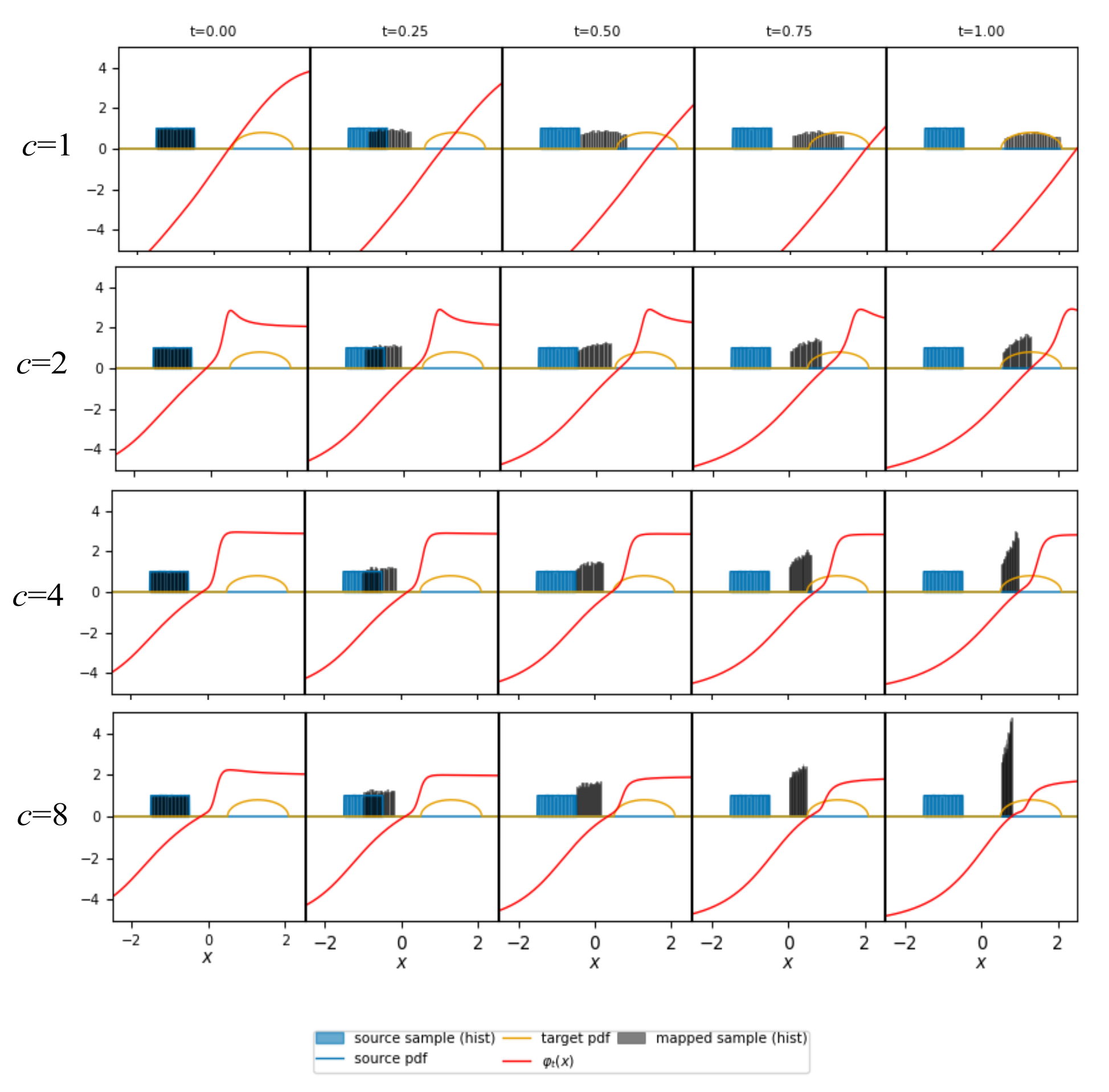}
        \caption{}
        \label{uniform2semicircle-evolution-all}
    \end{subfigure}\hspace{0mm}%
    
    \caption{Temporal snapshots at $t\in \{0.0, 0.25, 0.50, 0.75, 1.0\}$ for  $c\in{1,2,4,8}$. {(a)} Evolution between two 2D uniform distributions shown in Figure~\ref{squares-final-all}; the black contour line indicates the zero-level set. {(b)} Evolution from 1D uniform to semicircular density is shown in Figure~\ref{uniform2semicircle-final-all} with support subset alignment. $\varphi_{t}(\rvx)$ is denoted by red curve.}
    \label{fig_toy-add1}
\end{figure}

\begin{figure}
    \centering
    \begin{subfigure}[t]{0.58\textwidth}
        \centering
        \includegraphics[width=\linewidth,clip=true,trim={0 0 0 0}]{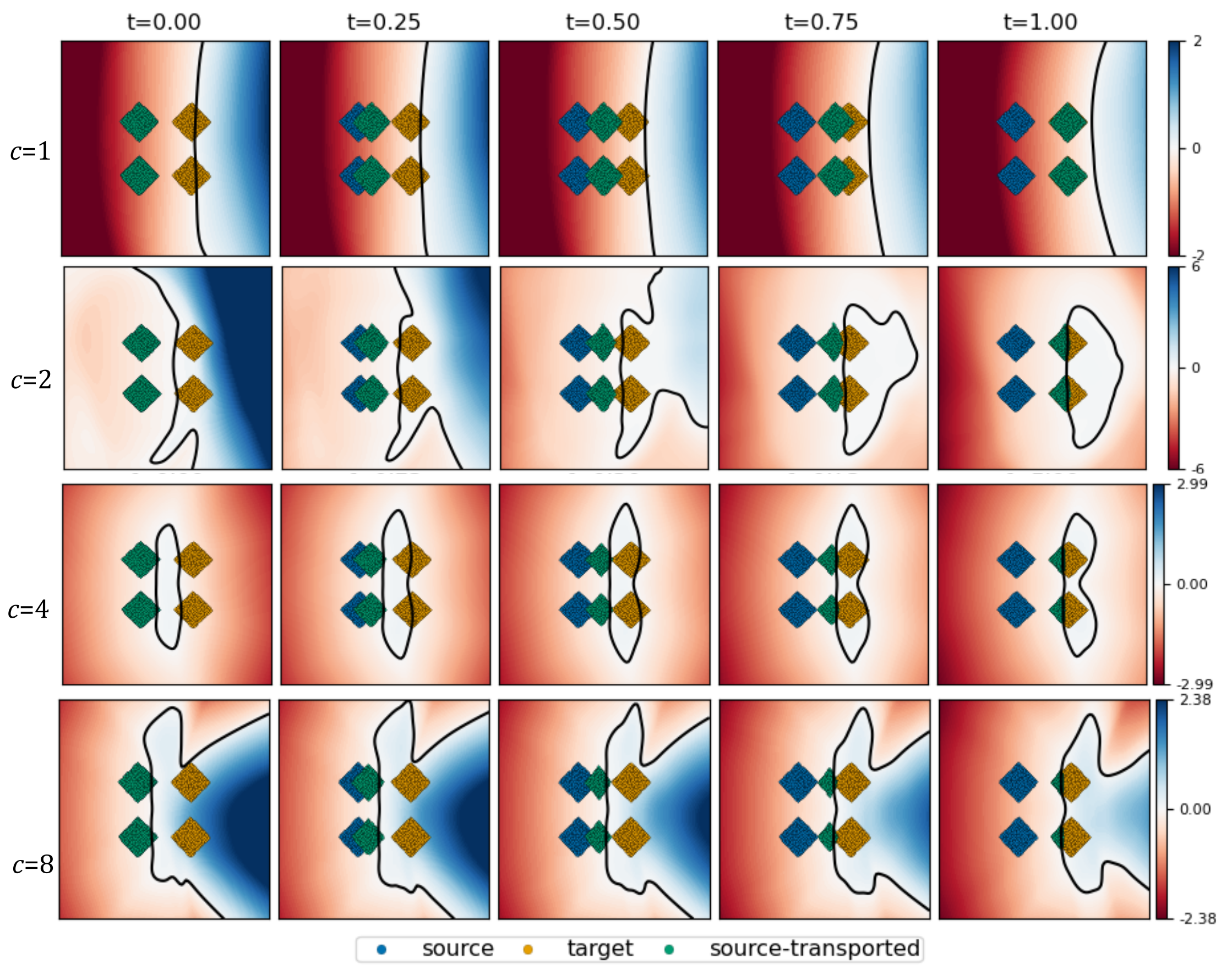}
        \caption{}
        \label{diamonds-evolution-all}
    \end{subfigure}\hspace{0mm}%
    \begin{subfigure}[t]{0.42\textwidth}
        \centering
        \includegraphics[width=\linewidth,clip=true,trim={0 0 0 0}]{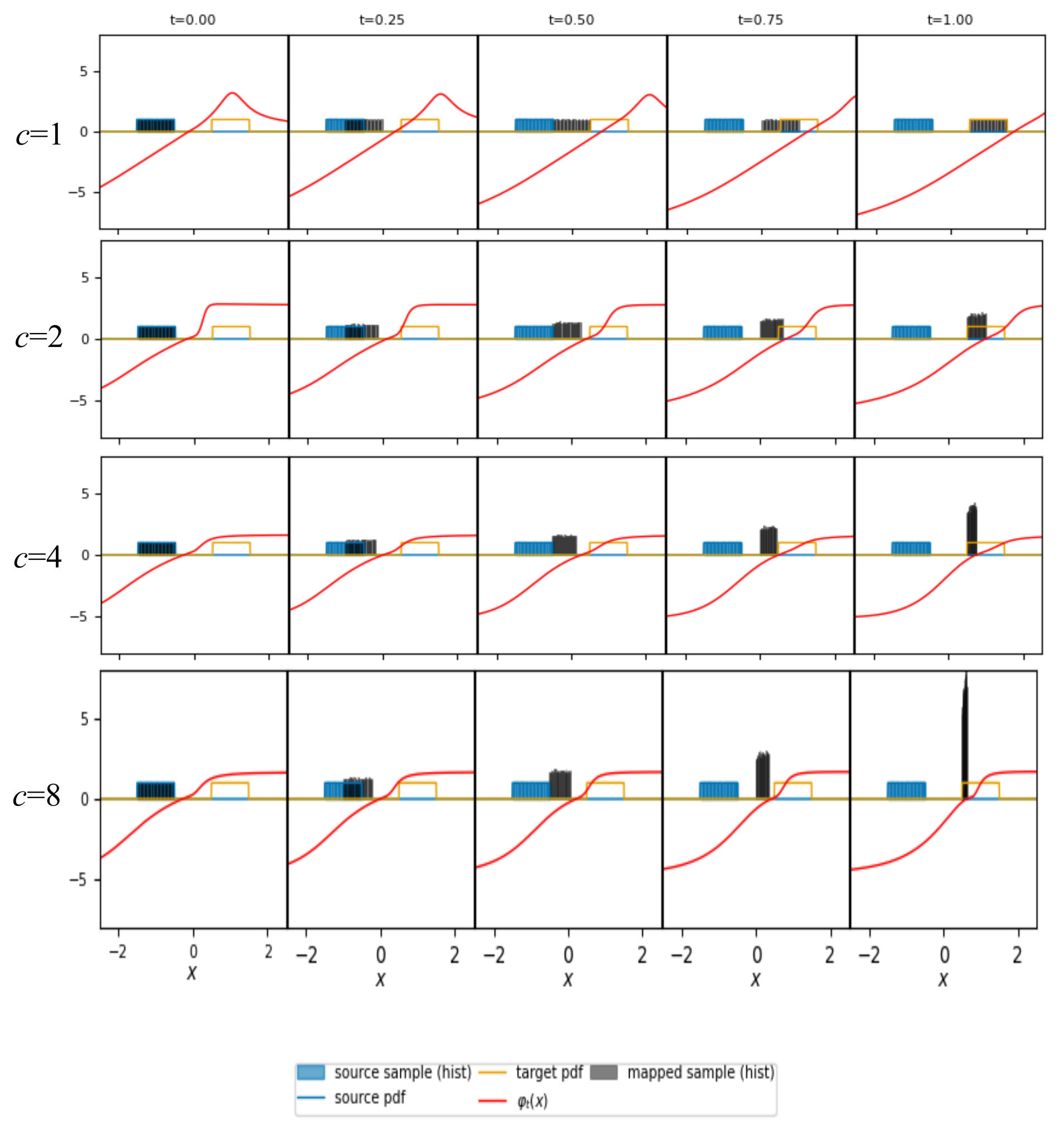}
        \caption{}
        \label{uniform2uniform-evolution-all}
    \end{subfigure}\hspace{0mm}%
    
    \caption{Temporal snapshots at $t\in \{0.0, 0.25, 0.50, 0.75, 1.0\}$ for $c\in{1,2,4,8}$. {(a)} Evolution between two 2D bimodal distributions; the black contour line indicates the zero-level set of $\varphi_{t}(\rvx)$. {(b)} Evolution from 1D uniform to another 1D uniform density, with dynamic subset subset alignment. $\varphi_{t}(\rvx)$ is denoted by red curve.}
    \label{fig_toy-add2}
\end{figure}

\begin{figure*}[htb]
    \centering
    \begin{subfigure}[t]{1.0\textwidth}
        \centering
    \includegraphics[width=\linewidth,clip=true,trim={0 0 0 0}]{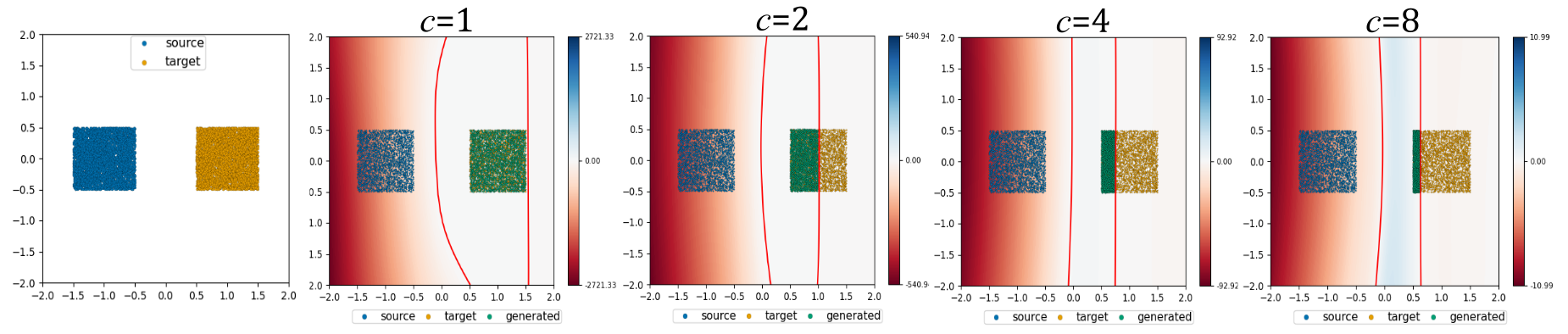}
        \caption{}
        \label{static-squares-all}
    \end{subfigure}
    \begin{subfigure}[t]{1.0\textwidth}
        \centering
    \includegraphics[width=\linewidth,clip=true,trim={0 0 0 0}]{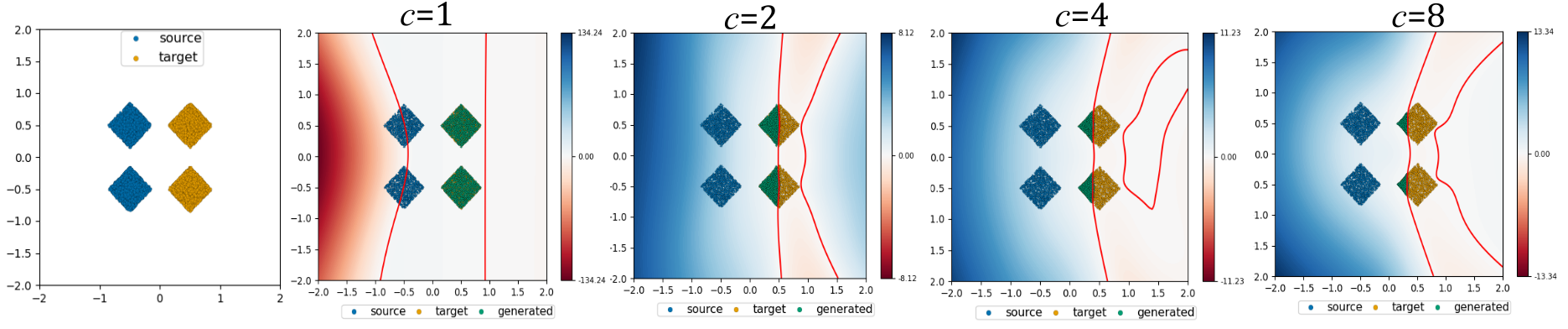}
        \caption{}
        \label{static-diamonds-all}
    \end{subfigure}\hspace{0mm}%
    \caption{Static subset alignment results for $c\in\{1,2,4,8\}$. {(a)} Translation between two 2D uniform distributions shown in Figure~\ref{squares-final-all}. {(b)} 2D bimodal distributions shown in Figure~\ref{diamonds-evolution-all}. Red curves indicate the zero-level sets of $\eta(\rvx)$.}
    \label{fig_toy-add3}
\end{figure*}

For the toy results given in Figures \ref{squares-final-all} and \ref{uniform2semicircle-final-all} the temporal snapshots of the evolution of points sampled from source along with the potential $\varphi_{t}$ are given in Figures~\ref{squares-evolution-all} and~\ref{uniform2semicircle-evolution-all}, respectively. Similar temporal snapshots between two 2D bimodal uniform squares and two 1D uniform distributions are shown in Figure~\ref{diamonds-evolution-all} and~\ref{uniform2uniform-evolution-all} respectively.

The static framework for subset alignment is equivalent to the relaxations of the inexact optimal transport problem discussed by \citet{gazdieva2023extremal}. Our 
approach for static subset alignment is discussed in Algorithm~\ref{neural-static} and Appendix \ref{Static support subset alignment}. 

\begin{figure*}[htb]
    \centering
    \begin{subfigure}[t]{1.0\textwidth}
        \centering
    \includegraphics[width=0.8\linewidth,clip=true,trim={0 0 0 0}]{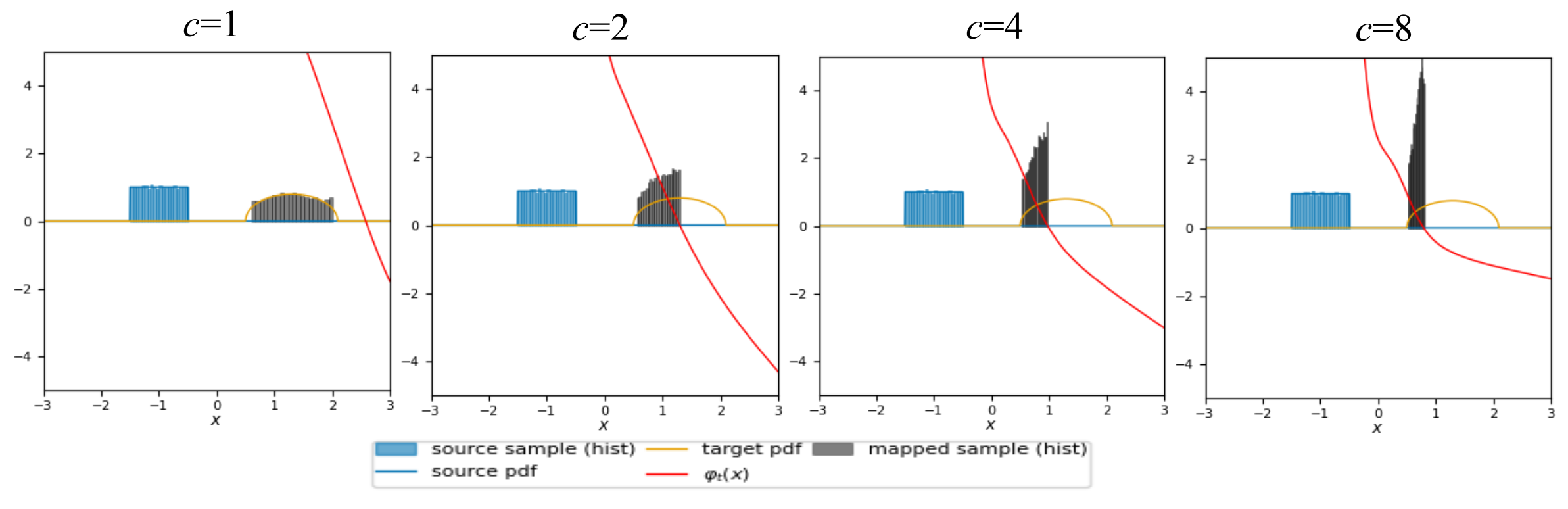}
        \caption{}
        \label{static-uniform2semicircle-final-all}
    \end{subfigure}
    \begin{subfigure}[t]{1.0\textwidth}
        \centering
    \includegraphics[width=0.8\linewidth,clip=true,trim={0 0 0 0}]{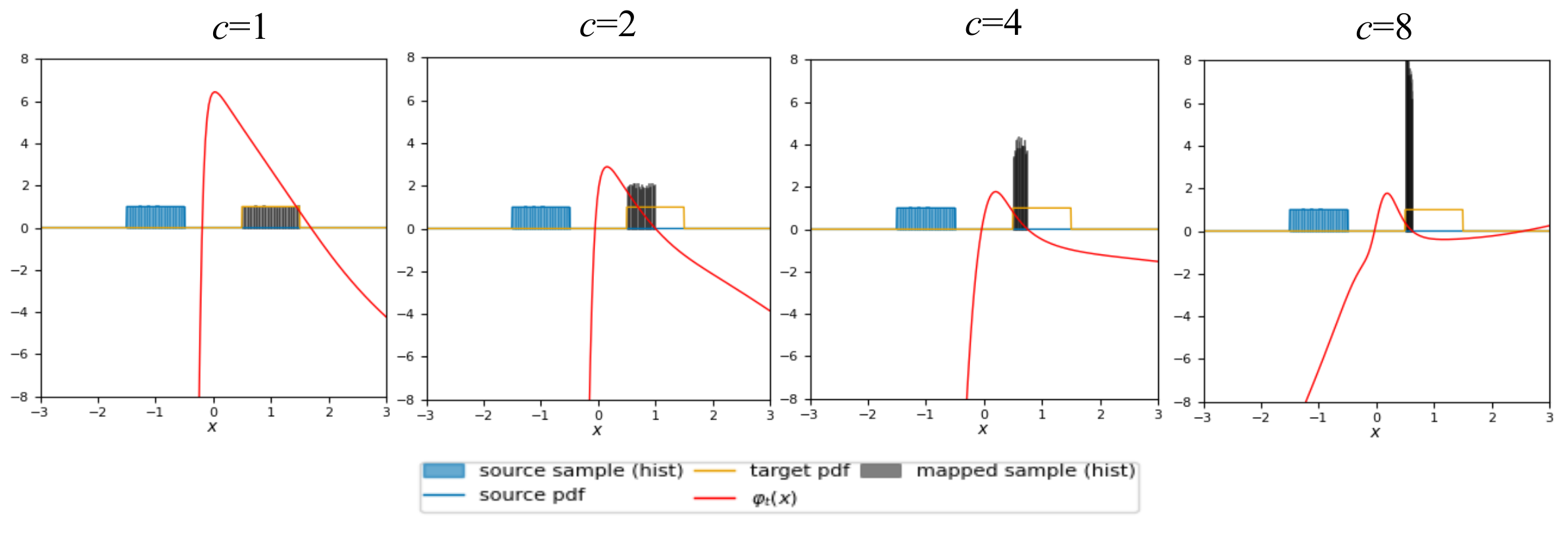}
        \caption{}
        \label{static-uniform2uniform-final-all}
    \end{subfigure}\hspace{0mm}%
    \caption{Static subset alignment results for $c\in\{1,2,4,8\}$. {(a)} Transport from 1D uniform distribution to semicircular density shown in Figure~\ref{uniform2semicircle-evolution-all}. {(b)} Transport from 1D uniform distribution to a uniform distribution shown in Figure~\ref{uniform2uniform-evolution-all}. Red curves indicate the  $\eta(\rvx)$.}
    \label{fig_toy-add4}
\end{figure*}
\section{Implementation Details}
\label{sec:implementation-details}
\subsection{Toy Results} 
\label{app:toy-results}
For dynamic subset alignment, we use the model architectures in Listings~\ref{lst:toy_model_architectures_1d} and~\ref{lst:toy_model_architectures_2d}, where \texttt{Smodel} and \texttt{Qmodel} parameterize $\theta_{\varphi}$ and $\theta_{\rho}$, respectively. Listing~\ref{lst:toy_model_architectures_1d} is used for the 1D toy experiments, while Listing~\ref{lst:toy_model_architectures_2d} is used for the 2D toy experiments. We set \texttt{num\_hid}$=512$ for all dynamic architectures. For both settings, we use Algorithm~\ref{neural-dynamic_0} with $n_{\varphi}=n_{\rho}=1$. We train for 50,000 iterations in 1D and 100,000 iterations in 2D. In all dynamic toy experiments, we use the Adam optimizer~\citep{kingma2014adam} with a learning rate of $1\times10^{-4}$ and a batch size of 512.

For static subset alignment, we use the model architecture in Listing~\ref{lst:toy_static_model_architectures} for both the 1D and 2D toy experiments. The \texttt{Tmodel} parameterizes the transport map $T$ via a residual connection, while \texttt{DualNET} parameterizes the dual function $\eta$. The same architecture is used in both settings, with the input and output dimensions adjusted accordingly. We set \texttt{num\_hid}$=512$ for all static architectures. For both settings, we use Algorithm~\ref{neural-static} with $n_T=5$ and $n_{\eta}=1$. We train for 20,000 iterations in 1D and 50,000 iterations in 2D. In all static toy experiments, we use the Adam optimizer~\citep{kingma2014adam} with a learning rate of $1\times10^{-4}$ and a batch size of 512.

\begin{lstlisting}[language=python,
style=pyflax,
float,
floatplacement=!ht,
caption={Model architectures used for toy 1D dynamic subset alignment experiments.},
label={lst:toy_model_architectures_1d}]
import jax
from jax import numpy as jnp
from flax import linen as nn
import math
from ml_collections import config_dict

class Smodel(nn.Module):
    num_hid: int
    num_out: int
    use_layernorm: bool = True
    boundary_zero: bool = False

    @nn.compact
    def __call__(self, t, x):
        if jnp.ndim(t) == 0:
            t = jnp.broadcast_to(t, x.shape[:-1] + (1,))
        elif t.shape[-1] != 1:
            t = t.reshape(x.shape[:-1] + (1,))

        h = jnp.concatenate([t, x], axis=-1)
        for _ in range(4):
            h = nn.Dense(self.num_hid)(h)
            if self.use_layernorm:
                h = nn.LayerNorm()(h)
            h = nn.swish(h)

        v = nn.Dense(self.num_out)(h)
        return t * (1.0 - t) * v if self.boundary_zero else v

class Qmodel(nn.Module):
    num_hid: int
    num_out: int

    @nn.compact
    def __call__(self, t, x_0, x_1):
        h = jnp.concatenate([t, x_0, x_1, t < 0.5], axis=-1)
        for _ in range(3):
            h = nn.swish(nn.Dense(self.num_hid)(h))

        h = nn.Dense(self.num_out)(h)
        return (1.0 - t) * x_0 + t * (x_1 + h)
\end{lstlisting}
\begin{lstlisting}[language=python,
style=pyflax,
float,
floatplacement=!ht,
caption={Model architectures used for toy 2D dynamic subset alignment experiments},
label={lst:toy_model_architectures_2d}]
import jax
from jax import numpy as jnp
from flax import linen as nn
import math
from ml_collections import config_dict

class Smodel(nn.Module):
    num_hid: int
    num_out: int

    @nn.compact
    def __call__(self, t, x):
        if jnp.ndim(t) == 0:
            t = jnp.broadcast_to(t, x.shape[:-1] + (1,))
        h = jnp.concatenate([t, x], axis=-1)
        for _ in range(3):
            h = nn.swish(nn.Dense(self.num_hid)(h))
        return nn.Dense(self.num_out)(h)

class Qmodel(nn.Module):
    num_hid: int
    num_out: int

    @nn.compact
    def __call__(self, t, x_0, x_1):
        h = jnp.concatenate([t, x_0, x_1, t < 0.5], axis=-1)
        for _ in range(3):
            h = nn.swish(nn.Dense(self.num_hid)(h))
        h = nn.Dense(self.num_out)(h)
        return (1.0 - t) * x_0 + t * (x_1 + h)
\end{lstlisting}
\begin{lstlisting}[language=python,
style=pyflax,
float,
floatplacement=!ht,
caption={Model architectures used for 1D and 2D static subset alignment toy experiments.},
label={lst:toy_static_model_architectures}]
import jax
from jax import numpy as jnp
from flax import linen as nn
import math
from ml_collections import config_dict

class Tmodel(nn.Module):
    num_hid: int
    num_out: int

    @nn.compact
    def __call__(self, inputs):
        h = inputs
        for _ in range(4):
            h = nn.swish(nn.Dense(self.num_hid)(h))
        h = nn.Dense(self.num_out)(h)
        return inputs + h
        
class DualNET(nn.Module):
    num_hid: int
    num_out: int

    @nn.compact
    def __call__(self, x):
        h = x
        for _ in range(4):
            h = nn.swish(nn.Dense(self.num_hid)(h))
        return nn.Dense(self.num_out)(h)
\end{lstlisting}

\subsection{EMNIST Classifier} 
\label{app:classifier}
We merge all letters into a single class and treat each digit as its own class (digits 0--9 keep their numeric label, and all letters are labeled 10). To mitigate class imbalance during training, we use a class-reweighted SoftMax loss. For $k$-class classification, let $\vz\in\mathbb{R}^k$ denote the per-class counts in the training set; we define reweighting coefficients $\bm{\omega}\in\mathbb{R}^k$ by
\begin{equation}
    {\omega}_i = \left(\sum_{j=1}^{N} \frac{z_{i}}{z_j} \right)^{-1}, \ \forall \ i \in [k].
\end{equation}
Given a one-hot label vector $\rvy$ and the network SoftMax output $\hat{\rvy}$, the reweighted loss (risk) is
\begin{equation}
    \ell(\rvy, \hat\rvy) = \mathbf{1}_{k}^{\top}(\bm{\omega} \odot \rvy \odot \hat{\rvy}) \label{reweighted-loss}
\end{equation}

We train the EMNIST classifier using the train/validation split provided with the EMNIST dataset \citep{cohen2017emnist}. The model is ResNet-18 \citep{he2016deep} and is optimized with the class-reweighted SoftMax loss in \eqref{reweighted-loss}. We use Adam \citep{kingma2014adam} with a warmup--cosine learning-rate schedule \citep{loshchilov2017sgdr}, a peak learning rate of $1\times10^{-3}$, and 500 warmup steps. The cosine schedule decays over $20{,}000$ steps to an end learning rate of $1\times10^{-5}$. Training is stopped after $20{,}000$ steps, once the classifier exceeds 90\% overall validation accuracy and 99\% accuracy on digits. Confusion matrices are reported in Appendix~\ref{appendix-mnist-emnist}.

\subsection{MNIST-EMNIST Translation Models} 
\label{app:emnist}

For dynamic subset selection, following \citet{neklyudov2023action}, we parameterize the vector field $\varphi_t$ with a U-Net equipped with DDPM-style time embeddings \citep{song2020improved}. As in action matching \citep{neklyudov2023action}, $\varphi_t$ returns a scalar via
\begin{equation}
\varphi_t(\rvx)=\langle \mathrm{UNet}(x),x\rangle.
\end{equation}
Likewise, $Q_t$, which parameterizes $\rho_t$, is implemented as a U-Net with time embeddings. For both U-Nets, we use the same architectural settings: \texttt{nonlinearity = swish}, \texttt{nf = 64}, \texttt{ch\_mult = (1, 2, 2, 2)}, \texttt{num\_res\_blocks = 2}, and \texttt{attn\_resolutions = (16, 8)}. We use the AdamW optimizer with learning-rate scheduling for $50{,}000$ training iterations. The optimizer parameters are $\beta=(0,0.999)$, \texttt{weight decay}$=0.1$, and \texttt{dropout}$=0.1$. We additionally employ exponential moving averages (EMA) during training with an EMA rate of $0.999$. These settings are very similar to those used in rectified flow matching and action matching \citep{liu2023flow,neklyudov2023action}. All input images in neural subset alignment are centered at $[0.5,0.5]$ and resized to $32\times32\times1$. The learning rate is linearly increased from zero to its maximum value during the first $5{,}000$ iterations and is then kept constant. The maximum learning rates are $2\times10^{-4}$ and $1\times10^{-4}$ for $\varphi_t$ and $Q_t$, respectively. Additionally, gradients are clipped element-wise to the interval $[-1,1]$. Algorithm \ref{neural-dynamic_0} is employed for $50{,}000$ training iterations with $n_\varphi=2$ and $n_\rho=1$. A batch size of $128$ is used to train the dynamic models for MNIST-EMNIST translation.

For static domain translation, the transport network $T$ is a U-Net \cite{ronneberger2015u} with base factor 64, and the critic network $\eta$ is a ResNet; we use the same choices as \citet{gazdieva2023extremal} for inexact extremal optimal transport. All inputs to the static models are centered at $[0.5,0.5]$ and resized to $32\times32\times1$. We train $T$ and $\eta$ with Adam \cite{kingma2014adam} using an initial learning rate of $1\times10^{-4}$, which is halved after $10{,}000+5{,}000c$, $20{,}000+5{,}000c$, $30{,}000+5{,}000c$, $40{,}000+5{,}000c$, and $70{,}000+5{,}000c$ steps. We run Algorithm~\ref{neural-static} for $50{,}000$ learning iterations with $n_T=10$ and $n_{\eta}=1$. A batch size of 128 is used for static MNIST--EMNIST translation.

\subsection{Models for PU-Learning Using Subset Alignment}
\label{app:pu-learning-models}

The architectures used for PU learning with dynamic subset alignment are given in Listing~\ref{lst:Model architectures for dynamic subset alignment}, where \texttt{Smodel} and \texttt{Qmodel} parameterize $\varphi_t$ and ${\rho_t}$, respectively. For both models, we use $\texttt{num\_hid}=1024$ and $\texttt{embedding\_dim}=64$. The output dimension is $\texttt{num\_out}=1$ for \texttt{Smodel}, while for \texttt{Qmodel} it equals the dataset dimension (Table~\ref{tab:PU-learning-ds}). We run Algorithm~\ref{neural-dynamic_0} for $50{,}000$ training iterations with $n_{\varphi}=2$ and $n_{\rho}=1$. Both models are optimized with AdamW using learning rate $5\times10^{-6}$, $\beta_1=0.9$, and $\epsilon=10^{-8}$. We use $10{,}000$ warmup steps and gradient clipping with maximum norm $0.1$, and we maintain an exponential moving average (EMA) with rate $0.999$. Batch sizes are dataset-dependent and are reported in Table~\ref{tab:PU-learning-ds}.

The architectures used for PU learning with static subset alignment are given in Listing~\ref{lst:Model architectures for PU learning with static subset alignment}, where \texttt{Tmodel} and \texttt{DualNET} parameterize $T$ and ${\eta}$, respectively. For both models, we use $\texttt{num\_hid}=1024$ and $\texttt{embedding\_dim}=128$. The output dimension for \texttt{Tmodel} matches the dataset dimension, while \texttt{DualNET} uses $\texttt{num\_out}=1$. We run Algorithm~\ref{neural-static} for $20{,}000$ training iterations with $n_T=10$ and $n_{\eta}=1$. Both models are optimized with learning rate $10^{-4}$ and use EMA with rate $0.999$. Batch sizes are dataset-dependent and are reported in Table~\ref{tab:PU-learning-ds}. For all PU-learning experiments, we fix $c=\frac{1}{\pi_{+}}$.

\begin{table}[h]
\centering
\caption{UCI datasets for PU learning, along with the total number of data points (n), dimension (dim), positive prior ($\pi$) and batch sizes employed in training the corresponding dynamic models. }
\label{tab:PU-learning-ds}
\begin{tabular}{l|llll}
Dataset     & data-set size & dimension & $\pi$   & batch size  \\ 
\hline
Abalone     & 4177             & 8   & 0.16 & 20           \\
Banknote    & 1372             & 4   & 0.44 & 10           \\
Breast-w    & 699              & 9   & 0.34 & 10           \\
Diabetes    & 768              & 8   & 0.35 & 6            \\
Haberman    & 306              & 3   & 0.26 & 6            \\
Heart       & 270              & 13  & 0.44 & 6            \\
Ionosphere  & 351              & 34  & 0.64 & 6            \\
Isolet      & 7797             & 617 & 0.04 & 4            \\
Jm1         & 10885            & 21  & 0.19 & 20           \\
Kc1         & 2109             & 21  & 0.15 & 20           \\
Madelon     & 2600             & 500 & 0.5  & 20           \\
Musk        & 6598             & 166 & 0.15 & 20           \\
Segment     & 2310             & 19  & 0.14 & 20           \\
Semeion     & 1593             & 256 & 0.1  & 4            \\
Sonar       & 208              & 60  & 0.53 & 4            \\
Spambase & 4601             & 57  & 0.39 & 20           \\
Vehicle     & 846              & 18  & 0.26 & 6            \\
Waveform    & 5000             & 40  & 0.34 & 20           \\
Wdbc        & 569              & 30  & 0.37 & 6            \\
Yeast       & 1484             & 8   & 0.31 & 10         
\end{tabular}
\end{table}

\begin{figure}
\begin{lstlisting}[language=python,
style=pyflax,
caption={Model architectures for PU-Learning for dynamic subset alignment},
label={lst:Model architectures for dynamic subset alignment}]
import jax
from jax import numpy as jnp
from flax import linen as nn
import math
from ml_collections import config_dict

def get_timestep_embedding(timesteps, embedding_dim: int,  max_positions: float = 10_000.0):
  t = jnp.asarray(timesteps)
  t = t.reshape((-1,)) if t.ndim == 1 else t.reshape((t.shape[0], -1))[:, 0]
  t = t.astype(jnp.float32)

  half = embedding_dim // 2
  if half < 1:
    return jnp.zeros((t.shape[0], embedding_dim), dtype=t.dtype)

  freqs = jnp.exp(-jnp.log(jnp.array(max_positions,dtype=t.dtype))
      * jnp.arange(half, dtype=t.dtype) / (half - 1 + 1e-8))  

  args = t[:, None] * freqs[None, :]          # (B, half)
  emb = jnp.concatenate([jnp.cos(args), jnp.sin(args)], axis=-1)  # (B, 2*half)

  if embedding_dim % 2 == 1:
    emb = jnp.pad(emb, ((0, 0), (0, 1)))

  return emb[:, :embedding_dim]
  
class Smodel(nn.Module):
  config: config_dict.ConfigDict
  @nn.compact
  def __call__(self, t, x, *, train: bool = True):
    config = self.config
    time_emb_dim = config.embedding_dim
    num_hid = config.num_hid
    num_out = config.num_out
    x = jnp.asarray(x)
    if x.ndim == 1:                 # (D,) -> (1,D)
      x = x[None, :]

    # make t broadcastable to x batch: (B,1)
    t = jnp.asarray(t)
    t = jnp.broadcast_to(t, x.shape[:-1] + (1,))

    temb = get_timestep_embedding(t, time_emb_dim)  # should return (B, time_emb_dim)
    temb = nn.swish(nn.Dense(num_hid)(temb))
    temb = nn.Dense(num_hid)(temb)          # (B, num_hid)

    h = jnp.concatenate([t, x], axis=-1)
    h = nn.swish(nn.Dense(num_hid)(h))
    h = h + temb

    for _ in range(3):
      h = nn.swish(nn.Dense(num_hid)(h))

    return nn.Dense(num_out)(h)

class Qmodel(nn.Module):
  config: config_dict.ConfigDict

  @nn.compact
  def __call__(self, t, x_0, x_1, *, train: bool = True):
    config = self.config
    time_emb_dim = config.embedding_dim
    num_hid = config.num_hid
    num_out = config.num_out

    x_0 = jnp.asarray(x_0)
    x_1 = jnp.asarray(x_1)
    if x_0.ndim == 1:                 # (D,) -> (1,D)
      x_0 = x_0[None, :]
      x_1 = x_1[None, :]

    # broadcast t to (B,1) to match x batch
    t = jnp.asarray(t)
    t = jnp.broadcast_to(t, x_0.shape[:-1] + (1,))     # (B,1)

    # time embedding -> project to hidden width
    temb = get_timestep_embedding(t, time_emb_dim)     # (B, time_emb_dim)
    temb = nn.swish(nn.Dense(num_hid)(temb))
    temb = nn.Dense(num_hid)(temb)                     # (B, num_hid)

    heaviside_step = (t < 0.5).astype(x_0.dtype)       # (B,1)
    h = jnp.concatenate([t, x_0, x_1, heaviside_step], axis=-1)

    h = nn.swish(nn.Dense(num_hid)(h))
    h = h + temb

    for _ in range(3):
      h = nn.swish(nn.Dense(num_hid)(h))

    h = nn.Dense(num_out)(h)

    x_t = (1.0 - t) * x_0 + t * (x_1 + h)
    return x_t
\end{lstlisting}
\end{figure}

\begin{figure}
\begin{lstlisting}[style=pyflax, language=Python,
caption={Model architectures for PU learning with static subset alignment},
label={lst:Model architectures for PU learning with static subset alignment}]
import jax
from jax import numpy as jnp
from flax import linen as nn
import math
from ml_collections import config_dict

class etamodel(nn.Module):
    num_hid: int
    num_out: int
    @nn.compact
    def __call__(self, x):
        h = x
        for _ in range(3):
            h = nn.swish(nn.Dense(self.num_hid)(h))
        return nn.Dense(self.num_out)(h)

class Tmodel(nn.Module):
    num_hid: int
    num_out: int
    @nn.compact
    def __call__(self, x):
        h = x
        for _ in range(4):
            h = nn.swish(nn.Dense(self.num_hid)(h))
        h = nn.Dense(self.num_out)(h)
        residual = nn.Dense(self.num_out)(x)
        return h + residual
\end{lstlisting}
\end{figure}
\subsection{Image-to-Image Translation on FFHQ}
\label{app:ffhq}

For dynamic subset alignment, the architectures for the networks parameterizing $\varphi_t$ and $\rho_t$ are given in Listing~\ref{lst:ffhq-dynamic-model-architectures}, where \texttt{Smodel} and \texttt{Qmodel} parameterize $\varphi_t$ and $\rho_t$ respectively . Both networks use \texttt{t\_embed\_dim}$=128$ and \texttt{num\_hid}$=1024$, with \texttt{num\_out}$=1$ for the $\varphi_t$ network and \texttt{num\_out}$=512$ for the $\rho_t$ network (matching the FFHQ latent dimensionality). We use AdamW with a fixed learning rate of $1\times10^{-5}$, gradient clipping with maximum norm $1.0$, EMA rate $0.999$, and batch size 128. We run Algorithm~\ref{neural-dynamic_0} for $500{,}000$ learning iterations with $n_{\varphi}=2$ and $n_{\rho}=1$.

For static subset alignment, the neural network architectures \texttt{TransportNET} and \texttt{DualNET} given in Listing~\ref{lst:ffhq-static-model-architectures} parameterize $T$ and $\eta$, respectively. Both networks use \texttt{hidden\_dim}$=1024$. \texttt{TransportNET} maps 512-dimensional FFHQ latents to 512-dimensional outputs, while \texttt{DualNET} returns a scalar. We use AdamW with a fixed learning rate of $1\times10^{-5}$, weight decay $10^{-8}$, EMA rate $0.999$, and batch size $1{,}024$. We run Algorithm~\ref{neural-static} for $50{,}000$ learning iterations with $n_T=5$ and $n_{\eta}=1$. The same architectures are used for the UOTM experiments in Table~\ref{ffhq_class_table}.

\begin{lstlisting}[language=python,
style=pyflax,
float,
floatplacement=!ht,
caption={Model architectures for FFHQ dynamic subset alignment.},
label={lst:ffhq-dynamic-model-architectures}]
import jax
from jax import numpy as jnp
from flax import linen as nn
import math

def get_timestep_embedding(t, dim, max_positions=10000):
    t = jnp.reshape(t, (-1,))
    half = dim // 2
    scale = jnp.exp(
        jnp.arange(half, dtype=jnp.float32)
        * (-math.log(max_positions) / (half - 1))
    )
    emb = t[:, None] * scale[None, :]
    emb = jnp.concatenate([jnp.sin(emb), jnp.cos(emb)], axis=1)
    return jnp.pad(emb, [(0, 0), (0, dim % 2)])

class Smodel(nn.Module):
    num_hid: int
    num_out: int
    t_embed_dim: int

    @nn.compact
    def __call__(self, t, x):
        if t.ndim == 0:
            t = jnp.broadcast_to(t, (x.shape[0], 1, 1))

        e = get_timestep_embedding(t, self.t_embed_dim)
        e = jnp.reshape(e, x.shape[:-1] + (self.t_embed_dim,))

        h = jnp.concatenate([e, x], axis=-1)
        for i in range(4):
            h = nn.swish(nn.Dense(self.num_hid)(h))
            if i < 3:
                h = jnp.concatenate([e, h], axis=-1)

        return nn.Dense(self.num_out)(h)
        
class Qmodel(nn.Module):
    num_hid: int
    num_out: int
    t_embed_dim: int

    @nn.compact
    def __call__(self, t, x_0, x_1):
        e = get_timestep_embedding(t.squeeze(), self.t_embed_dim)
        h = jnp.concatenate([e, x_0, x_1, t < 0.5], axis=-1)

        for i in range(4):
            h = nn.swish(nn.Dense(self.num_hid)(h))
            if i < 3:
                h = jnp.concatenate([e, h], axis=-1)

        h = nn.Dense(self.num_out)(h)
        return (1.0 - t) * x_0 + t * (x_1 + h)
\end{lstlisting}

\begin{lstlisting}[language=python,
style=pyflax,
float,
floatplacement=!ht,
caption={Model architectures for FFHQ static subset alignment.},
label={lst:ffhq-static-model-architectures}]
import jax
from jax import numpy as jnp
from flax import linen as nn
import math

class TransportNET(nn.Module):
    input_dim: int
    hidden_dim: int
    @nn.compact
    def __call__(self, x):
        h = x
        for _ in range(3):
            h = nn.swish(nn.Dense(self.hidden_dim)(h))
        return nn.Dense(self.input_dim)(h) + nn.Dense(self.input_dim)(x)

class DualNET(nn.Module):
    hidden_dim: int
    @nn.compact
    def __call__(self, x):
        h = x
        for _ in range(3):
            h = nn.swish(nn.Dense(self.hidden_dim)(h))
        return nn.Dense(1)(h)
\end{lstlisting}

\subsection{Melanoma Experiment}
\label{app: Melanoma Experiment}
Unlike the other experiments in this paper, the melanoma setting involves multiple intermediate marginals that must be matched. Because we want to preserve all marginals except the terminal one, we generalize \eqref{eq:dynamic-interp} to the multi-marginal setting. Assuming we can sample from the marginals at times $t_k$ and $t_{k+1}$, for $t\in[t_k,t_{k+1}]$ we define $\tau=\frac{t-t_k}{t_{k+1}-t_k}$ and use the interpolant
  \begin{equation} \label{melanoma_interpolant}
    \rvx_t = (1-\tau)\,\rvx_{t_k} + \tau\,\rvx_{t_{k+1}}
           + w_k(\tau)\, Q\big(t,\rvx_{t_k},\rvx_{t_{k+1}};\theta_\rho\big),
  \end{equation}
  where
  \begin{equation}
      w_k(\tau) =
    \begin{cases}
      1-\tau^2-(1-\tau)^2, & k<K-1,\\[2pt]
      1-(1-\tau)^2,        & k=K-1,
    \end{cases}
  \end{equation}
  where $Q$ is a neural network shared across segments. Since $w_k$ vanishes at
  $\tau=0$ for all $k$ and at $\tau=1$ for $k<K-1$, the interpolant matches
  $\mu_0,\ldots,\mu_{K-1}$ exactly, while $w_{K-1}(1)=1$ leaves the terminal
  marginal unconstrained: $\rvx_{t_K}\mapsto\rvx_{t_K}+Q(t_K,\rvx_{t_{K-1}},\rvx_{t_K};\theta_\rho)$.
  For $K=1$, this reduces to $(1-t)\rvx_0 + t\rvx_1 + t(2-t)\,Q$, which, like
  $(1-t)\rvx_0 + t(\rvx_1+Q)$, fixes $\mu_0$ while relaxing the terminal marginal.
  In addition, we use the Wasserstein gradient-flow updates of \citet{neklyudov24a} to improve stability and convergence. 
  
  The Algorithm~\ref{neural-dynamic_modified} summarizes the training procedure for the melanoma experiment. The Neural Network Architectures employed in Melanoma experiments are provided in Listing \ref{lst:Melanoma-model-architectures}. We should emphasize that the interpolation equation \ref{melanoma_interpolant} has been implemented as part of interpolant neural network architecture, but for clear description of learning procedure, we included interpolation equation a part of learning algorithm. 
  
  We use the first $50$ principal components of the melanoma data \citep{bunne2023learning}, centered and divided by the largest per-dimension standard deviation of the training set. The three observed populations are ordered in reverse time, $\mu_0$ (48h), $\mu_1$ (24h) and $\mu_2$ (8h), at normalised times $t_0=0$, $t_1=0.6$ and $t_2=1$ ($K=2$), so that the sub-setting constraint acts on the $\mu_{K=2}$. The potential $\varphi_t(\rvx;\theta_\varphi)=\langle f(t,\rvx;\theta_\varphi),\rvx\rangle$ is parameterized by a \texttt{Smodel} in Listing \ref{lst:Melanoma-model-architectures} with $\mathtt{nf}=512$ and activations $\mathtt{act} = \mathtt{Swish}$; the time $t$ is mapped by a two-layer embedding network of width $512$ and added to the input of each of the three inner hidden layers. The interpolant network $Q_t(\rvx_{t_k},\rvx_{t_{k+1}};\theta_\rho)$  is parameterized by \texttt{Qmodel} in Listing \ref{lst:Melanoma-model-architectures}, which takes concatenation $(t,\rvx_{t_k},\rvx_{t_{k+1}})$ as input and has a single hidden layer of width $512$ with swish activation and a linear output in $\mathbb{R}^{50}$. Both networks are trained with AdamW ($\beta_1=0.9$, $\beta_2=0.999$, weight decay $10^{-4}$) at learning rate $2\times10^{-4}$ with a linear warm-up over the first $5{,}000$ iterations and element-wise gradient clipping at $1$, using batch size $N=512$, $n_\varphi=n_\rho=1$ update per iteration, $M=10$ Wasserstein gradient steps with step size $\eta=10^{-2}$, and $200{,}000$ iterations; the times $t^i$ are drawn from a low-discrepancy sequence on $[0,1]$. We report results for sub-setting constants $c\in\{1.0,2.0, 4.0, 8.0\}$ in table \ref{tab:melanoma_results}. For evaluation we use an exponential moving average of the parameters with rate $0.999$, and transport cells with the learned velocity field $\nabla\varphi_t$ by integrating the ODE from $t_K$ back to $t_0$ with the explicit Euler scheme and step size $1\times10^{-3}$.

\begin{algorithm}[ht]
{
\caption{\texttt{Modified-Dynamic-Neural-SS}: Modified Learning Algorithm for Dynamic Subset Alignment on Melanoma }
\label{neural-dynamic_modified}
\footnotesize
\begin{algorithmic}[1]
\REQUIRE Samples from distributions $\{\mu_k\}_{k=0}^{K}$ observed at times $\{t_k\}_{k=0}^{K}$, time-dependent neural network
$\varphi_t(\cdot;\theta_{\varphi})$, interpolant network $\mathit{Q}_t(\cdot,\cdot;\theta_\rho)$, sub-setting constant $c\geq 1$, batch size $N$, number of updates $n_\varphi$ and $n_\rho$, gradient clipping parameter $\varepsilon$, 
number of Wasserstein gradient steps $M$ with step size $\eta$, gradient-based optimizers $\mathrm{optim}_\varphi$ and
$\mathrm{optim}_\rho$.
\ENSURE Sample-based neural estimate for $\varphi_t(\cdot;\theta_{\varphi})$
\FOR{learning iteration $\ell = 0,1,\ldots$}
  \FOR{$u = 1,\ldots,n_\varphi$}
    \STATE Sample $\{\rvx_{t_k}^i\}_{i=1}^N \sim \mu_k$ for $k=0,\ldots,K$, and $\{t^i\}_{i=1}^N \sim \mathrm{Uniform}([0,1])$.
    \STATE $
    \begin{aligned}
        &\Big\{{\rvx}_{t^i}^i \leftarrow (1-\tau^i)\,\rvx_{t_{k^i}}^i + \tau^i\,\rvx_{t_{k^i+1}}^i
        + w_{k^i}(\tau^i)\, Q_{t^i}\big(\rvx_{t_{k^i}}^i,\rvx_{t_{k^i+1}}^i;\theta_\rho\big) \Big\}_{i=1}^{N}, \\
        & \text{where } t^i\in[t_{k^i},t_{k^i+1}],\ \tau^i=\tfrac{t^i-t_{k^i}}{t_{k^i+1}-t_{k^i}},\\
        & w_k(\tau)=1-\tau^2-(1-\tau)^2 \text{ for } k<K-1, \text{ and } w_{K-1}(\tau)=1-(1-\tau)^2
    \end{aligned}$
    \FOR{$j = 1,\ldots,M$}
      \STATE $
      \begin{aligned}
          \Big\{{\rvx}_{t^i}^i \leftarrow {\rvx}_{t^i}^i + \eta\, w_{k^i}(\tau^i)\,\mathrm{clip}\Big(\nabla_{\rvx}\Big[\frac{\partial}{\partial t}\varphi_{t^i}({\rvx}_{t^i}^i;\theta_{\varphi})
          + \frac{\|\nabla \varphi_{t^i}({\rvx}_{t^i}^i;\theta_{\varphi})\|^2}{2}\Big],-\varepsilon, \varepsilon\Big) \Big\}_{i=1}^{N}
      \end{aligned}$
    \ENDFOR
    \STATE
    $\displaystyle
    \begin{aligned}
    \mathrm{grad}_{\theta_{\varphi}}
    & \leftarrow \nabla_{\theta_{\varphi}} \frac{1}{N}\sum_{i=1}^{N} \bigg[
      \frac{\partial}{\partial t}\varphi_{t^i}({\rvx}_{t^i}^i;\theta_{\varphi})
      + & \frac{\|\nabla \varphi_{t^i}({\rvx}_{t^i}^i;\theta_{\varphi})\|^2}{2} + \varphi_{t_0}(\rvx_{t_0}^{i};\theta_{\varphi}) \\
      & & +  c \cdot \max\!\Bigl(0,-\varphi_{t_K}(\rvx_{t_K}^{i};\theta_{\varphi})\Bigr)
    \bigg]
    \end{aligned}$

    \STATE Using $\mathrm{optim}_\varphi$ and $\mathrm{grad}_{\theta_{\varphi}}$ to update $\theta_{\varphi}$ via gradient descent.
  \ENDFOR

  \FOR{$u = 1,\ldots,n_\rho$}
    \STATE Sample $\{\rvx_{t_k}^i\}_{i=1}^N \sim \mu_k$ for $k=0,\ldots,K$, and $\{t^i\}_{i=1}^N \sim \mathrm{Uniform}([0,1])$.
    \STATE  $
    \begin{aligned}
        \Big\{{\rvx}_{t^i}^i(\theta_\rho) \leftarrow (1-\tau^i)\,\rvx_{t_{k^i}}^i + \tau^i\,\rvx_{t_{k^i+1}}^i
        + w_{k^i}(\tau^i)\, Q_{t^i}\big(\rvx_{t_{k^i}}^i,\rvx_{t_{k^i+1}}^i;\theta_\rho\big) \Big\}_{i=1}^{N}
    \end{aligned}$

    \STATE
    $\displaystyle
    \begin{aligned}
    \mathrm{grad}_{\theta_{\rho}}
    &\leftarrow \nabla_{\theta_{\rho}} \frac{1}{N}\sum_{i=1}^{N} \bigg[
      \frac{\partial}{\partial t}\varphi_{t^i}({\rvx}_{t^i}^i(\theta_{\rho});\theta_{\varphi}) + \frac{\|\nabla \varphi_{t^i}({\rvx}_{t^i}^i(\theta_{\rho});\theta_{\varphi})\|^2}{2}
    \bigg]
    \end{aligned}$

    \STATE Update $\theta_{\rho}$ using $\mathrm{optim}_\rho$ and $\mathrm{grad}_{\theta_{\rho}}$ via gradient ascent.
  \ENDFOR
\ENDFOR
\end{algorithmic}
}
\end{algorithm}

\begin{lstlisting}[language=python,
style=pyflax,
float,
floatplacement=!ht,
caption={Model architectures for Melanoma Experiment.},
label={lst:Melanoma-model-architectures}]
import jax
from jax import numpy as jnp
from flax import linen as nn
import math

class Smodel(nn.Module):
  config: ml_collections.ConfigDict

  @nn.compact
  def __call__(self, t: jnp.ndarray, x: jnp.ndarray, train: bool):
    config = self.config
    act = get_act(config)
    nf = config.nf

    if config.embed_time:
      temb = t
      temb = nn.Dense(nf)(temb)
      temb = nn.Dense(nf)(act(temb))
      h = x
    else:
      h = jnp.hstack([x, t])
    
    h = act(nn.Dense(nf)(h))
    for _ in range(config.n_layers):
      if config.embed_time:
        h += temb
      h = nn.Dropout(config.dropout)(h, deterministic=not train)
      if config.skip:
        h = act(nn.Dense(nf)(h)) + h
      else:
        h = act(nn.Dense(nf)(h))
    h = act(nn.Dense(nf)(h)) 
    h = nn.Dense(config.input_dim)(h)
    return (h*x).sum(1, keepdims=True)
        
class Qmodel(nn.Module):
  config: ml_collections.ConfigDict

  @nn.compact
  def __call__(self,t:jnp.ndarray,batch:jnp.ndarray,train:bool):
    config=self.config
    act=get_act(config)
    nf=config.nf

    timesteps,x=batch
    B=x.shape[0]
    n_marginals=timesteps.shape[1]

    t_reshaped=t.reshape(-1,1,1)
    t_right=(timesteps<t_reshaped).sum(1)
    t_right=jnp.fmax(t_right,jnp.ones_like(t_right).astype(int))
    t_left=t_right-1

    batch_idx=jnp.arange(B)
    x_left=x[batch_idx,t_left.ravel(),:]
    x_right=x[batch_idx,t_right.ravel(),:]
    t_0=timesteps[batch_idx,t_left.ravel(),:]
    t_1=timesteps[batch_idx,t_right.ravel(),:]

    temb=layers.get_timestep_embedding(t.ravel(),nf)
    temb=nn.Dense(nf)(temb)
    temb=nn.Dense(nf)(act(temb))

    h=jnp.concatenate([x_left,x_right],axis=1)
    h=jnp.hstack([t,h])

    if config.indicator:
      h=jnp.hstack([h,t<0.5*(t_0+t_1)])

    h=act(nn.Dense(nf)(h))

    for _ in range(config.n_layers):
      h=nn.Dropout(config.dropout)(h,deterministic=not train)
      h=act(nn.Dense(nf)(h))+h if config.skip else act(nn.Dense(nf)(h))

    h=nn.Dense(config.input_dim)(h)

    alpha=(t-t_0)/(t_1-t_0)
    beta=(t_1-t)/(t_1-t_0)

    w_inner=1.0-alpha**2-beta**2
    w_last=1.0-beta**2

    w=jnp.where(t_right==(n_marginals-1),w_last,w_inner)

    if config.get('pin_last',False):
      w=w_inner

    return beta*x_left+alpha*x_right+w*h
\end{lstlisting}

\section{Confusion matrices for MNIST to EMNIST domain translation}
\label{appendix-mnist-emnist}
\begin{figure}[h]
    \captionsetup{skip=0pt} 
    \centering
    \begin{subfigure}[t]{0.30\textwidth}
        \centering
        \includegraphics[width=\linewidth]{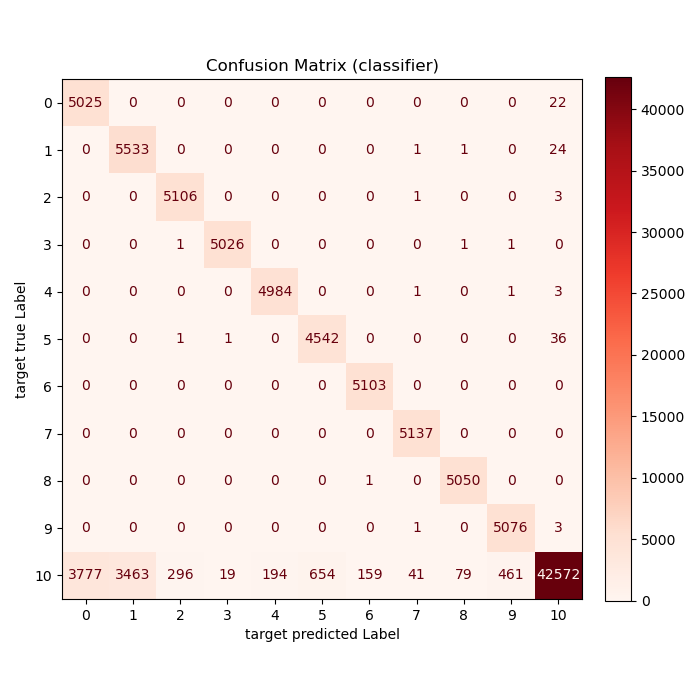}
        \caption{}
        \label{fig:emnist-cm-unnorm}
    \end{subfigure}\hspace{5pt}%
    \hspace{4pt}
    \begin{subfigure}[t]{0.30\textwidth}
        \centering
        \includegraphics[width=\linewidth]{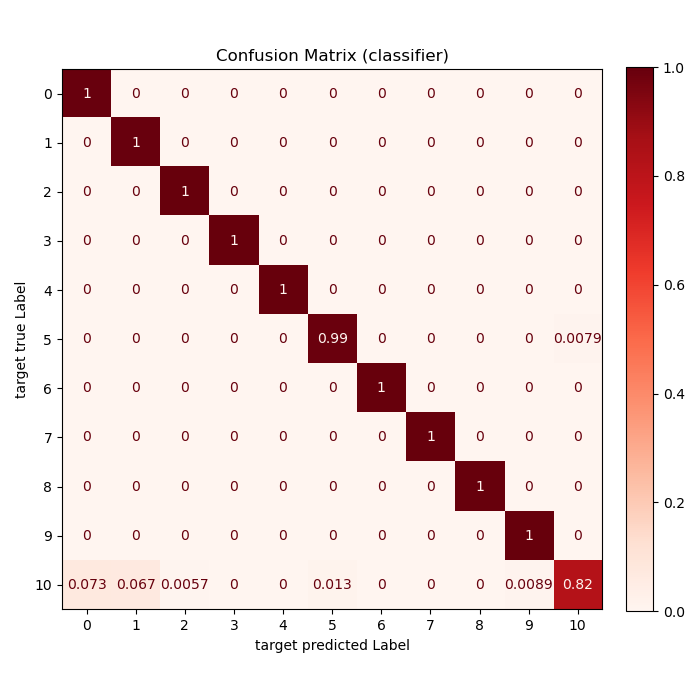}
        \caption{}
        \label{fig:emnist-cm-norm}
    \end{subfigure}

    \caption{(a) Normalized and (b) unnormalized confusion matrices for the EMNIST classifier discussed in Section~\ref{mnist_to_emnist}.}
    \label{emnist-classifier-confusion-matrix}
\end{figure}
\begin{figure}[h]
    \captionsetup{skip=2pt} 
    \centering
    \begin{subfigure}[t]{0.30\textwidth}
        \centering
        \includegraphics[width=\linewidth]{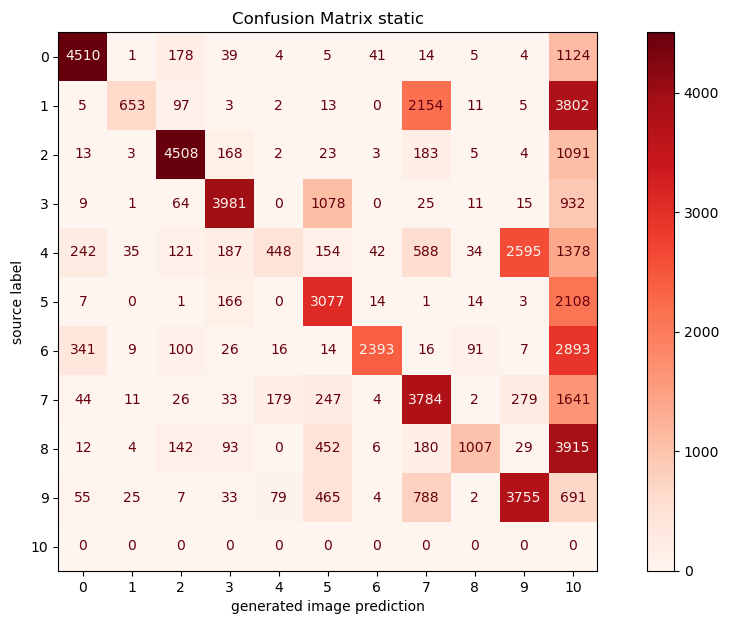}
        \caption{\textit{c}=1}
        \label{fig:mnist-emnist-static-c1}
    \end{subfigure}
    \begin{subfigure}[t]{0.30\textwidth}
        \centering
        \includegraphics[width=\linewidth]{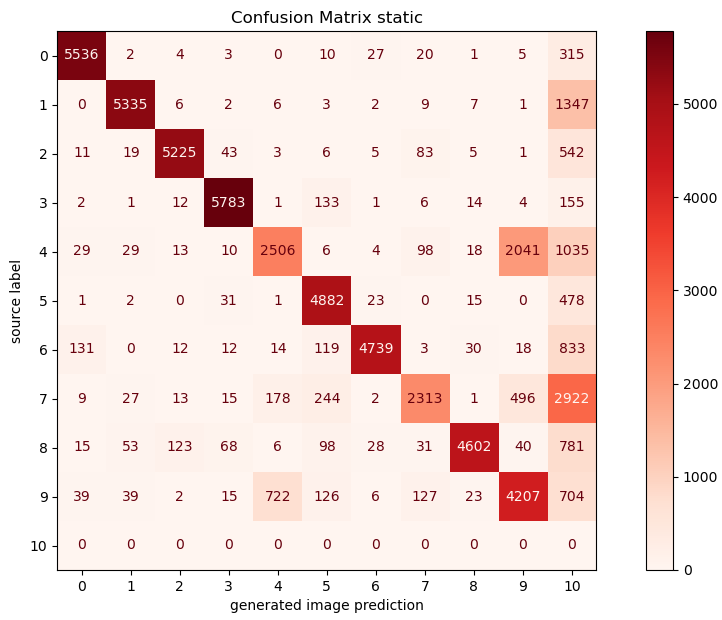}
        \caption{\textit{c}=2}
        \label{fig:mnist-emnist-static-c2}
    \end{subfigure} \\
    \begin{subfigure}[t]{0.30\textwidth}
        \centering
        \includegraphics[width=\linewidth]{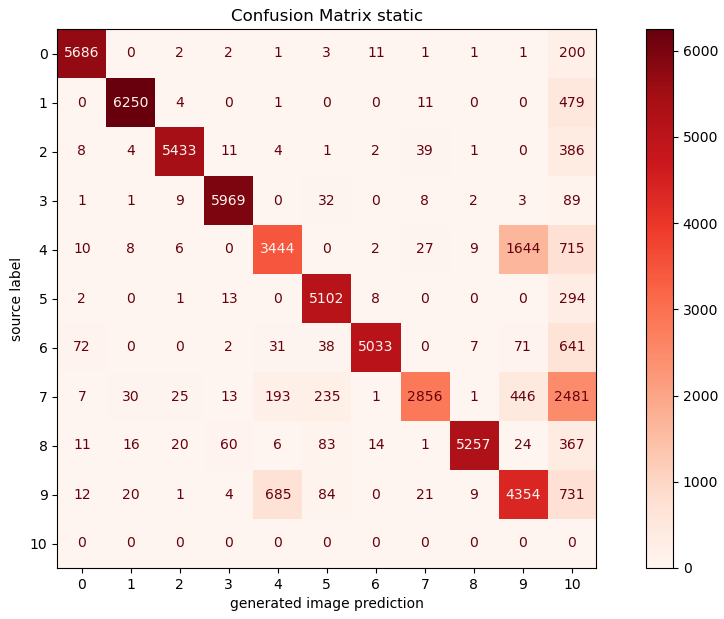}
        \caption{\textit{c}=4}
        \label{fig:mnist-emnist-static-c4}
    \end{subfigure}\hspace{2pt}%
    \begin{subfigure}[t]{0.30\textwidth}
        \centering
        \includegraphics[width=\linewidth]{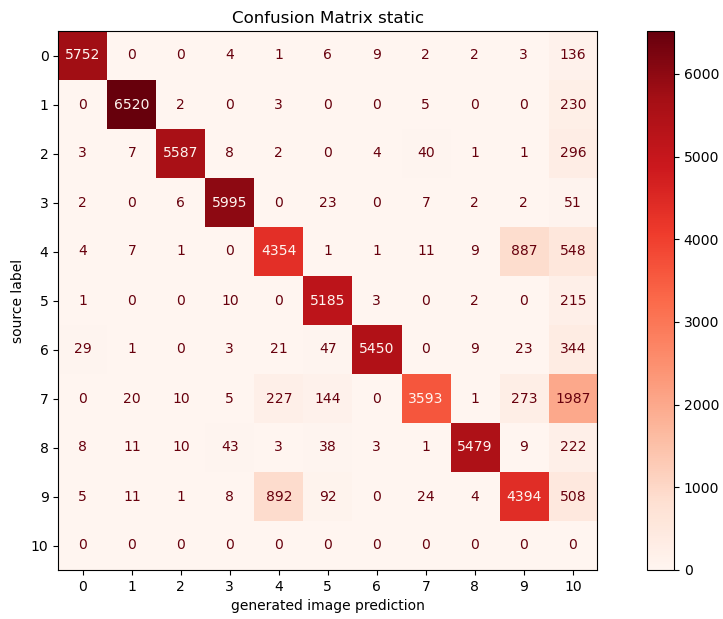}
        \caption{\textit{c}=8}
        \label{fig:mnist-emnist-static-c8}
    \end{subfigure}

    \caption{Confusion matrices for MNIST$\rightarrow$EMNIST domain translation using static subset selection. Accuracy is computed as the ratio between the trace and the sum of all entries in the confusion matrices.}
    \label{mnist-emnist-static}
\end{figure}

\begin{figure}[h!]
    \centering
    \includegraphics[width=0.75\textwidth]{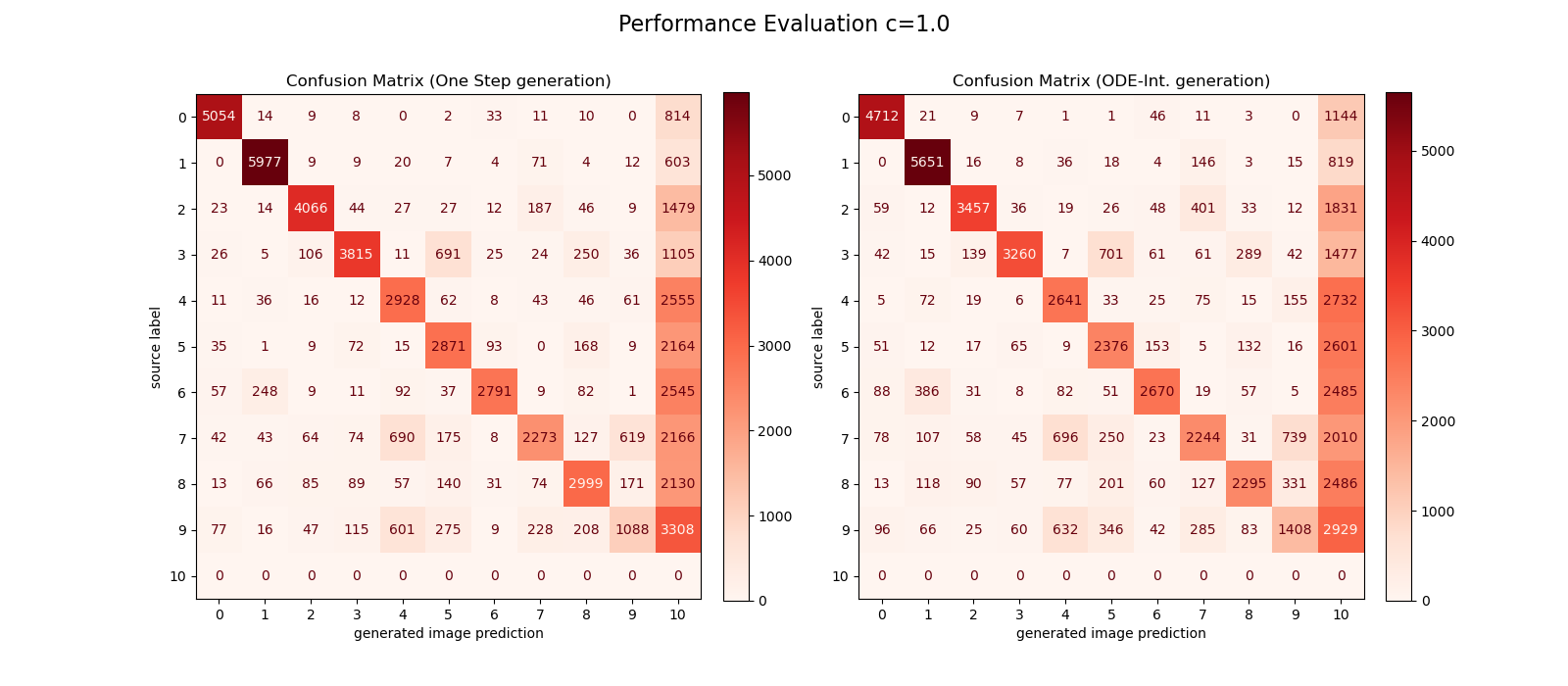}

    \includegraphics[width=0.75\textwidth]{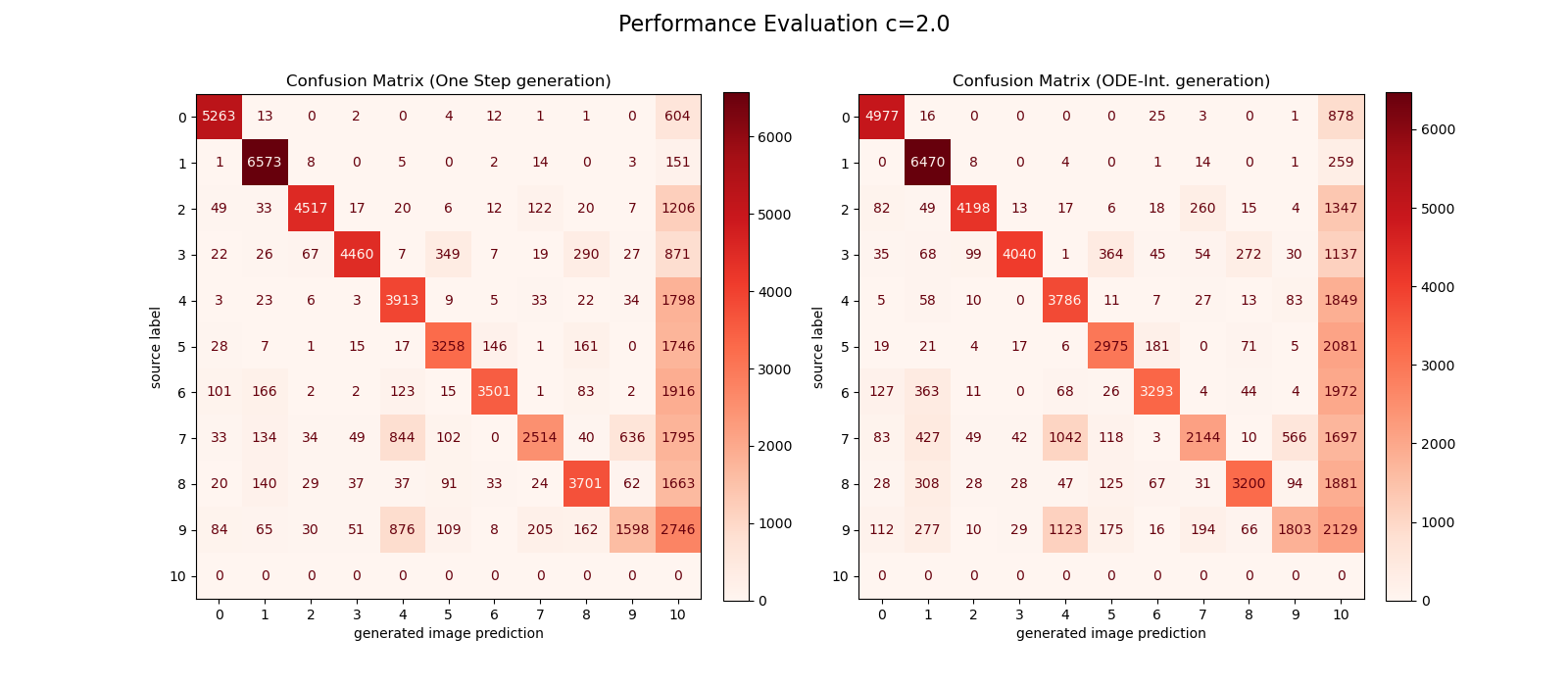}

    \includegraphics[width=0.75\textwidth]{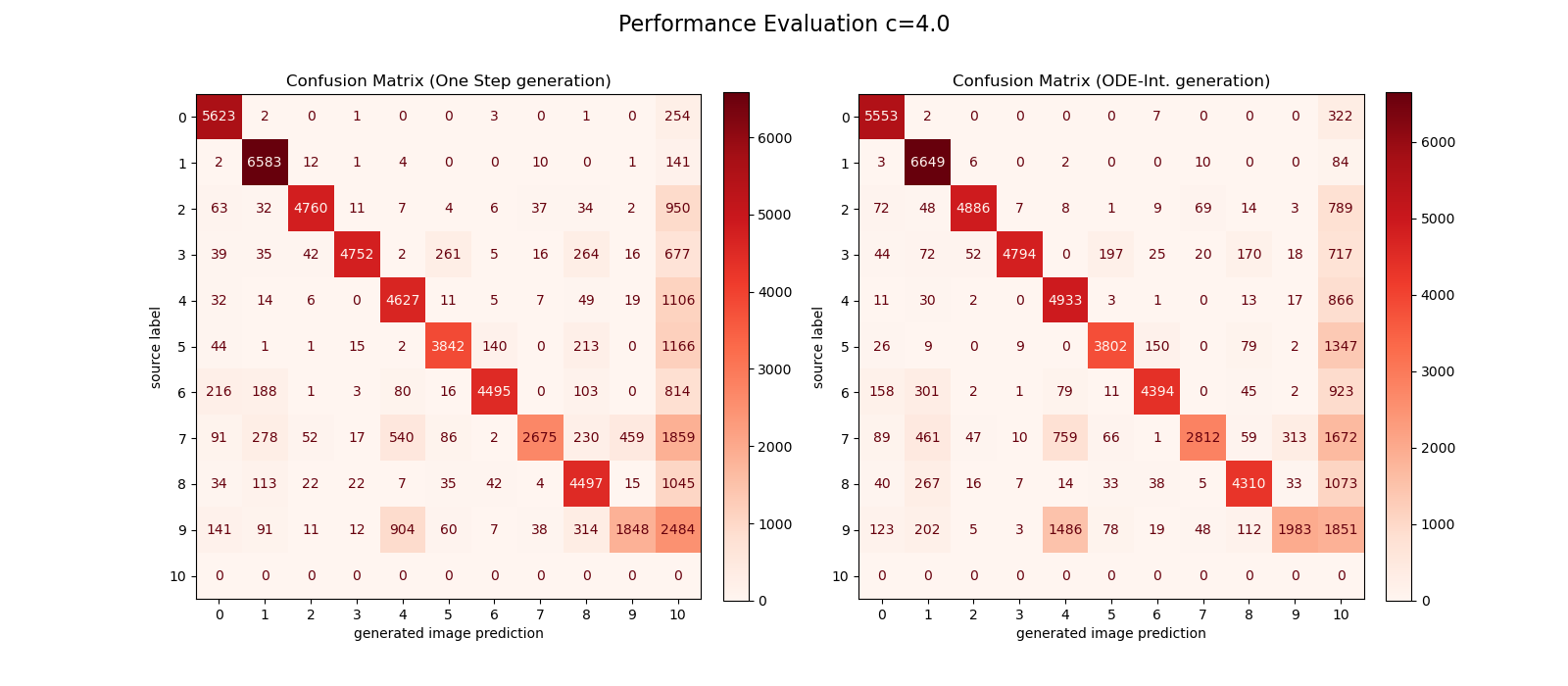}

    \includegraphics[width=0.75\textwidth]{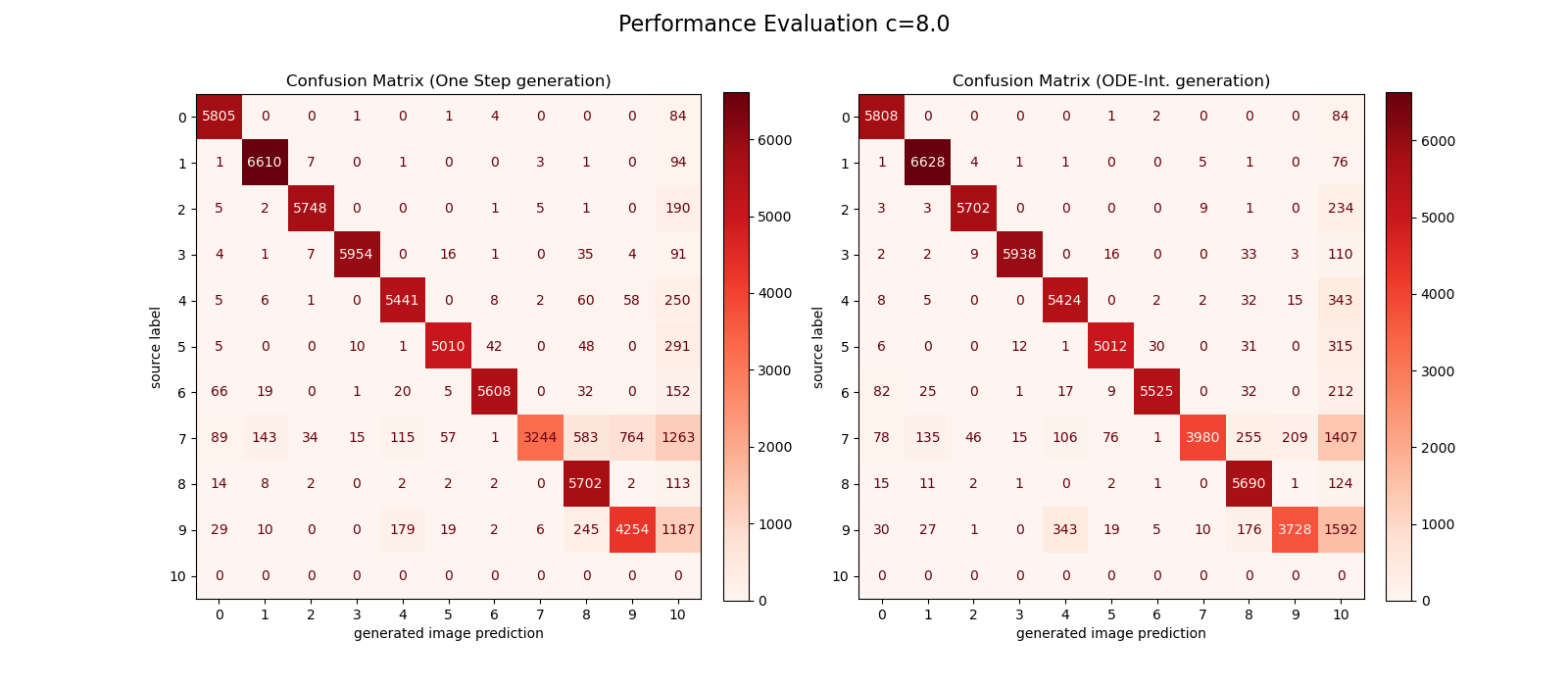}

    \caption{Confusion matrices for MNIST$\rightarrow$EMNIST domain translation using dynamic subset selection.}
    \label{mnist-emnist-dynamic}
\end{figure}

\clearpage
\section{t-SNE embeddings of UCI datasets}
\label{app:tsne-uci}

\begin{figure}[!ht]
    \centering
    \captionsetup[subfigure]{skip=0pt}    

    \begin{subfigure}[t]{0.24\textwidth}\centering
        \includegraphics[width=\linewidth]{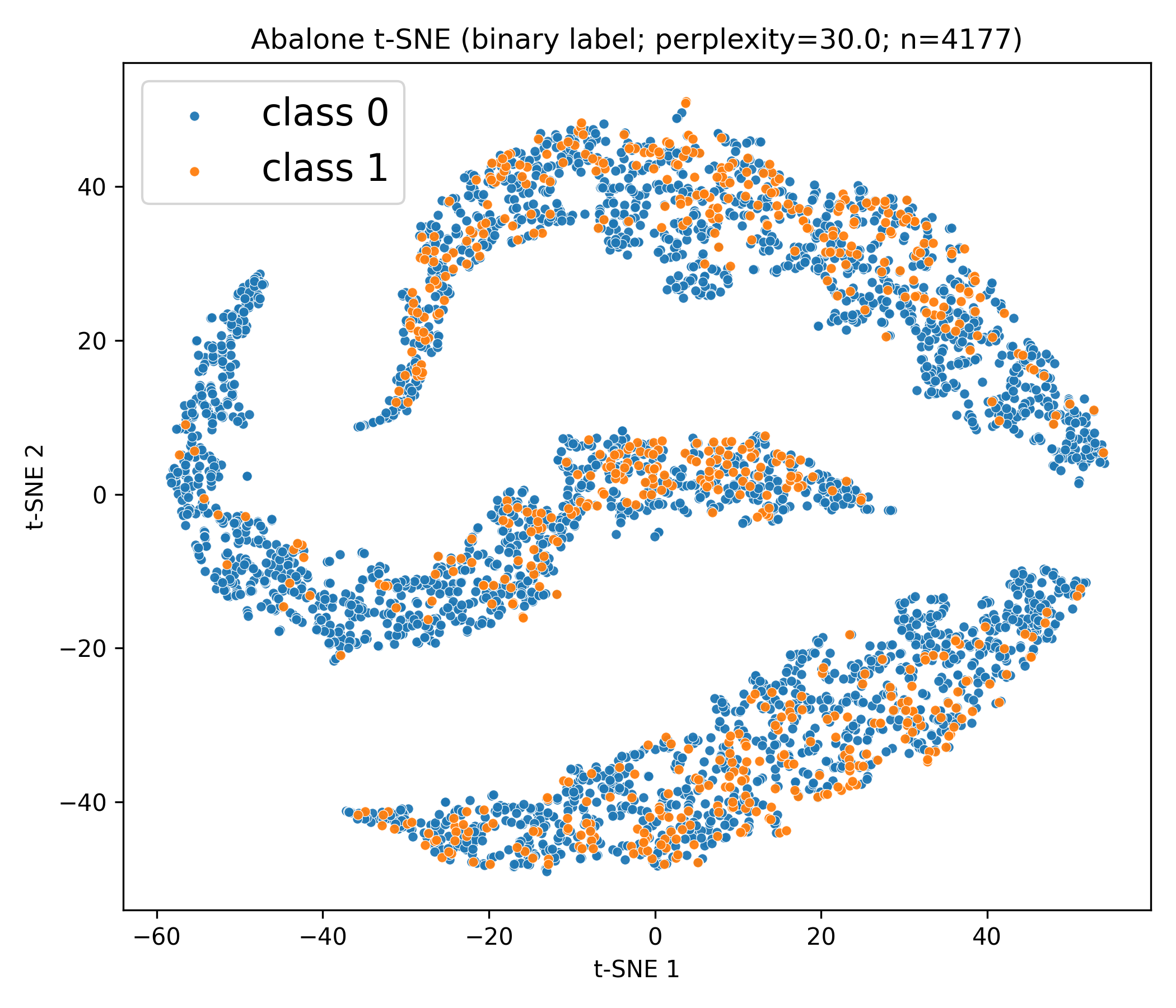}
        \caption{}\label{fig:tsne-abalone}
    \end{subfigure}
    \begin{subfigure}[t]{0.24\textwidth}\centering
        \includegraphics[width=\linewidth]{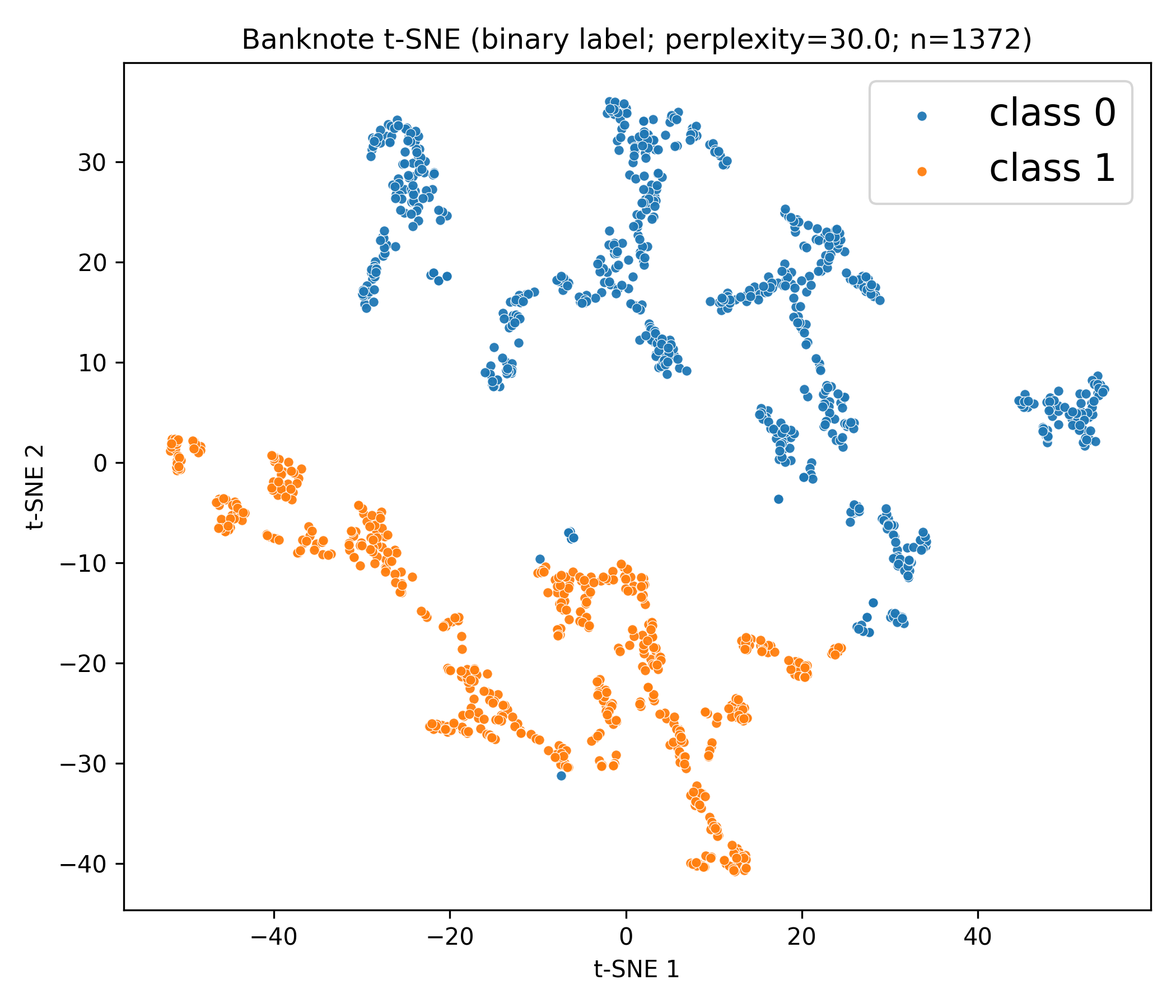}
        \caption{}\label{fig:tsne-banknote}
    \end{subfigure}
    \begin{subfigure}[t]{0.24\textwidth}\centering
        \includegraphics[width=\linewidth]{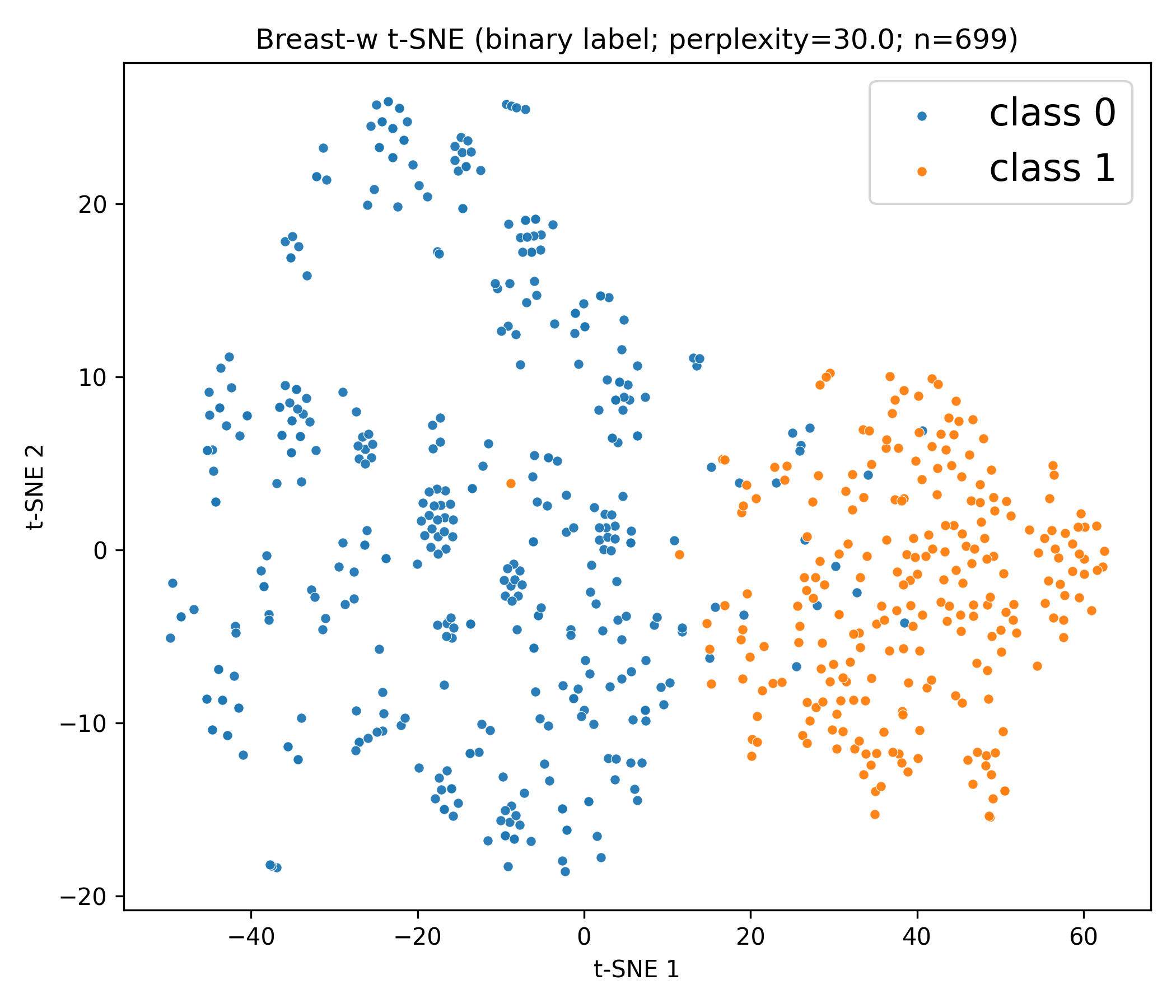}
        \caption{}\label{fig:tsne-breastw}
    \end{subfigure}
    \begin{subfigure}[t]{0.24\textwidth}\centering
        \includegraphics[width=\linewidth]{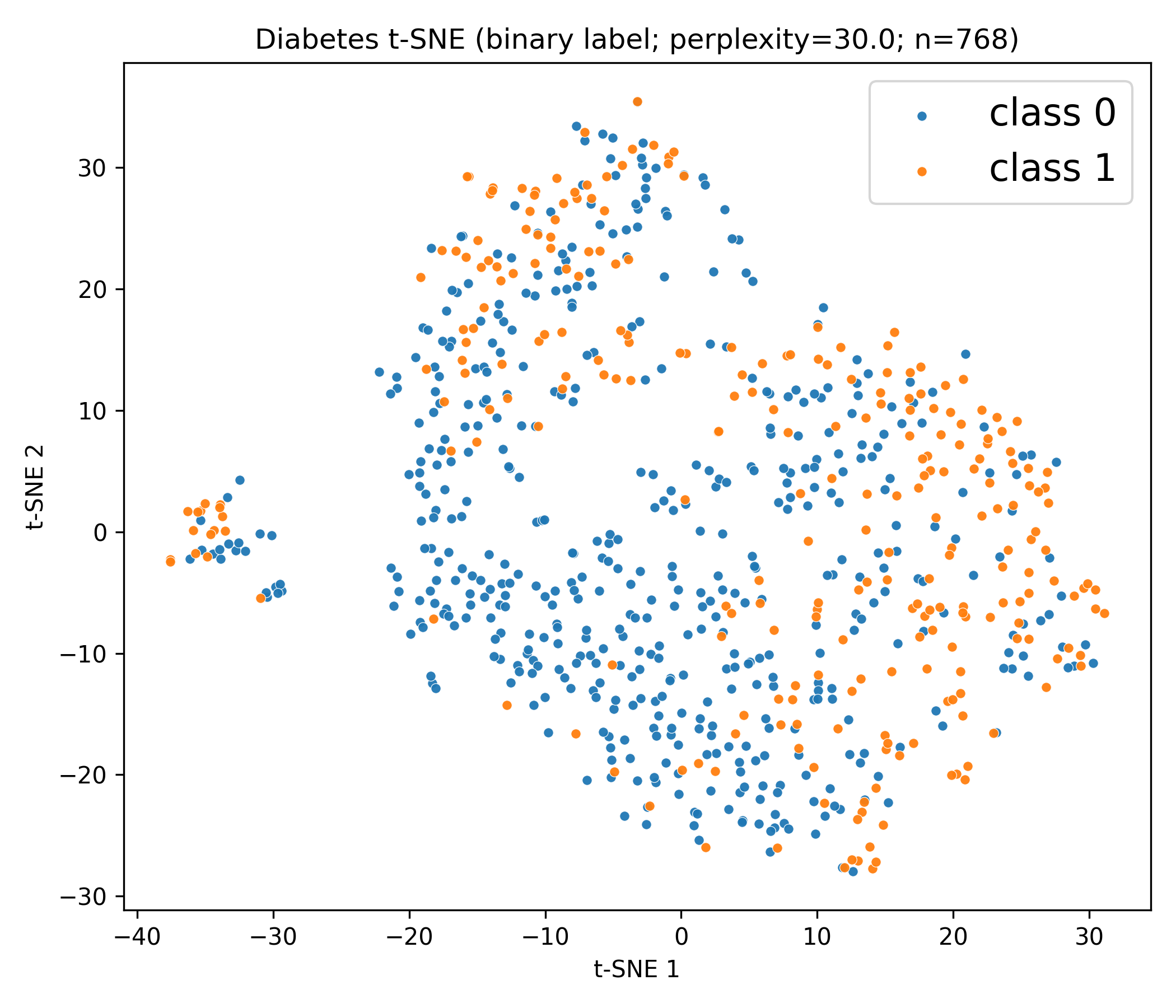}
        \caption{}\label{fig:tsne-diabetes}
    \end{subfigure}

    \vspace{1mm}

    \begin{subfigure}[t]{0.24\textwidth}\centering
        \includegraphics[width=\linewidth]{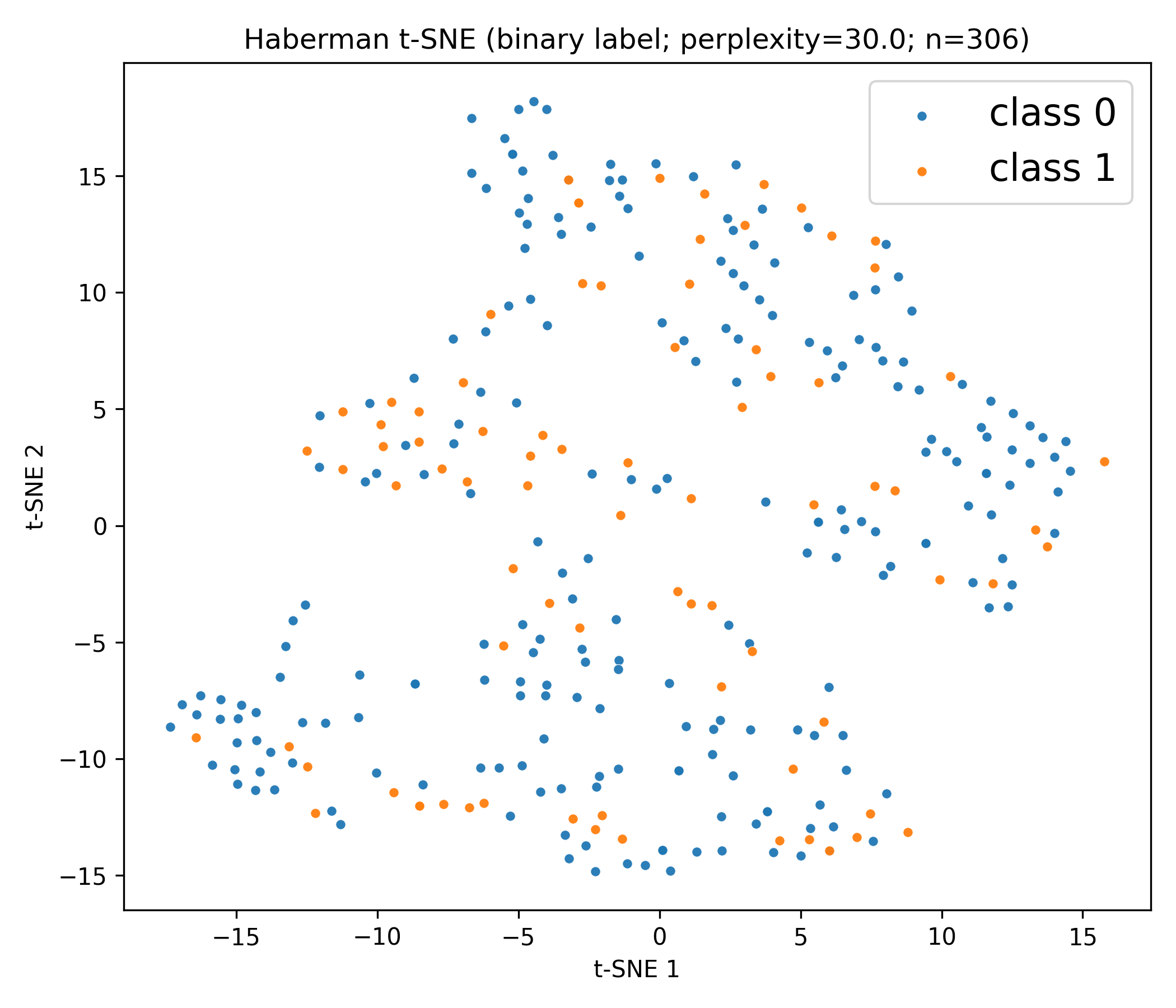}
        \caption{}\label{fig:tsne-haberman}
    \end{subfigure}
    \begin{subfigure}[t]{0.24\textwidth}\centering
        \includegraphics[width=\linewidth]{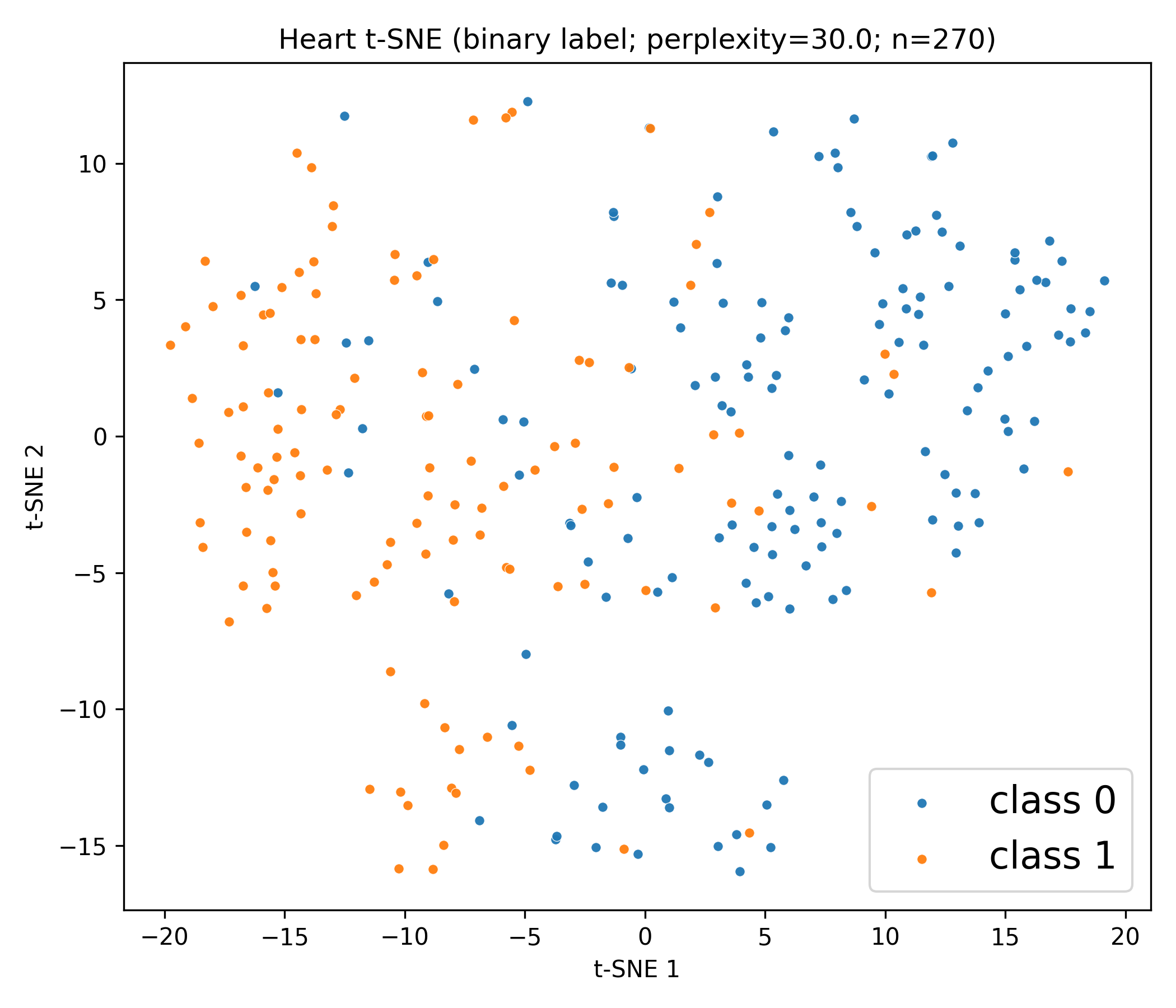}
        \caption{}\label{fig:tsne-heart}
    \end{subfigure}
    \begin{subfigure}[t]{0.24\textwidth}\centering
        \includegraphics[width=\linewidth]{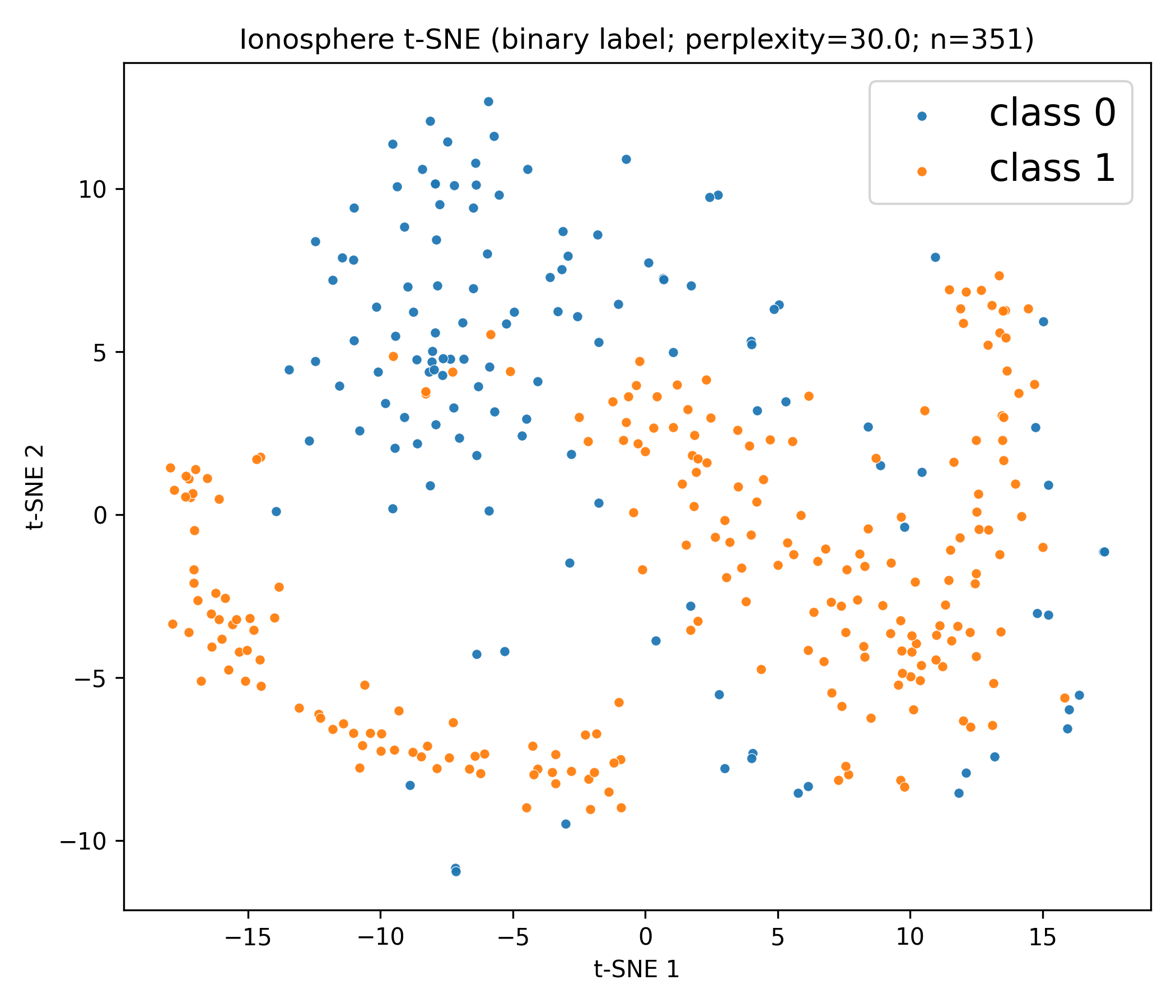}
        \caption{}\label{fig:tsne-ionosphere}
    \end{subfigure}
    \begin{subfigure}[t]{0.24\textwidth}\centering
        \includegraphics[width=\linewidth]{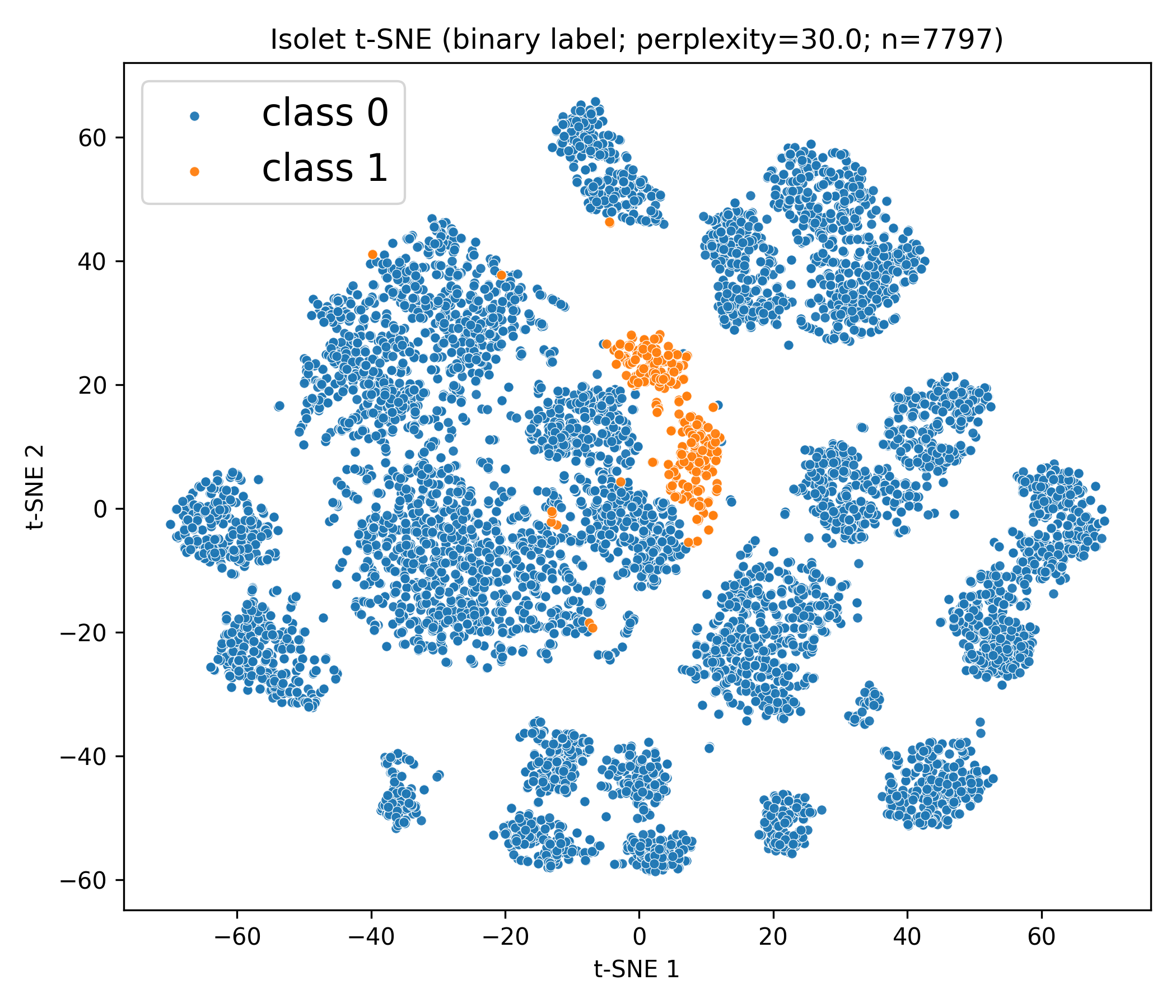}
        \caption{}\label{fig:tsne-isolet}
    \end{subfigure}

    \vspace{1mm}

    \begin{subfigure}[t]{0.24\textwidth}\centering
        \includegraphics[width=\linewidth]{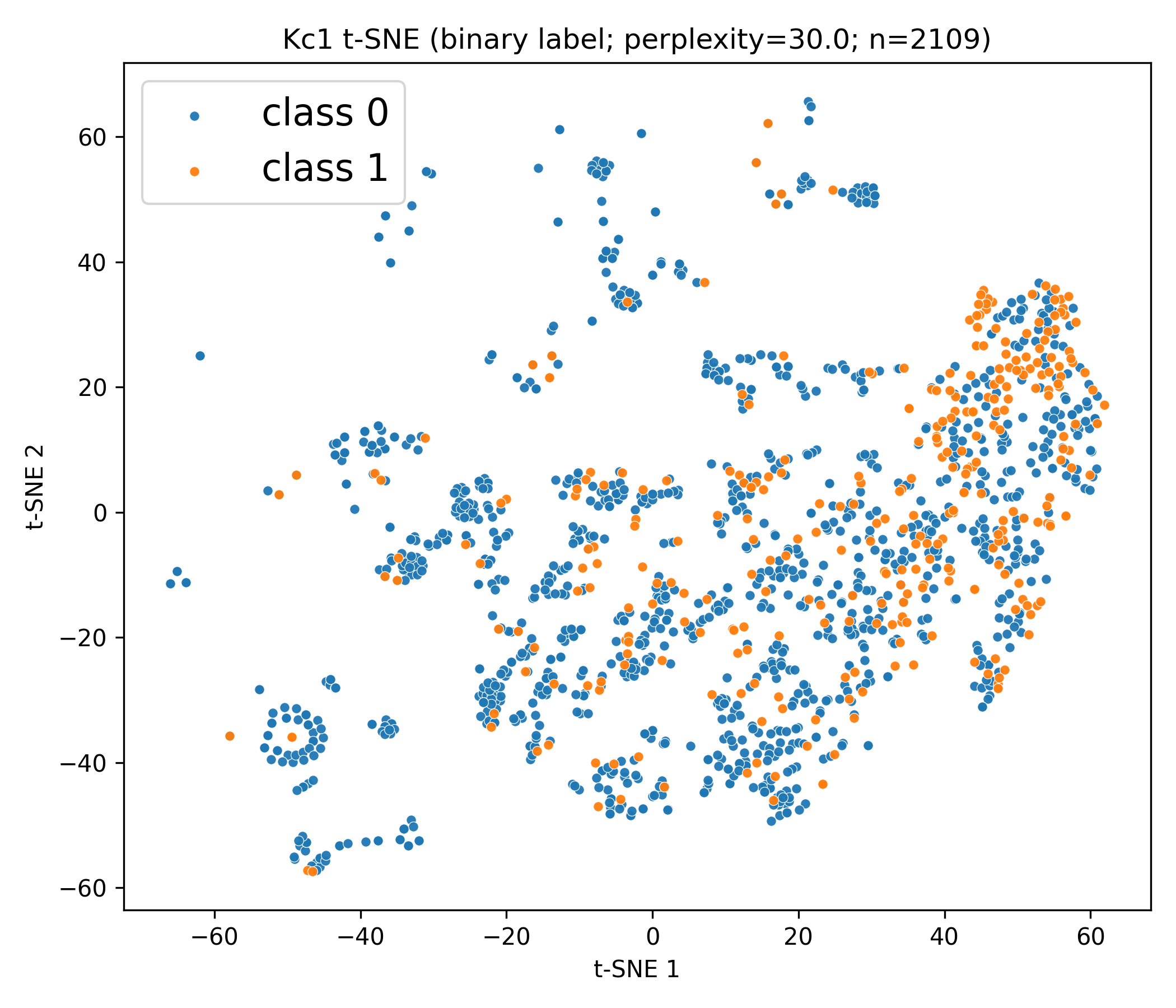}
        \caption{}\label{fig:tsne-kc1}
    \end{subfigure}
    \begin{subfigure}[t]{0.24\textwidth}\centering
        \includegraphics[width=\linewidth]{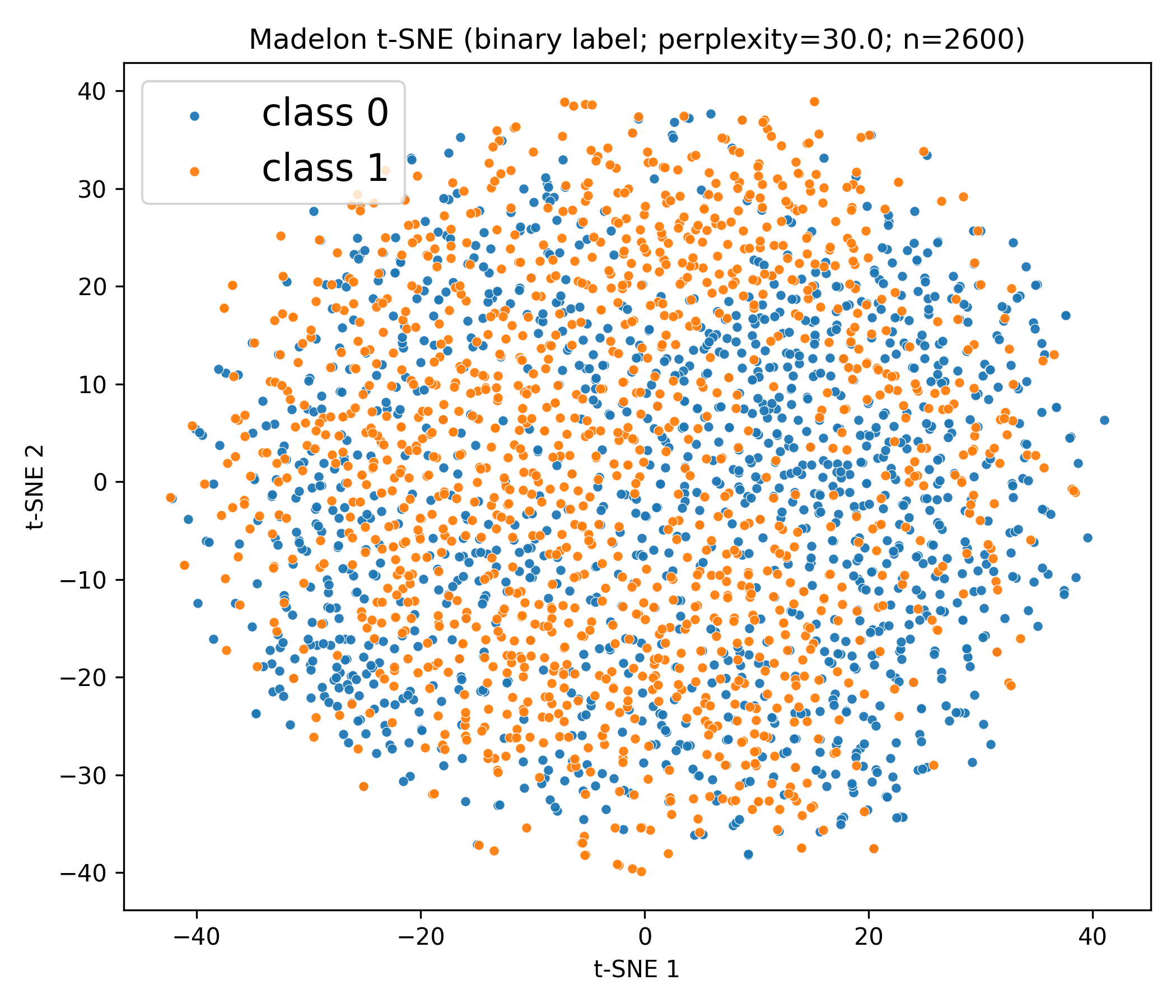}
        \caption{}\label{fig:tsne-madelon}
    \end{subfigure}
    \begin{subfigure}[t]{0.24\textwidth}\centering
        \includegraphics[width=\linewidth]{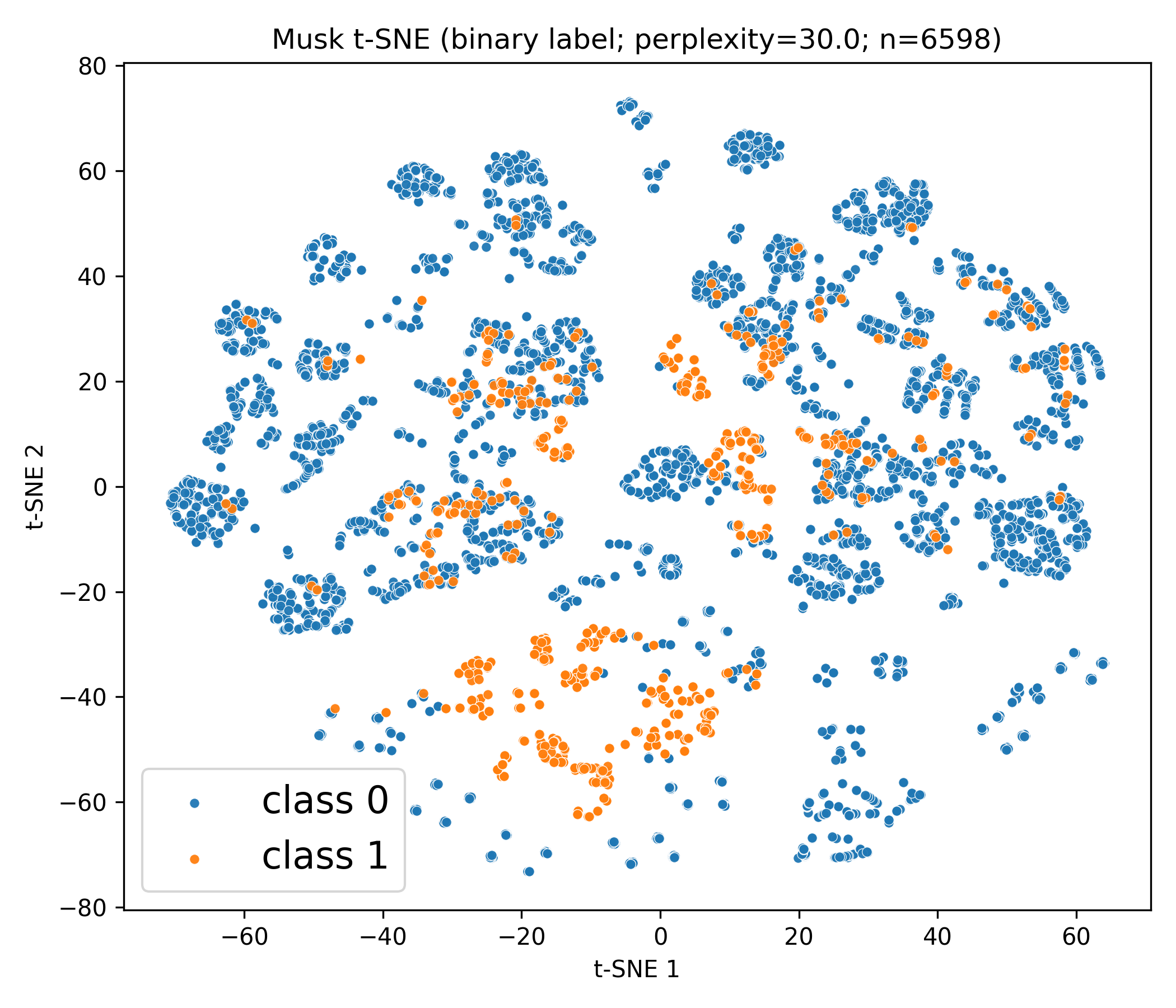}
        \caption{}\label{fig:tsne-musk}
    \end{subfigure}
    \begin{subfigure}[t]{0.24\textwidth}\centering
        \includegraphics[width=\linewidth]{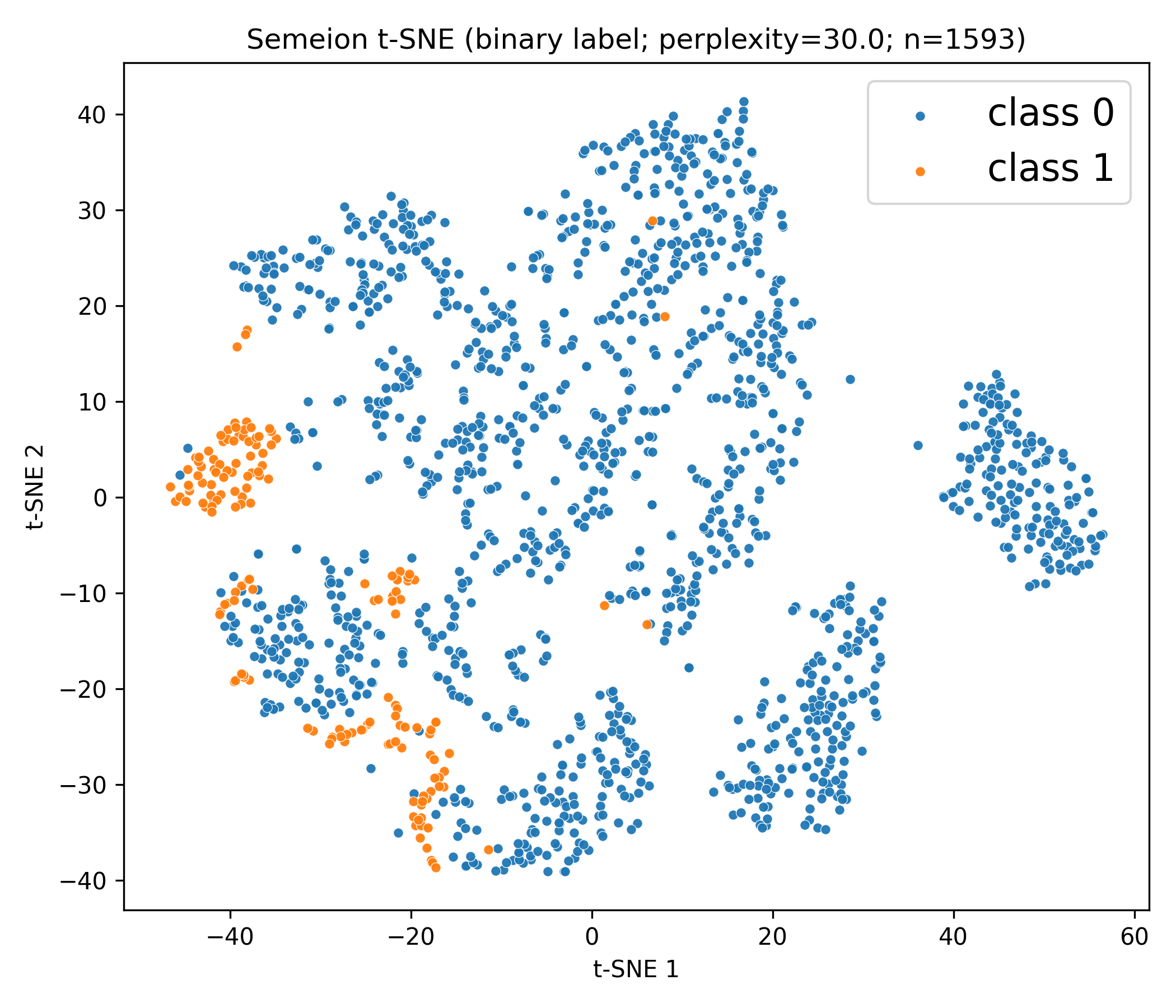}
        \caption{}\label{fig:tsne-semeion}
    \end{subfigure}

    \vspace{1mm}

    \begin{subfigure}[t]{0.24\textwidth}\centering
        \includegraphics[width=\linewidth]{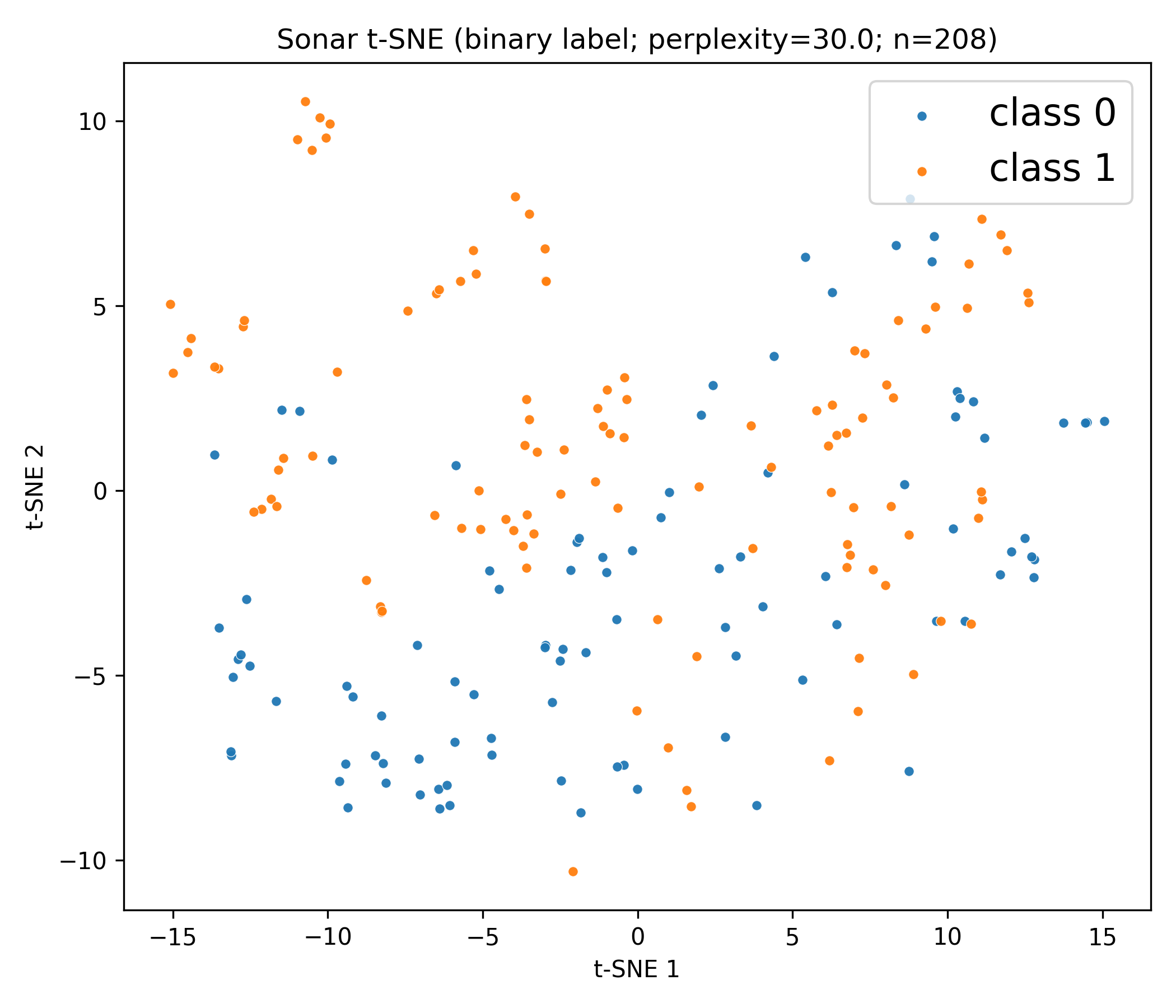}
        \caption{}\label{fig:tsne-sonar}
    \end{subfigure}
    \begin{subfigure}[t]{0.24\textwidth}\centering
        \includegraphics[width=\linewidth]{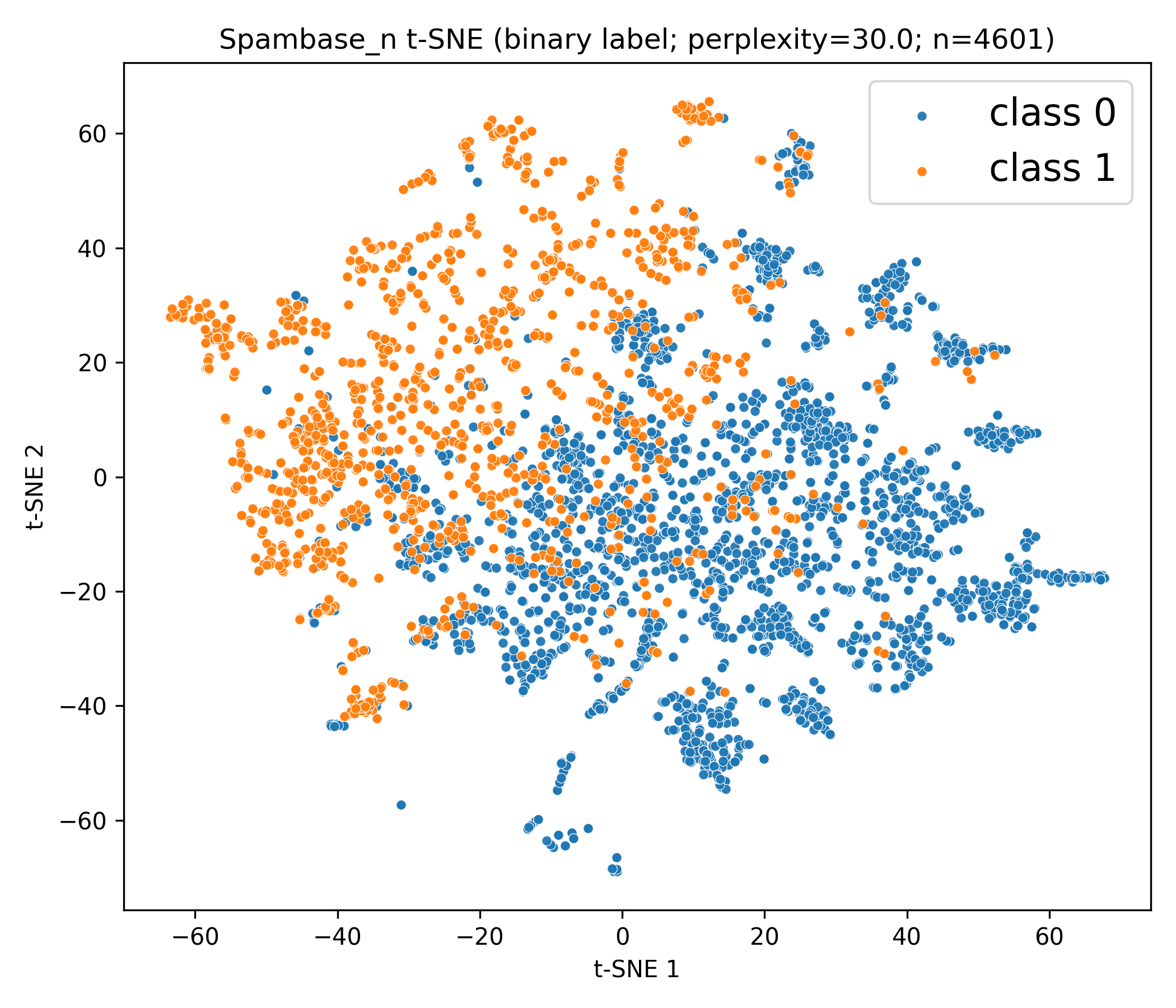}
        \caption{}\label{fig:tsne-spambase}
    \end{subfigure}
    \begin{subfigure}[t]{0.24\textwidth}\centering
        \includegraphics[width=\linewidth]{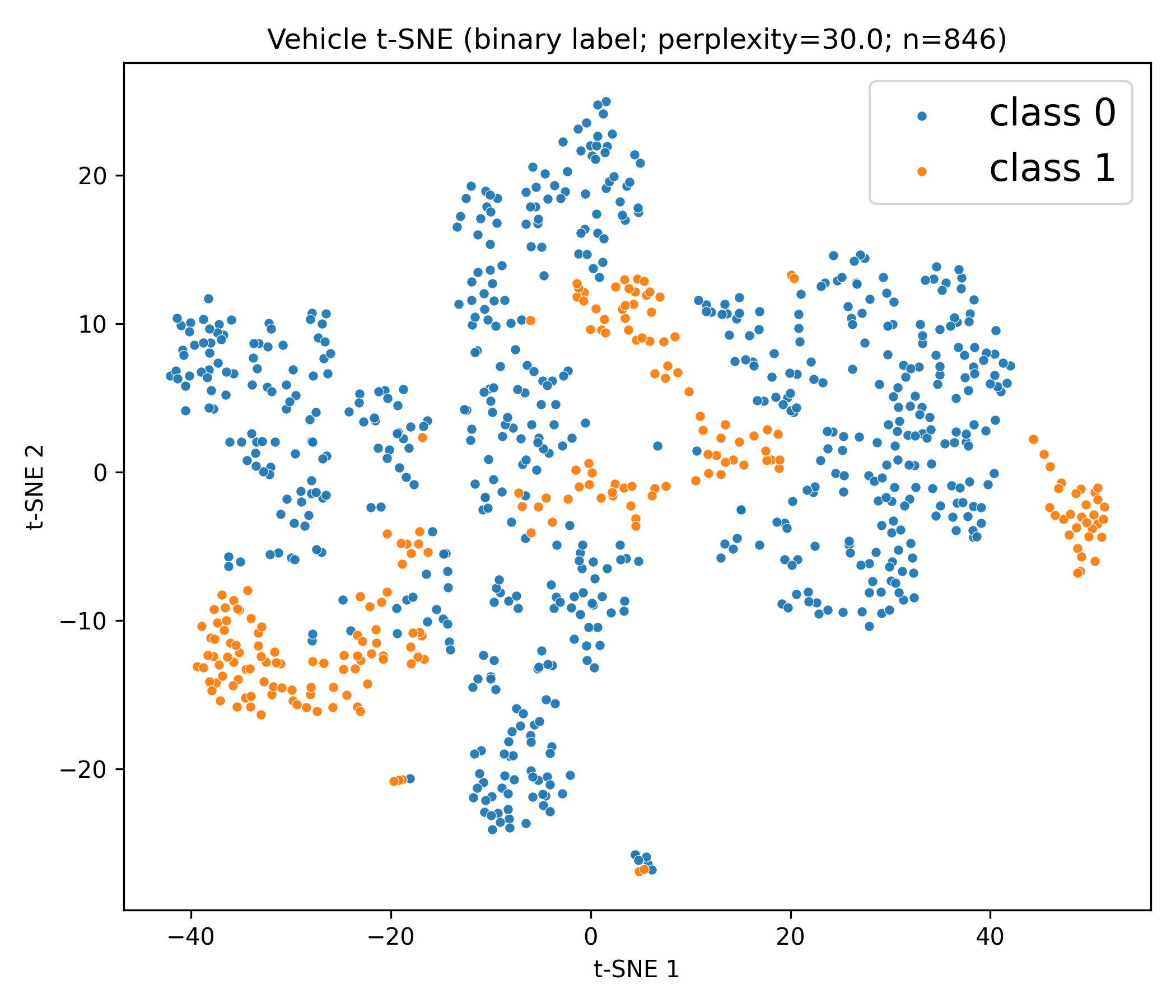}
        \caption{}\label{fig:tsne-vehicle}
    \end{subfigure}
    \begin{subfigure}[t]{0.24\textwidth}\centering
        \includegraphics[width=\linewidth]{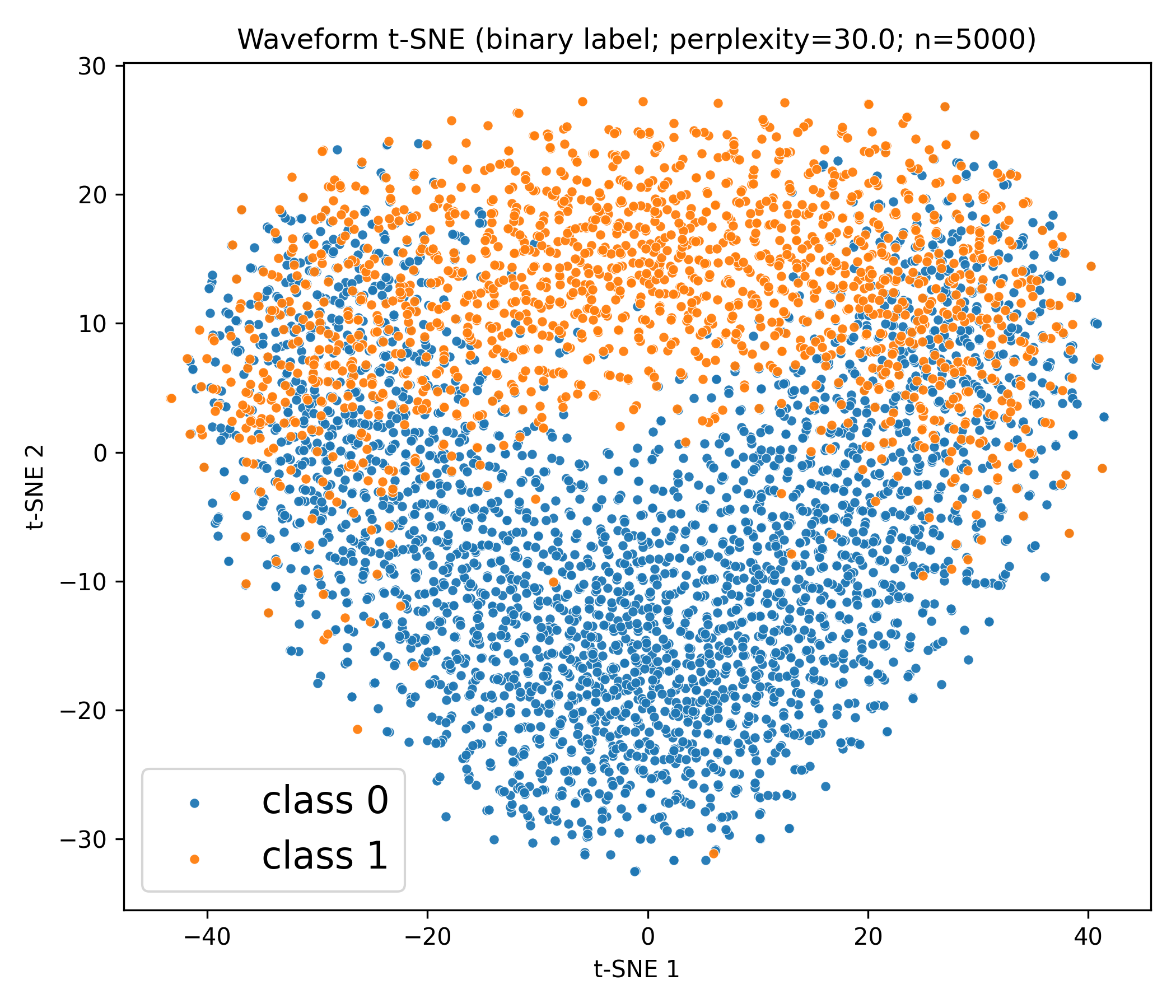}
        \caption{}\label{fig:tsne-waveform}
    \end{subfigure}

    \vspace{1mm}

    \begin{subfigure}[t]{0.24\textwidth}\centering
        \includegraphics[width=\linewidth]{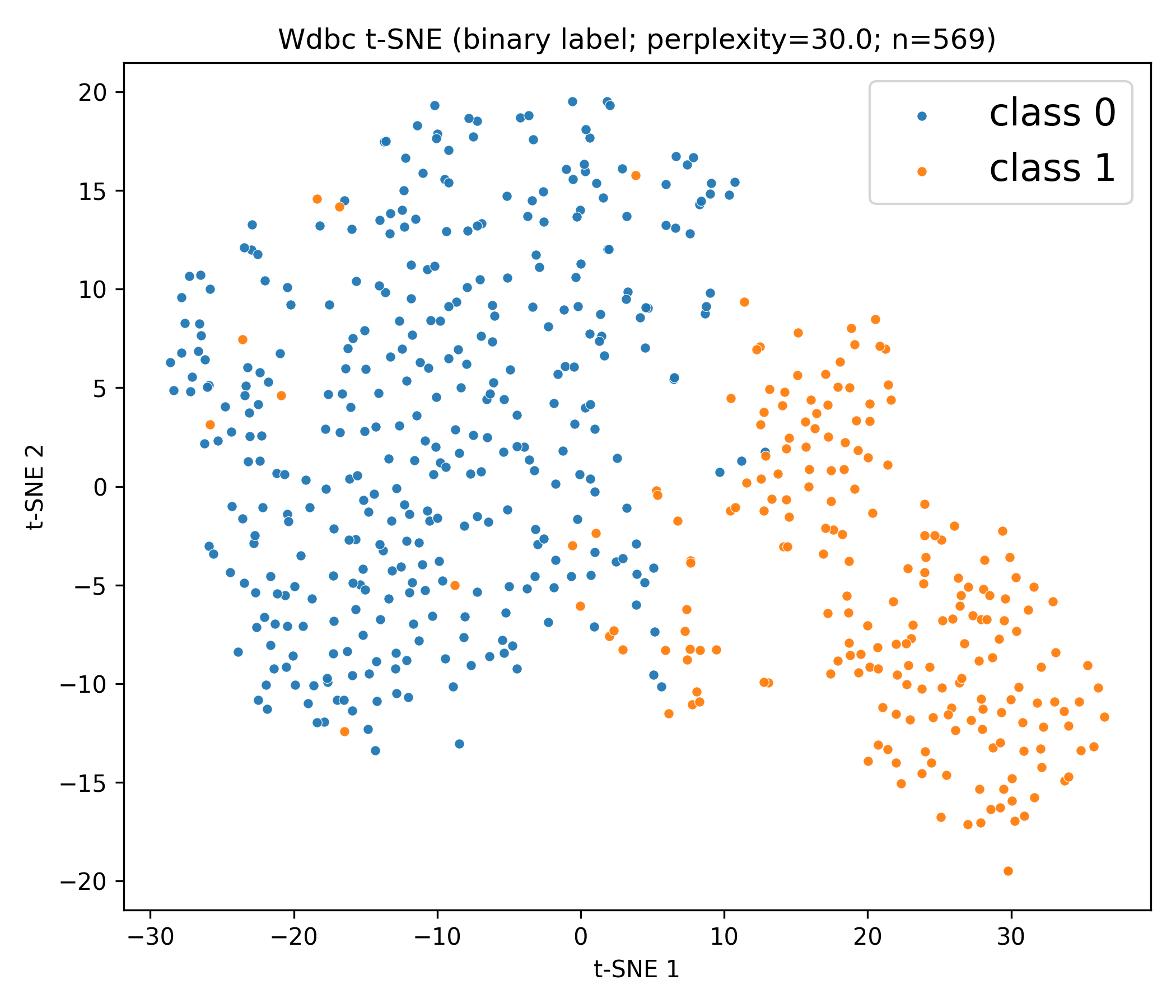}
        \caption{}\label{fig:tsne-wdbc}
    \end{subfigure}
    \begin{subfigure}[t]{0.23\textwidth}\centering
        \includegraphics[width=\linewidth]{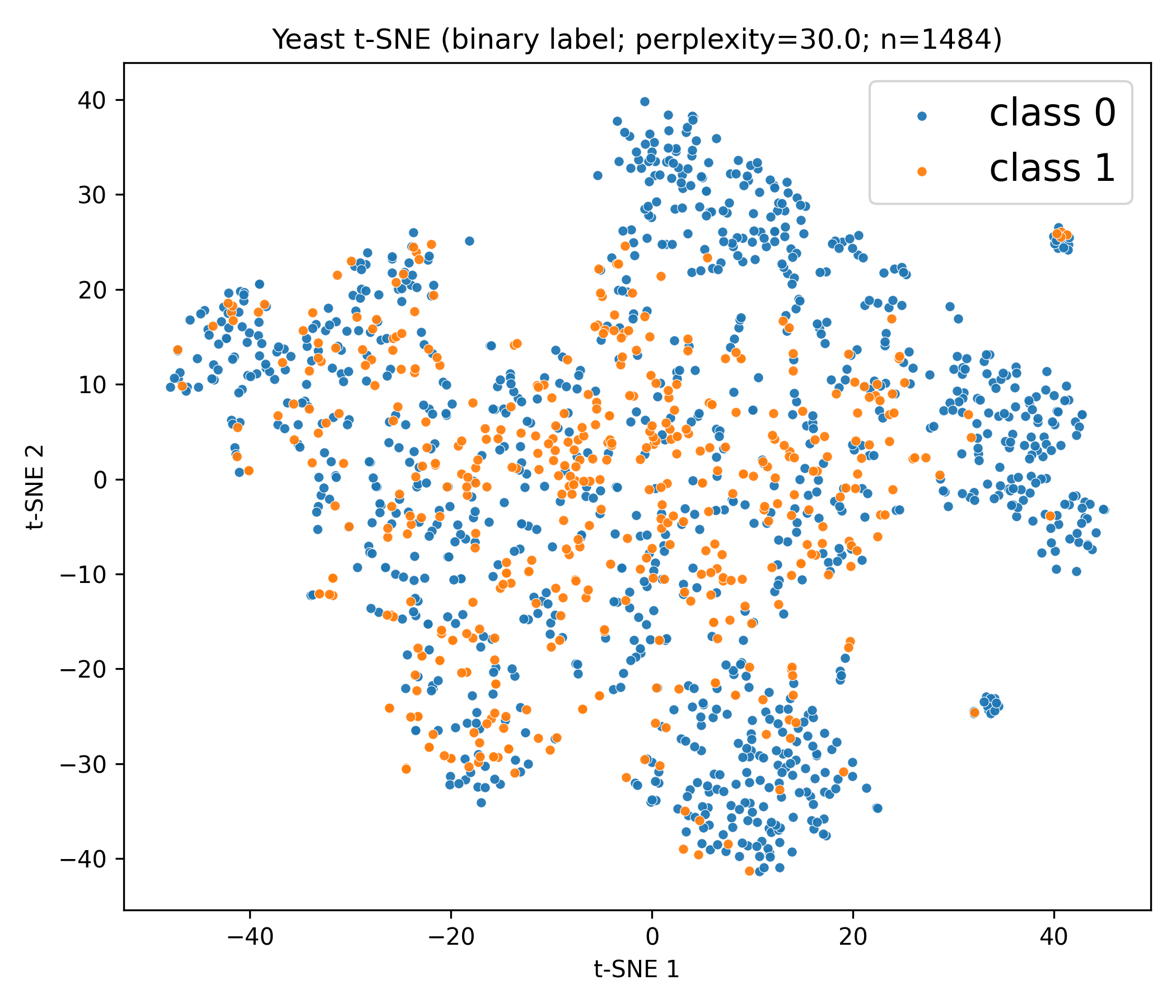}
        \caption{}\label{fig:tsne-yeast}
    \end{subfigure}
    \begin{subfigure}[t]{0.24\textwidth}\centering
        \includegraphics[width=\linewidth]{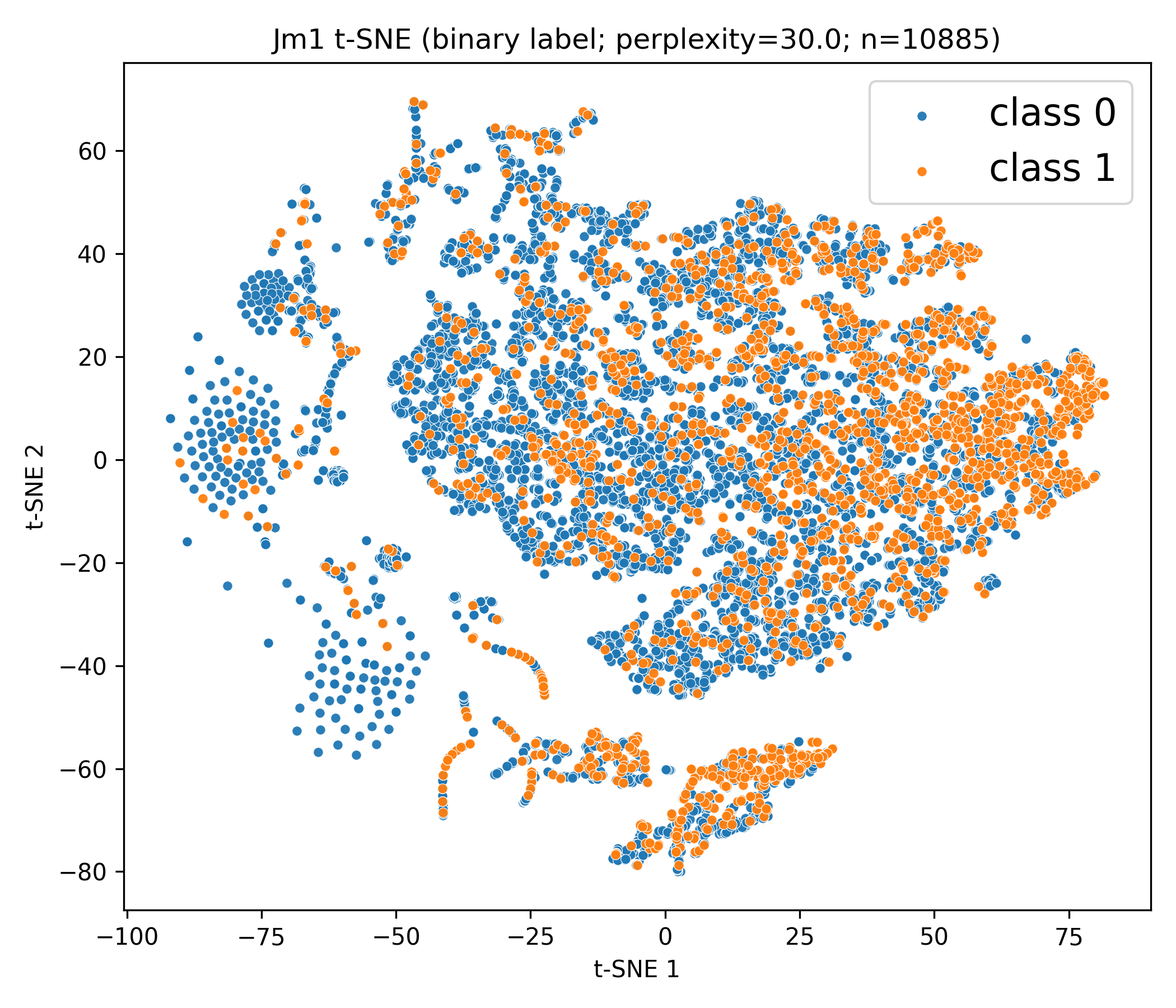}
        \caption{}\label{fig:tsne-Jm1}
    \end{subfigure}
    \begin{subfigure}[t]{0.23\textwidth}\centering
        \includegraphics[width=\linewidth]{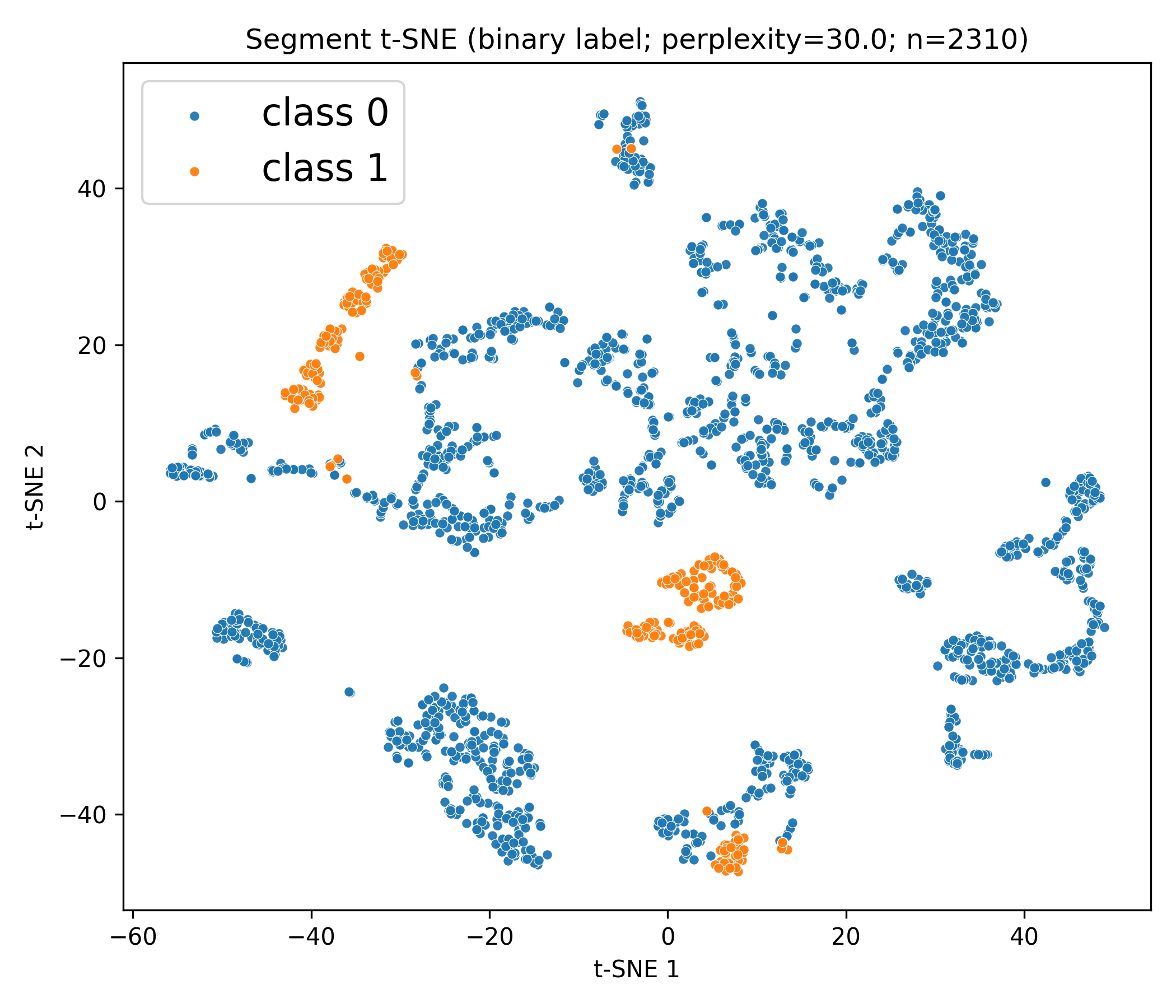}
        \caption{}\label{fig:tsne-segmant}
    \end{subfigure}

    \caption{The t-SNE embeddings for (a) Abalone, (b) Banknote, (c) Breast-w, (d) Diabetes, (e) Haberman, (f) Heart, (g) Ionosphere, (h) Isolet, (i) Kc1, (j) Madelon, (k) Musk, (l) Semeion, (m) Sonar, (n) Spambase, (o) Vehicle, (p) Waveform, (q) Wdbc, (r) Yeast, (s) Jm1, and (t) Segment datasets.}
    \label{tsne-illustrations-1}
\end{figure}

\clearpage
\section{Results from FFHQ}
\label{app:ffhq_more}
\begin{figure}[!h]
    \centering
    \captionsetup[subfigure]{skip=0pt} 

    \begin{subfigure}[t]{0.49\textwidth}
        \centering
        \includegraphics[width=\linewidth]{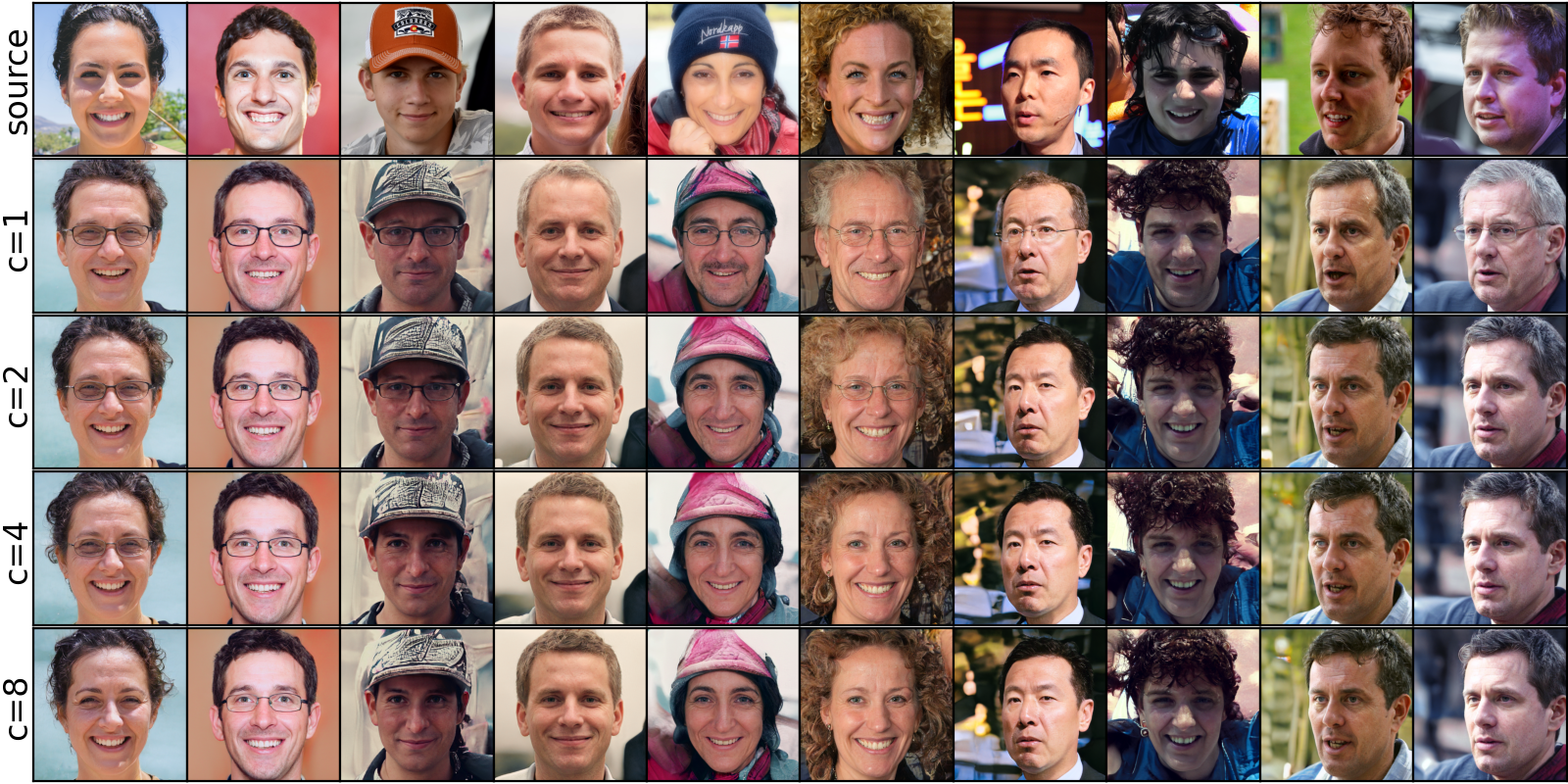}
        \caption{Static subset selection}
        \label{fig:young-old-static}
    \end{subfigure}\hfill
    \begin{subfigure}[t]{0.49\textwidth}
        \centering
        \includegraphics[width=\linewidth]{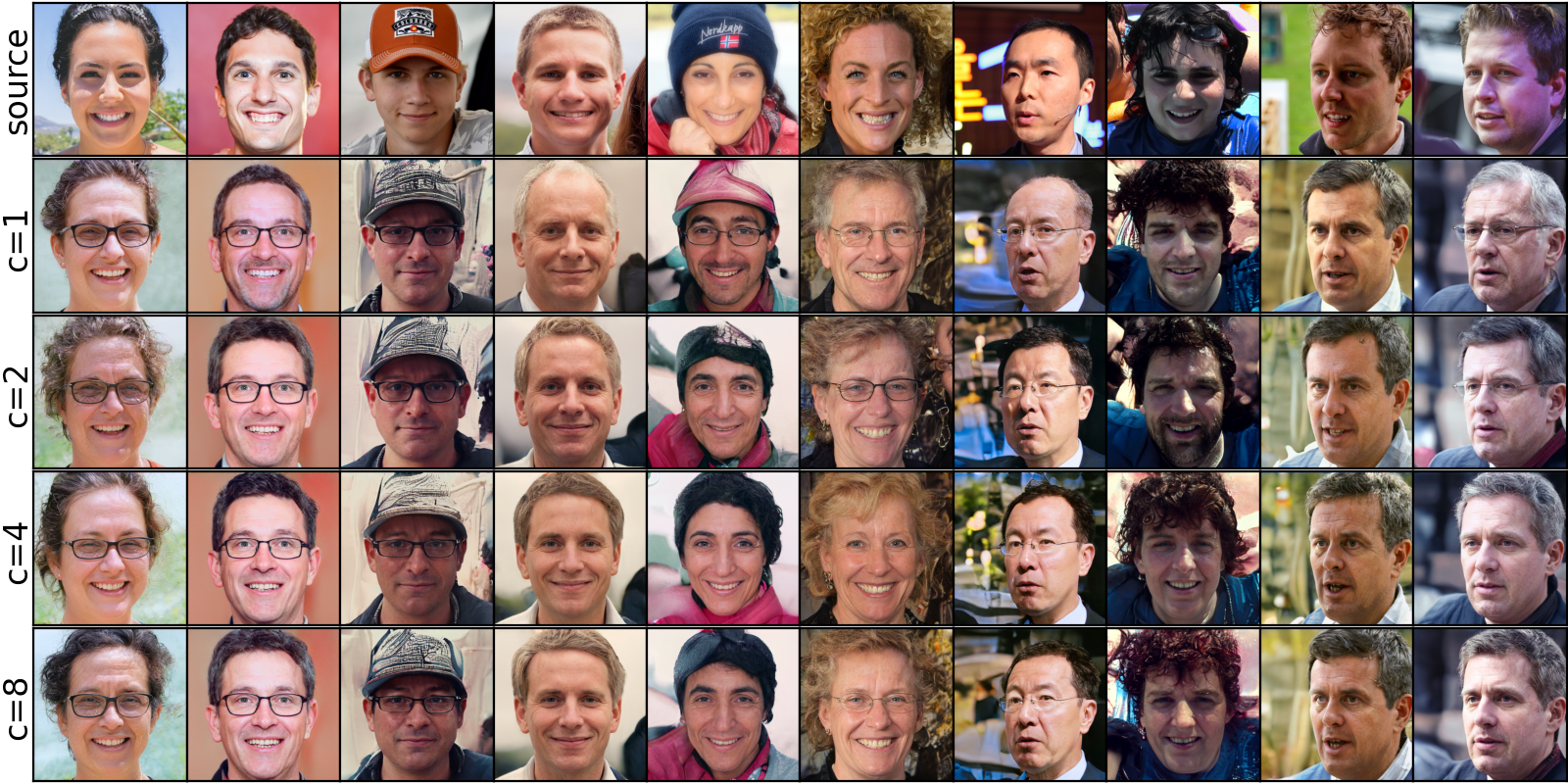}
        \caption{Dynamic subset selection}
        \label{fig:young-old-dynamic}
    \end{subfigure}

    \caption{FFHQ young$\rightarrow$old translation using (a) static and (b) dynamic subset selection. Dynamic subset selection is evaluated using Euler integration with 100 steps.}
    \label{young-to-old}
\end{figure}

\begin{figure}[!h]
    \centering
    \captionsetup[subfigure]{skip=0pt} 

    \begin{subfigure}[t]{0.49\textwidth}
        \centering
        \includegraphics[width=\linewidth]{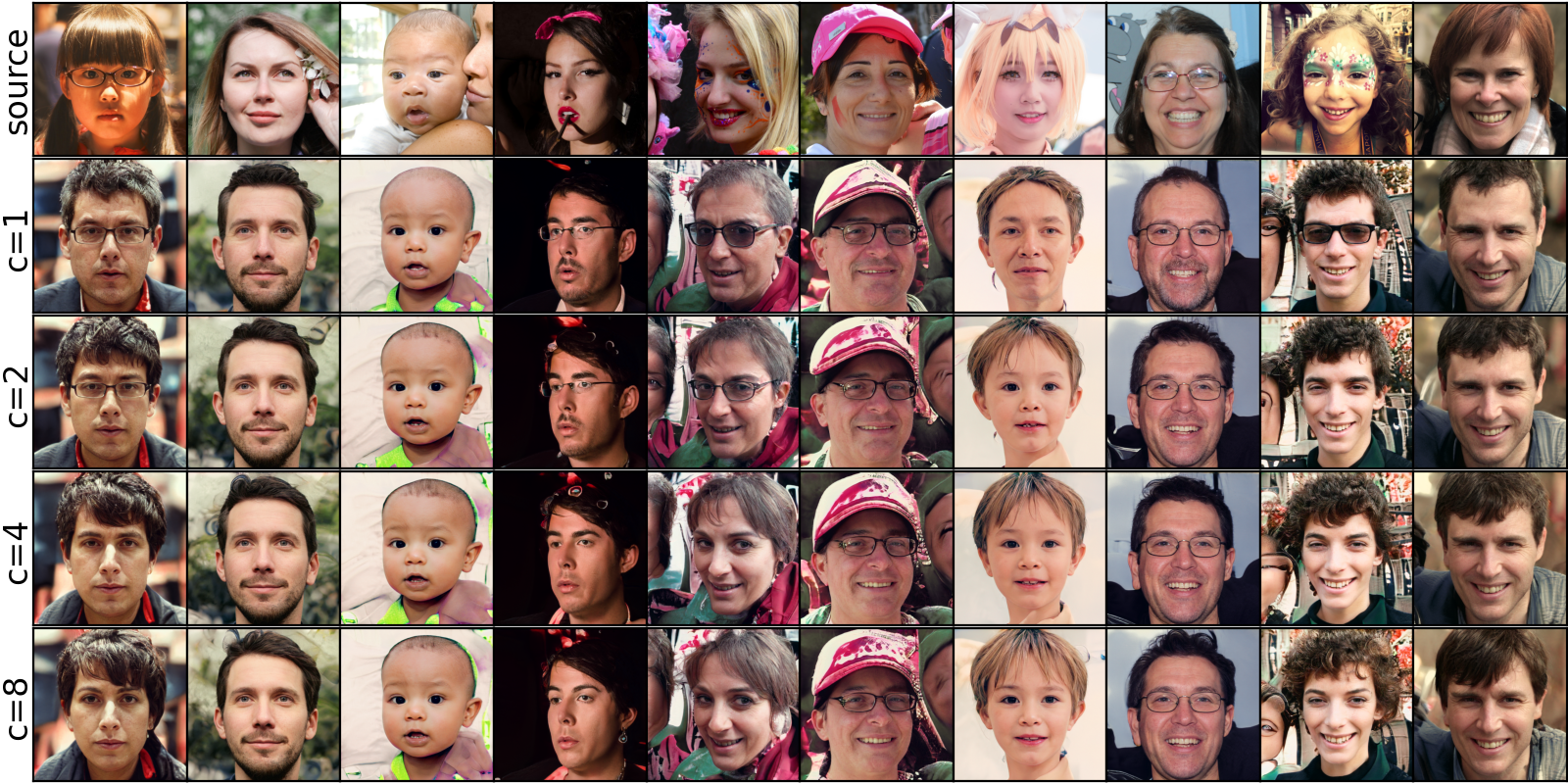}
        \caption{Static subset selection}
        \label{fig:woman-man-static}
    \end{subfigure}\hfill
    \begin{subfigure}[t]{0.49\textwidth}
        \centering
        \includegraphics[width=\linewidth]{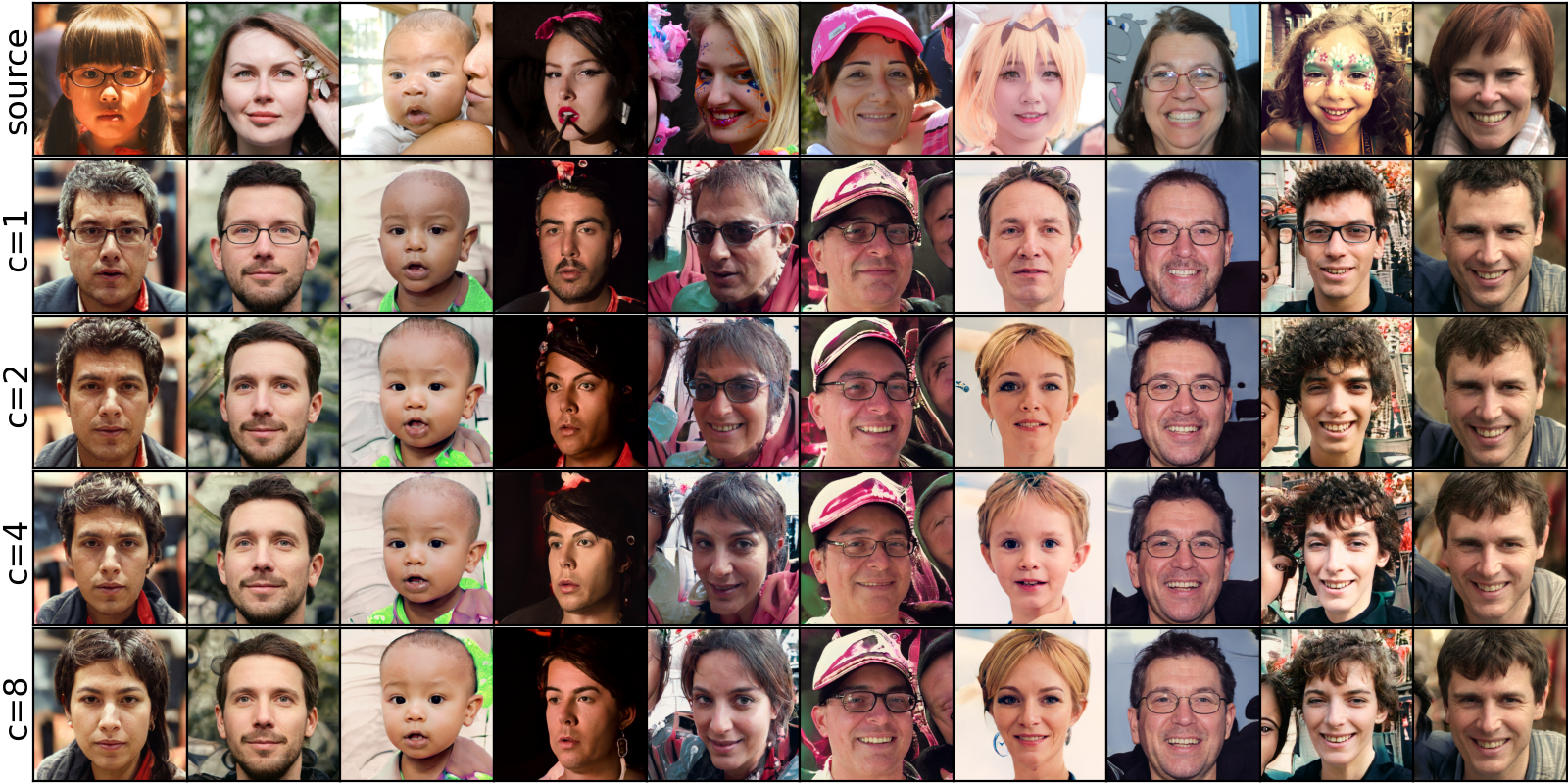}
        \caption{Dynamic subset selection}
        \label{fig:woman-man-dynamic}
    \end{subfigure}

    \caption{FFHQ woman$\rightarrow$man translation using (a) static and (b) dynamic subset selection. Dynamic subset selection is evaluated using Euler integration with 100 steps.}
    \label{woman-to-man}
\end{figure}

\begin{figure}[!h]
    \centering
    \captionsetup[subfigure]{skip=0pt} 

    \begin{subfigure}[t]{0.49\textwidth}
        \centering
        \includegraphics[width=\linewidth]{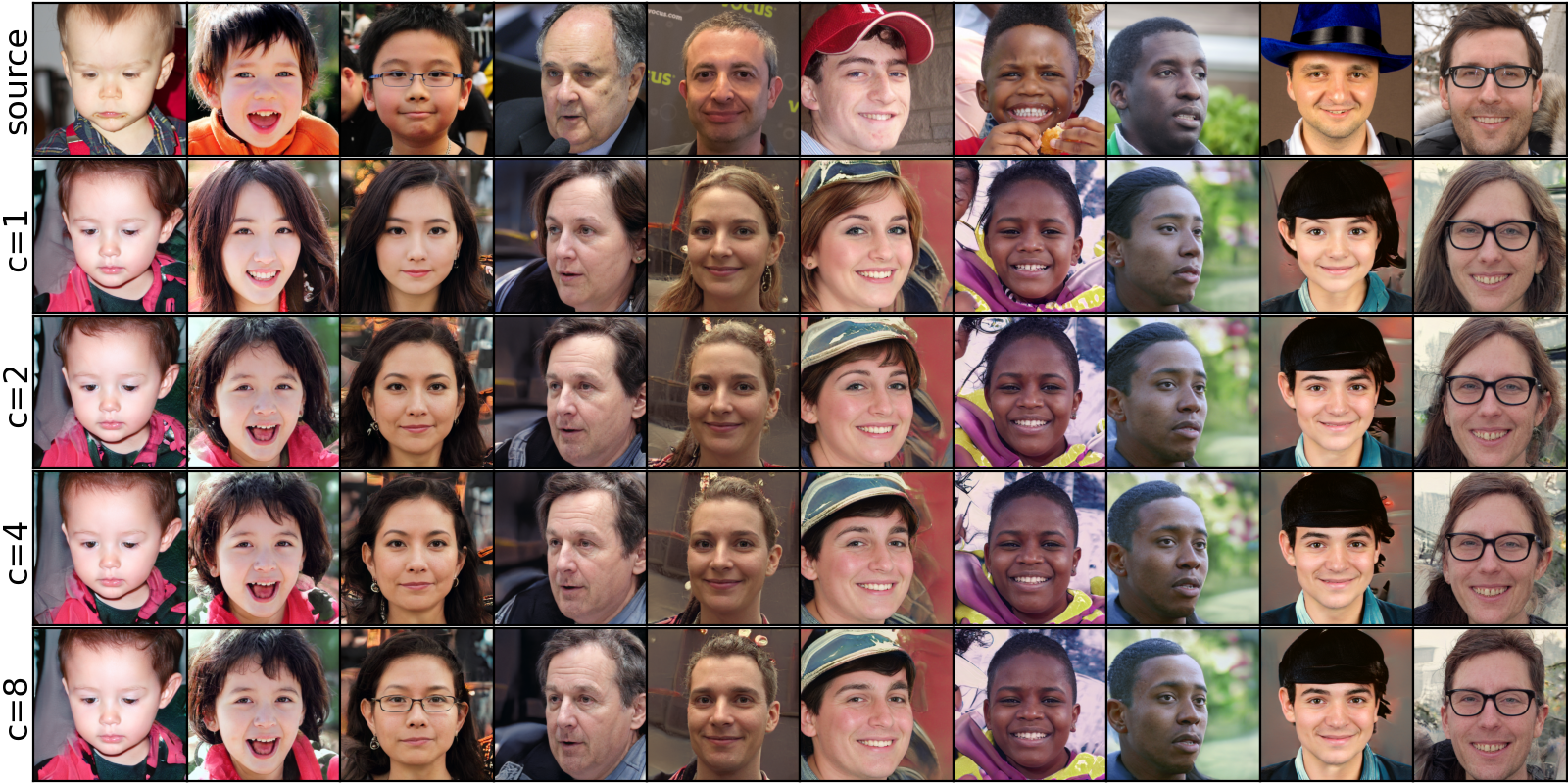}
        \caption{Static subset selection}
        \label{fig:man-woman-static}
    \end{subfigure}\hfill
    \begin{subfigure}[t]{0.49\textwidth}
        \centering
        \includegraphics[width=\linewidth]{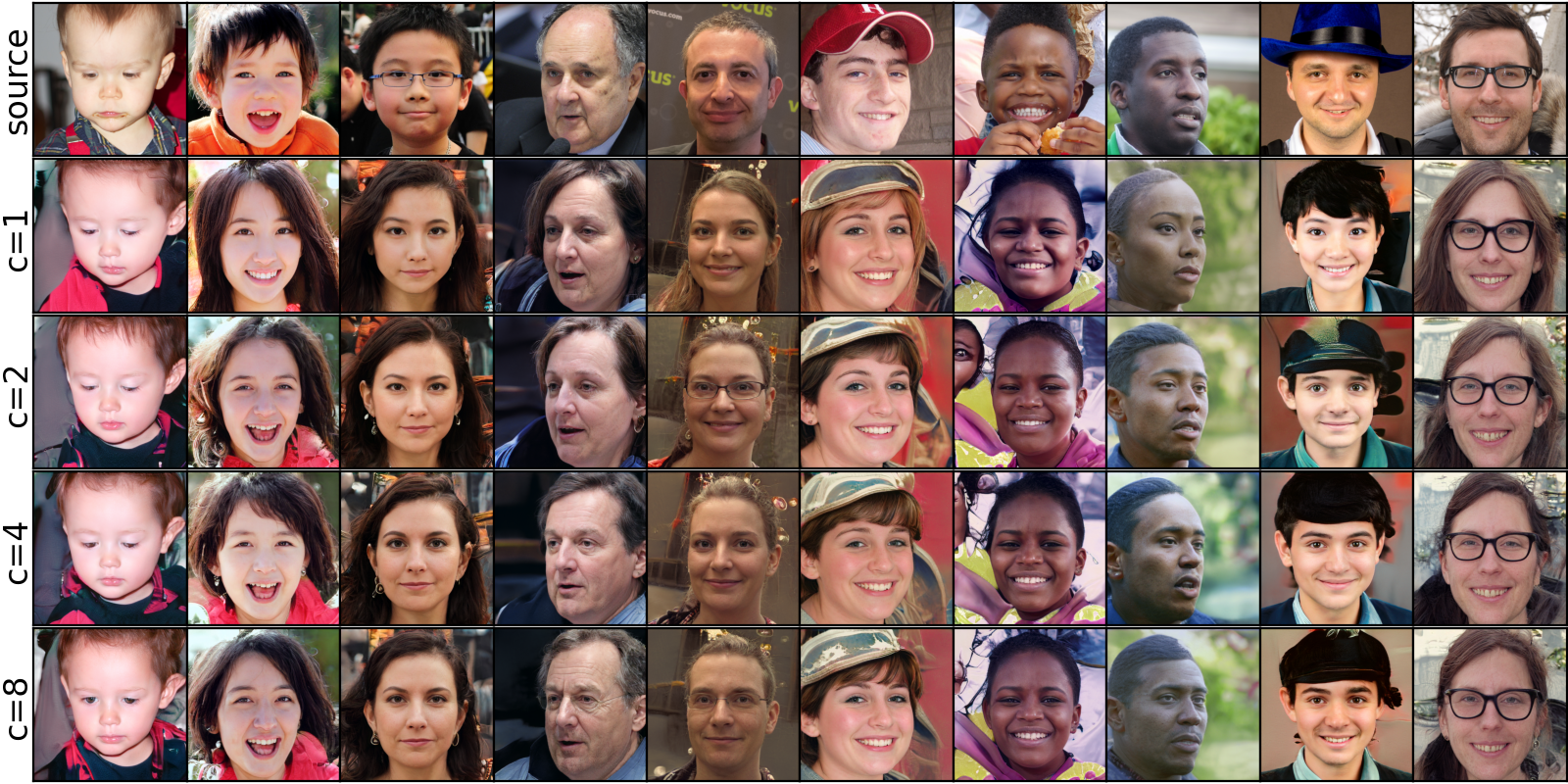}
        \caption{Dynamic subset selection}
        \label{fig:man-woman-dynamic}
    \end{subfigure}

    \caption{FFHQ man$\rightarrow$woman translation using (a) static and (b) dynamic subset selection. Dynamic subset selection is evaluated using Euler integration with 100 steps.}
    \label{man-to-woman}
\end{figure}

\end{document}